\documentclass{article}

\usepackage[preprint]{neurips_2022}
\usepackage[utf8]{inputenc} 
\usepackage[T1]{fontenc}
\usepackage{url}
\usepackage{booktabs}
\usepackage{amsfonts}
\usepackage{nicefrac}
\usepackage{microtype}
\usepackage{xcolor}     
\usepackage{caption}
\usepackage{titlesec}
\usepackage{subcaption}
\usepackage{graphicx}
\usepackage{hyperref}
\usepackage{multirow}
\usepackage{wrapfig}
\usepackage{floatrow}
\usepackage{amsmath}
\usepackage{hyperref}
\usepackage{url}
\usepackage{microtype}
\usepackage{graphicx}
\usepackage{booktabs}
\usepackage{hyperref}
\usepackage{url}
\usepackage{caption}
\usepackage{subcaption}
\usepackage{wrapfig}
\usepackage[capitalize]{cleveref}
\newfloatcommand{capbtabbox}{table}[][\FBwidth]
\usepackage{blindtext}
\definecolor{mydarkblue}{rgb}{0,0.08,0.45}

\definecolor{mygray}{gray}{0.4}
\newcommand{\g}[2]{#1\textsubscript{\textcolor{mygray}{$\pm$#2}}}
\newcommand{\highlight}[1]{\colorbox{blue!10}{#1}}

\titlespacing*{\paragraph}
{0ex}{-0.5ex}{0.75ex}

\hypersetup{colorlinks=true,
    linkcolor=mydarkblue,
    citecolor=mydarkblue,
    filecolor=mydarkblue,
    urlcolor=mydarkblue}

\title{On Neural Architecture Inductive Biases for Relational Tasks}

\author{Giancarlo Kerg\textsuperscript{1} \thanks{Correspondence to Giancarlo Kerg:  \href{mailto:giancarlo.kerg@gmail.com}{giancarlo.kerg@gmail.com}, Guillaume Lajoie: \href{mailto:g.lajoie@umontreal.ca}{g.lajoie@umontreal.ca} \newline\hspace*{1.35em} \textsuperscript{1}Universit\'e de Montr\'eal, \textsuperscript{2}McGill University, \textsuperscript{3}CIFAR Senior Fellow,\ \textsuperscript{4}CIFAR AI Chair,\newline\hspace*{1.75em}Code is available at: \href{https://github.com/giancarlok/relationaltasks}{\texttt{https://github.com/giancarlok/relationaltasks}}} \quad Sarthak Mittal \textsuperscript{1} \quad David Rolnick \textsuperscript{2,4} \quad Yoshua Bengio \textsuperscript{1,3,4}\\[0.2em]
\textbf{Blake Richards}\,\textsuperscript{2,4} \quad \textbf{Guillaume Lajoie}\,\textsuperscript{1,4,*}\\[0.5em]
Mila, Quebec AI Institute
}

\begin{document}

\maketitle

\begin{abstract}
Current deep learning approaches have shown good in-distribution generalization performance, but struggle with out-of-distribution generalization. 
This is especially true in the case of tasks involving abstract relations like recognizing rules in sequences, as we find in many intelligence tests.
Recent work has explored how forcing relational representations to remain distinct from sensory representations, as it seems to be the case in the brain, can help artificial systems.
Building on this work, we further explore and formalize the advantages afforded by ``partitioned'' representations of relations and sensory details, and how this inductive bias can help recompose learned relational structure in newly encountered settings. 
We introduce a simple architecture based on similarity scores which we name \textit{Compositional Relational Network (CoRelNet)}.
Using this model, we investigate a series of inductive biases that ensure abstract relations are learned and represented distinctly from sensory data, and explore their effects on out-of-distribution generalization for a series of relational psychophysics tasks. We find that simple architectural choices can outperform existing models in out-of-distribution generalization.
Together, these results show that partitioning relational representations from other information streams may be a simple way to augment existing network architectures' robustness when performing out-of-distribution relational computations.
\end{abstract}

\section{Introduction}
\looseness=-1
Current deep learning systems have performed astonishingly well on a multitude of domains, ranging from natural language \citep{devlin2018bert,child2019generating,dai2019transformer} and speech recognition \citep{pratap2019wav2letter++,oord2016wavenet} to image classification \citep{dosovitskiy2020image,he2015delving}. This progress has been obtained with extensive compute and data \citep{brown2020language}. However, a line of research shows that deep learning models still struggle to perform well in tasks that require abstract relational reasoning, especially in low data regimes and out-of-distribution (OoD) settings \citep{ESBN,kim2020cogs,johnson2017clevr,yi2019clevrer,newman2020eos,nogueira2021investigating,mittal2021compositional,ke2021systematic}.

\looseness=-1
A working hypothesis we explore here is that most artificial systems do not work well on relational reasoning because they do not encode an explicit notion of relations between different objects, and rather rely too much on representations of object features. There is a body of machine learning approaches exploiting relational structure \citep{scarselli2008graph, velivckovic2017graph, kipf2018neural, battaglia2018relational, bengio2019meta, TCN, zhang2019raven}, but the majority of machine learning systems do not explicitly encode such relations. In contrast, in our brains, relational dependencies across high-level entities form a key ingredient to our understanding of the world and augment our ability to reason in potentially unseen scenarios \citep{whittington2019tolman}, 
For example, the relationship between a parent and their child often generalizes to different animal groups and thus explicit modeling of relationships between objects could lead to better out-of-distribution (OoD) generalization, since a number of generic relation types generalize over multiple domains. 

Recently, there has been substantial work on attention-based systems \citep{Bahdanau2015NeuralMT,luong2015effective,vaswani2017attention,hudson2018compositional} that do compute a notion of similarity between objects, which can be understood as relations between said objects. For example, transformers \citep{vaswani2017attention} compute multiple parallel relational connections between the objects in a scene \citep{dosovitskiy2020image,image-transformer}. However, the relational connections computed by transformers are only used as a means to route information between different objects (i.e.~via attention scores), and they are not explicitly leveraged for predictions themselves.
However, some work does address this shortcoming by explicitly focusing on encoding relational information between objects. One approach is to group objects with similar features, and to attend to these groupings based on context \citep{santoro2017simple,locatello2020object,santoro2018relational}.
This has shown promising aptitudes, but suffers from a few issues such as scaling, and may generalize poorly in OoD settings where group membership is not straightforward. Another approach, first outlined by~\citet{ESBN} with the Emergent Symbols through Binding in Memory (ESBN) method, leverages the abstract nature of relations (e.g.~parent and child) by explicitly segregating sensory information from relational encoding. While showcasing the importance of separate sensory and relational encodings, ESBN also includes a number of architectural design choices that may or may not be important for OoD generalization in relational tasks.

Here we build on previous relational approaches and explore in depth the inductive bias of separating sensory information from abstract, relational representations. We do so using a model that we call  \textit{Compositional Relational Network (CoRelNet)}. CoRelNet uses the simplest possible architecture to distill a minimal set of inductive biases that allow for OoD generalization in relational tasks. Thanks to this simplicity, CoRelNet allows for considerably more robust performance and generalization than several benchmarks, including ESBN. Through a series of numerical experiments involving relational tasks on data with spurious features, scaled input size, and variable contexts, we demonstrate and analyze why the simple act of building representations of similarity relations between objects, irrespective of sensory encoding, is enough to improve a range of widely used model architectures (e.g.~multi-layer perceptrons and transformers). Thus, CoRelNet variants confirm the importance of abstract relational encoding and helps to establish minimal inductive biases to enhance other architectures. We find that focusing on these inductive biases to design new and simpler architectures considerably improves OoD generalization across all settings tested, and thus is a promising approach for other machine learning researchers to adopt.

\section{Motivation and Related Work}
\looseness=-1
Human cognition is heavily dependent on the ability to understand relationships between different entities in the world, regardless of their sensory attributes~\citep{Kriete16390}. This ability allows us to excel at a number of tasks that require understanding of abstract rules that govern the relationships between objects in a way that abstracts out the purely perceptual qualities of those objects (see e.g.~\citep{JONES2007196}). For example, people can easily grasp that different institutions can have the same organizational structure, e.g.~different high schools will have a principal, teachers, and students with similar relations between these individuals regardless of the specific people who hold those roles. 

Ongoing work in neuroscience aims to understand how our brains represent such abstract relationships. Early results identified cognitive maps~\citep{tolman48} and so-called relational memories~\citep{cohenEichenbaum93} which seem to encode abstract relational information invariant to some perceptual features (see also \citet{sedda2012dorsal, goodale1992separate}). While this abstraction of relational knowledge from memory is not entirely disconnected from sensory modality~\citep{BARSALOU200384}, there is evidence that isolated perceptual features alone are not principally driving relational reasoning, and that complex interdependence of processed features and memory gives rise to abstraction~\citep{Goldstone89relationsrelating}.

Consistent with this neuro-cognitive picture, AI systems which work well on a variety of domains like machine translation~\citep{vaswani2017attention,universal_tsf,devlin2018bert}, image classification~\citep{dosovitskiy2020image}, etc.~do not perform as well on tasks that explicitly require inference of relational structure between entities~\citep{ESBN,johnson2017clevr,yi2019clevrer}. An example of such a task is Raven's Progressive Matrices \citep{raven1938raven} which, given a sequence of objects, tests the ability to infer the relations between objects and to use it to select candidate objects that satisfy the underlying relational structure. In response to this shortcoming, a number of AI elements inspired from neuroscience have been proposed to tackle relational abstractions. A fruitful approach has been to endow AI systems with memory and with various mechanisms that enforce abstract representations~\citep{santoro2018relational,graves2014neural,mittal2020learning,pritzel2017neural,hill2020grounded,fortunato2019generalization}. Recent work combines this approach with the explicit separation of memory-stored representations and sensory inputs in the ESBN model~\citep{ESBN} (see also~\citet{whittington2019tolman}), and reveals impressive performance and OoD generalization properties on relational tasks. 
In this work, we develop CoRelNet, with the goal of identifying a minimal architecture needed to impart an effective inductive bias towards OoD generalizable relational representations.
\section{Relational Tasks}
\label{sec:basic_tasks}
In this section we describe the collection of tasks used to evaluate relational learning, these range from commonly used tasks in cognitive tests to tasks purposely designed to evaluate artificial systems. 

\subsection{Relational Cognitive Tests}
\label{sec:basic_psychophysics_tasks}
\begin{wrapfigure}{r}{0.6\textwidth}
    \centering
    \vspace{-6mm}
    \includegraphics[width=\textwidth]{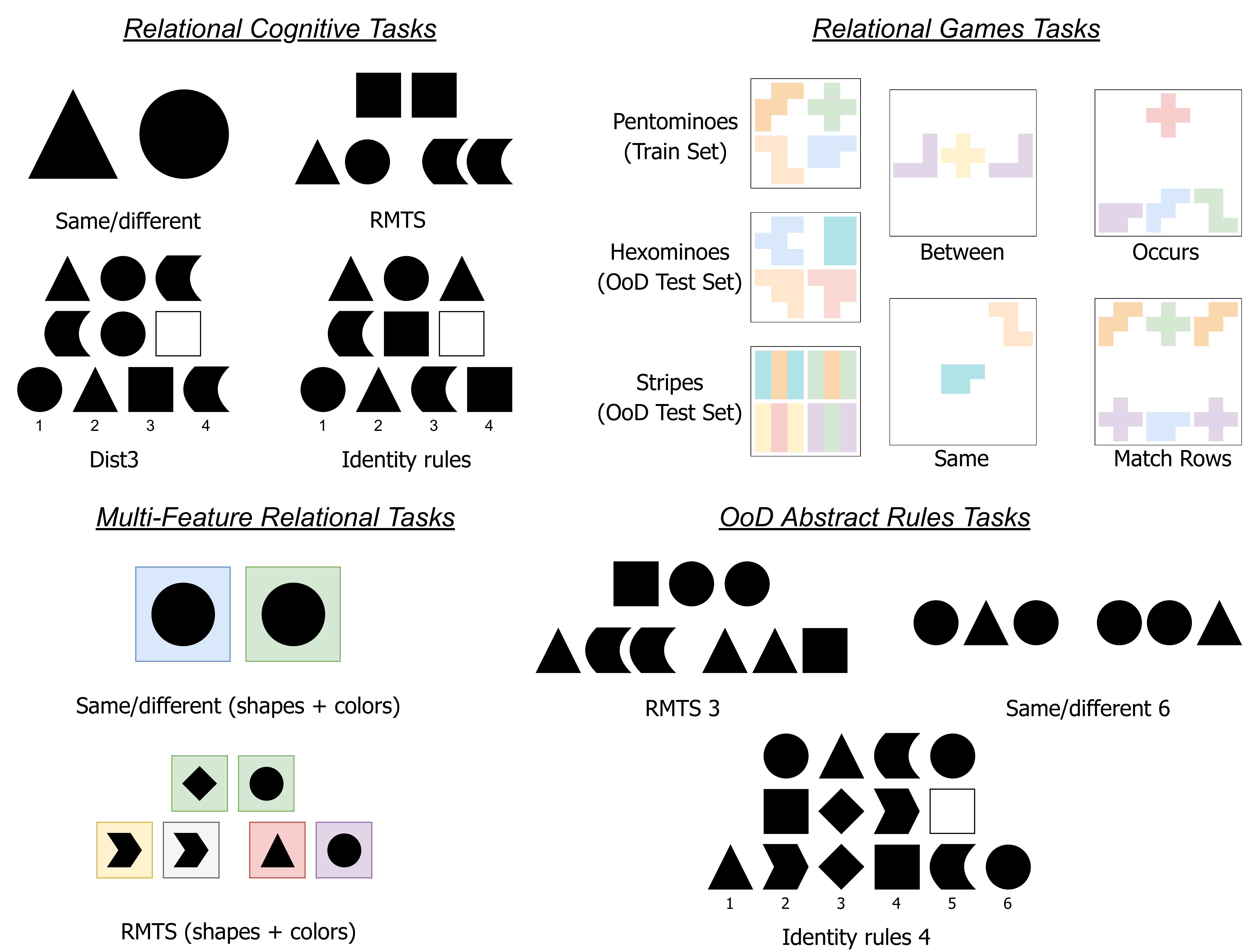}
    \caption{{\small\textbf{Top left.} Relational cognitive tasks. Same/different discrimination task (answer: ``different''). RMTS task (here the source pair (top row) has two identical objects matching the second candidate pair in the bottom row). Dist3 task, choosing a candidate in the bottom row to replace the white square (answer: 2). Identity rules task (ABA pattern, answer: 3). \textbf{Top right.} Relational Games tasks. Example objects from the training set ('pentominoes'), two held-out test sets ('hexominoes' and 'stripes'), as well as examples from the tasks 'between' (label: True), 'occurs' (label: False) ,'same' (label: False), 'match rows' (label: True, matching pattern: ABA) \textbf{Bottom left.} Examples from relational tasks with a spurious feature (background color). In same/diff. (shapes+colors), the label is 'same' as the two shapes are identical. In RMTS (shapes+colors), the source pair (top row) has two distinct shapes, matching the second candidate pair in the bottom row. \textbf{Bottom right.} In RMTS3, the source triplet matches the first target triplet. In same/diff 6, the two triplets are 'different'. In identity rules 4, the answer is 4.}}
    \vspace{-2mm}
    \label{fig:task_figures}
\end{wrapfigure}
We first describe a set of four relational tasks which are used to assess abstract relational reasoning in cognitive tests. In each of the tasks, object images are provided sequentially such that some specific abstract rule governs the relationship among them. To test for OoD generalization, the objects used for training are different from those used during testing while keeping the abstract rule consistent. See Figure~\ref{fig:task_figures} for illustrations. Even though these tasks might appear simple at first, it is indeed very challenging for deep neural network architectures to generalize well in OoD settings like this~\citep{not-so-clevr, ESBN}, including LSTM, Transformers, Relation Networks, Predi-Net, ESBN and others.

\paragraph{Same/different discrimination.} Two objects are shown, and the task is to determine whether they are the same or different. 

\looseness=-1
\paragraph{Relational match-to-sample (RMTS).} Three pairs of objects are presented: a \textit{source} pair and two \textit{target} pairs. The task consists of identifying which target pair has the same relational structure as the source pair, i.e., if the source pair consists of two identical/different objects, the task is to identify the target pair which contains two identical/different objects. 

\paragraph{Distribution-of-three (Dist3).} Two rows of three objects followed by a row of four objects are presented. The same set of 3 (distinct) objects that are shown in the first row are in the second row in a permuted order, but with the last object masked. The task is to identify the missing object of the second row from the set of four objects shown in the third row. See~\citet{carpenter1990one} for more details.

\paragraph{Identity rules.} Objects are presented in the same manner as in Dist3. However here, the first row contains a sequential pattern of objects (e.g. ABA or ABB or AAA). The task is to pick the object from the last row which completes the second row's sequence such that the sequential pattern of the first row matches the one of the second row. See~\citet{marcus1999rule} for more details.

\subsection{Relational Games}
\label{sec:basic_relationalgames_tasks}
This is a set of binary classification tasks first outlined by \citet{shanahan2019explicitly}. Each Relational Game is governed by some underlying abstract rule (see list below) and involves the presentation of an image containing objects laid out on a $3\times 3$ grid. An image is labelled as ``True" if the objects in the image satisfy the abstract rule, and False otherwise. The training set consists of simple geometric shapes called ``pentominoes", and we use two held-out OoD test sets consisting of different shapes called  ``hexominoes" and ``stripes". See Figure~\ref{fig:task_figures} for illustrations. Contrary to the implementation by \citet{shanahan2019explicitly}, and more in line with the tasks above, here the $3\times 3$ grid is presented sequentially. The rules considered are as follows.

\paragraph{Same/Different.} Objects are ``same" if they have the same shape, colour and orientation.

\paragraph{Between.} This relation holds if the image is composed of three objects displayed in a straight line where the outer objects are ``same".

\paragraph{Occurs.} This relation holds if the object in the top row is the ``same" as one of the objects in the bottom row.

\paragraph{Row Matching.} This relation holds that the first row and last row of objects are governed by the same underlying abstract pattern such as AAB, AAA, or ABB.

\paragraph{Xoccurs.} This relation holds if the object in the top row is  the ``same" as one of the objects in the bottom row, while the other two objects in the bottom row are different.

\paragraph{Colour/Shape.} This relation has four labels: same-same, same-different, different-same and different-different, depending on whether the two objects shown in the image have the same or different colour/shape.

\paragraph{Left-of. } This positional task is slightly different. Only two objects with different luminance are presented somewhere in the $3 \times 3$ grid (other places are left blank). The relation holds if the object of lower luminance is to the 'left-of' the other object. Note that this relation is antisymmetric, since interchanging the objects changes the label of the image.

\subsection{Enhanced Relational Cognitive Tests}
\label{sec:harder_unseen_psychophysics_tasks}
\label{sec:harder_spurious_psychophysics_tasks}
We modify tasks from \ref{sec:basic_psychophysics_tasks} to probe more difficult and abstract relations in order to further explore OoD generalization and model limitations. Modifications probe two aspects: unseen relations, and spurious correlations/distractors.

\paragraph{Unseen Relations.} We construct tasks where we restrict not only the set of objects but also the set of relations seen during training, while testing on unseen objects as well as unseen relations. These include 
{\bf RMTS 3:} similar to RMTS where for training only triplets of the form ($AAA$, $ABA$ and $BAA$) are shown, while testing \textit{only} involves triplets of the form ($ABC$, $AAB$). 
{\bf Same/diff 6:} Similar to Same/Different but comparing 2 triplets instead of single objects. Each triplet consists of at most two distinct objects (thereby of forms $AAA$, $AAB$, $ABA$ or $BAA$). During training, only triplets of the form ($AAA$, $ABA$, $BAA$) are shown, while evaluation is done on examples that have at least one triplet of the form ($AAB$).
{\bf Identity rules 4:} Similar to 'Identity rules' but with rows of 4 objects and multiple choice of 6. 
Note that for each quadruple in row 1 or 2, we can have at most three distinct objects. During training, only quadruples with at most two distinct objects are shown, while testing relies \textit{only} on quadruples with exactly three distinct objects.
{\bf Identity rules 4 Missing:} Identical to Identity rules 4 but the training set does not include quadruples of the form ($ABAA$, $ABAB$). See also Appendix~\ref{app: idrules4_flipped} where we include another variation called \textit{identity rules 4 [flipped]} where testing and training sets are swapped.
{\bf Distribution-of-$N$:} Same as Dist-3 but with $N$ objects. See also Appendix~\ref{app: distN_results} for results. 

\paragraph{Spurious Correlations.} In this last set of tasks, we add spurious features to objects. We again consider the four relational tasks from Section~\ref{sec:basic_psychophysics_tasks}, but this time we provide each object with a (distracting) background color. The tasks consist of identifying the same relational structure but based \textit{solely} on shapes.

\section{Models}

Using tasks described in Section~\ref{sec:basic_tasks}, we compare five models\footnote{All models were trained on single RTX8000 GPUs and each experiment took a few hours.}: Transformers, PrediNet, ESBN, and two versions of our model: CoRelNet, and CoRelNet-T. All the models use an encoder $q$ which is sequentially applied to each of the inputs $\{x_t\}_{t=1}^{T}$ followed by a temporal context normalization (TCN) step, which has shown to significantly improve OoD generalization in relational reasoning tasks \citep{TCN}.
$$\{x_t\}_{t=1}^{T} \longmapsto q(\{x_t\}_{t=1}^{T}) \longmapsto \textrm{TCN}(q(\{x_t\}_{t=1}^{T})) = \{z_t\}_{t=1}^{T}$$
The encoded inputs $\{z_t\}_{t=1}^{T}$ are then being fed to the respective models.
\paragraph{Transformer.} Introduced in \cite{vaswani2017attention}, transformers rely on multiple parallel attention mechanisms (Multi-Head Attention or MHA) to route information between different input tokens, which can be elements in a set or in a sequence, etc. While these systems do compute multiple similarity matrices in parallel, these are further combined with sensory information to drive final prediction and hence the dichotomy between explicit relational and sensory structure is absent. Prior work \citep{ESBN} hypothesizes that the absence of this dichotomy leads to poor OoD generalization, and we analyze this further through our ablations in Section \ref{sec:results}.

\paragraph{PrediNet.} Introduced by \citet{shanahan2019explicitly}, PrediNet leverages Multi-Head Attention (MHA) to map inputs onto objects, relations and propositions, thereby attempting to extract the underlying relational structure via propositional representations compatible with predicate calculus. 
\begin{wrapfigure}{r}{0.45\textwidth}
    \vspace{-30mm}
    \centering
    \includegraphics[width=\textwidth]{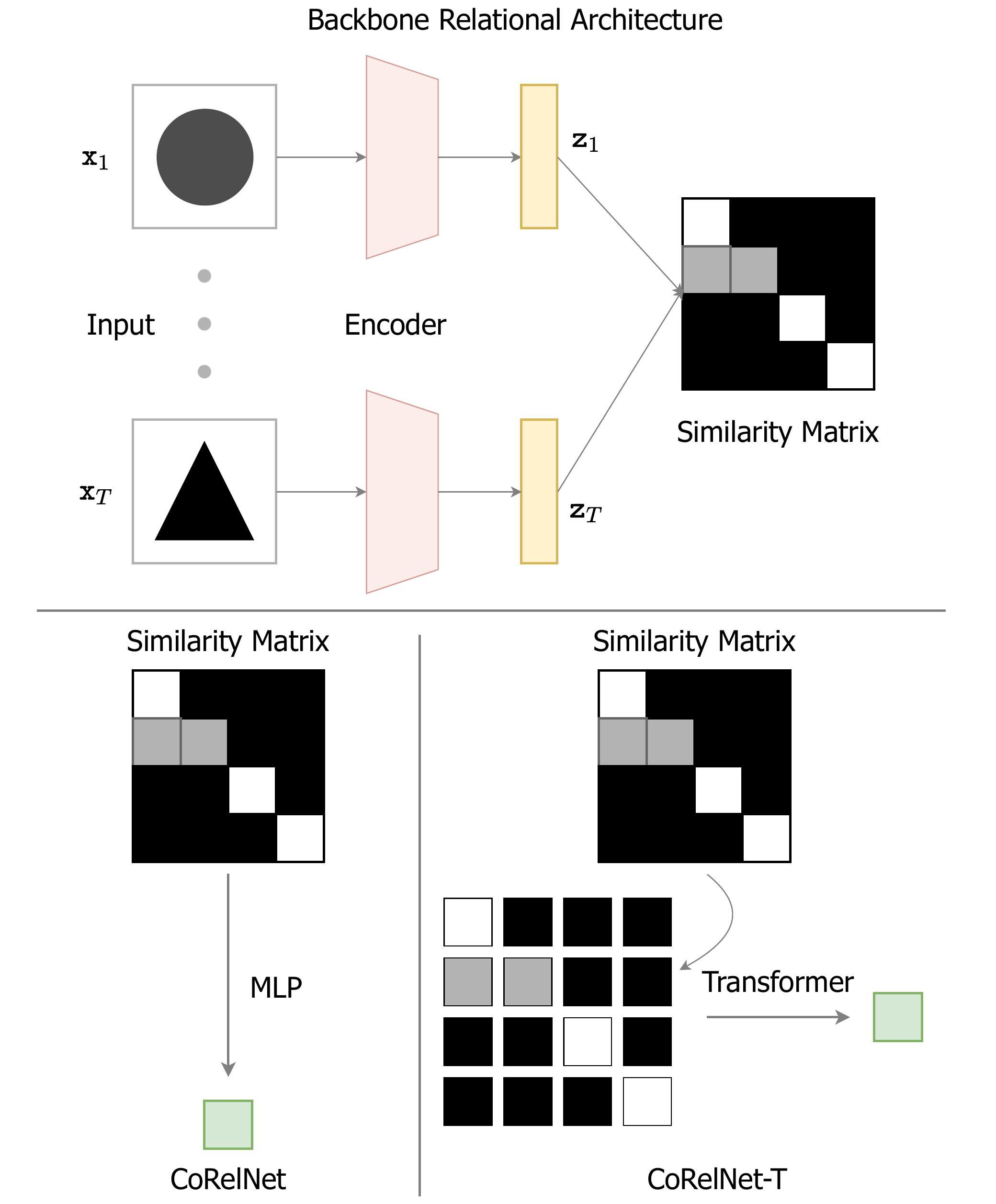}
    \caption{{\small \textbf{Illustration of Relational Architecture.} (top) The backbone relational architecture describes the common backbone present in all of the relational architectures considered in this work, with a similarity matrix with entries for each pair of input objects. (bottom) CoRelNet considers an MLP to process the similarity matrix obtained at the end of the backbone while CoRelNet-T uses a Transformer.}}
    \label{fig:model_pic}
    \vspace{-2mm}
\end{wrapfigure}

\paragraph{ESBN.} Emergent Symbol Binding Network introduced in \citet{ESBN} constructs an external memory through a recurrent controller and is the first model to make use of the separation of sensory and relational information. This means the model is separated into two distinct information processes. An LSTM is used to process relational data, incurring additional processing.

\paragraph{CoRelNet.} In our main model, \textit{CoRelNet}, we compute the self-attention coefficients $$\{R_{i,\cdot}\}_{i=1}^T = \{\textrm{softmax}(z_i^\top \cdot \mathbf{z})\}_{i=1}^T$$ over the encoded inputs $\{z_t\}_{t=1}^{T}$, and feed those \textit{directly} into an MLP decoder to produce the final output $o$ (see Figure \ref{fig:model_pic}):
$$R \overset{\textrm{flatten}}{\longmapsto} \textrm{flatten}(R) \overset{\textrm{MLP}}{\longmapsto} o$$

First note that self-attention is computed via dot products and hence can be interpreted as giving rise to a (symmetric) similarity measure encoding relational information of the objects in the input sequence. Since the input to the MLP decoder is only composed of self-attention coefficients $R_{i,j}$ (carrying the relational information), and not any sensory information, we make use of the mentioned important inductive bias of \textit{separating sensory details from relational information}. Meanwhile, the MLP applies independent weights to each $R_{i,j}$ which allows for maximal flexibility and enables the model to remain sensitive to positional information when making use of the mentioned relational information provided. This advantage will be of specific use in later experiments (see Figure \ref{fig:results_unseen_main})

\paragraph{CoRelNet-T.} This is a variation of the CoRelNet model where the MLP decoder is replaced by a Multi-Head Attention layer (Transformer) with $8$ heads to produce the final output. Each row of the similarity matrix is treated as an individual token, and these tokens, combined with learned positional embeddings, are fed to the transformer system to provide the prediction (see Figure \ref{fig:model_pic}; \textit{bottom right}). 

In both CoRelNet and CoRelNet-T, we also explore different sensory encoding schemes based on either a learned or a random encoder, as part of our analysis which is described in more detail in Section \ref{sec:results}. We also refer the reader to Appendix \ref{app:hyperparameters} for hyperparameters and more architectural details.

\begin{table}[t]
\renewcommand{\arraystretch}{1.}
\centering
\resizebox{\columnwidth}{!}{
\begin{tabular}{ c  c | c c | c c c } 
\toprule
\textit{Task} & \textit{Test Set} & \textit{CoRelNet} & \textit{CoRelNet-T} & \textit{ESBN} & \textit{PrediNet*} & \textit{Transformer}\\
\midrule
\multirow{2}{4em}{same} & Hex. & \g{94.7}{5.4} & \g{98.7}{0.5} & \g{60.9}{18.0} & \highlight{\g{99.1}{0.5}} & \g{96.3}{3.6}\\ 
& Str. & \g{90.4}{9.0} & \highlight{\g{98.4}{0.6}} & \g{60.5}{17.0} & \g{97.7}{1.5} &  \g{93.6}{4.5} \\ 
\midrule
\multirow{2}{4em}{between} & Hex. & \highlight{\g{96.6}{2.2}} & \g{91.9}{3.3} & \g{79.5}{2.7} & \g{94.4}{3.6} & \g{90.9}{4.2}\\ 
& Str. & \highlight{\g{93.1}{5.4}} & \g{87.0}{4.7} & \g{82.5}{4.5} & \g{92.4}{5.4} & \g{87.5}{6.4} \\ 
\midrule
\multirow{2}{4em}{occurs} & Hex. & \highlight{\g{96.2}{2.2}} & \g{91.6}{4.6} & \g{74.3}{0.9} & \g{92.9}{3.3} & \g{95.9}{2.4}\\ 
& Str. & \highlight{\g{88.7}{5.3}} & \g{79.3}{12.1} & \g{73.3}{0.9} & \g{87.4}{6.5} &  \highlight{\g{88.7}{5.4}} \\ 
\midrule
\multirow{2}{4em}{xoccurs} & Hex. & \highlight{\g{92.2}{6.4}} & \g{91.7}{6.9} & \g{63.0}{0.8} & \g{67.2}{8.7} & \g{75.1}{11.9}\\ 
& Str. & \g{83.6}{10.9} & \highlight{\g{85.4}{6.4}} & \g{64.7}{2.1} & \g{61.3}{6.3} & \g{69.4}{13.0} \\ 
\midrule
\multirow{2}{4em}{row matching} & Hex. & \highlight{\g{97.7}{0.8}} & \g{95.4}{5.1} & \g{81.1}{2.7} & \g{50.3}{0.5} & \g{50.3}{0.6}\\ 
& Str. & \highlight{\g{94.8}{1.3}} & \g{90.5}{5.2} & \g{76.8}{4.3} & \g{50.5}{0.5} &  \g{50.4}{0.5} \\ 
\midrule
col./shape & Hex. & \g{47.2}{3.7} & \g{49.6}{0.8} & \g{31.2}{10.1} & \g{74.9}{10.6} &  \highlight{\g{89.1}{2.9}}\\ 
\midrule
left-of & Hex. & \highlight{\g{99.2}{0.7}} & \g{97.6}{1.2} & \g{49.9}{0.3} & \g{94.9}{0.9} & \g{96.0}{1.9}\\ 
\bottomrule
\end{tabular}
}
\caption{{\small Out-Of-Distribution test accuracies for the Relational Game tasks on the two held-out sets "Hexominoes" (Hex.) and "Stripes" (Str.). Results reflect test accuracies averaged over 10 seeds (Details in Appendix~\ref{app:relational_games}).}}
    \label{table:predinet_basic_results}  
\end{table}

\section{Results}
\label{sec:results}
\begin{wrapfigure}{r}{0.48\textwidth}
\vspace{-16mm}
\includegraphics[width=\textwidth]{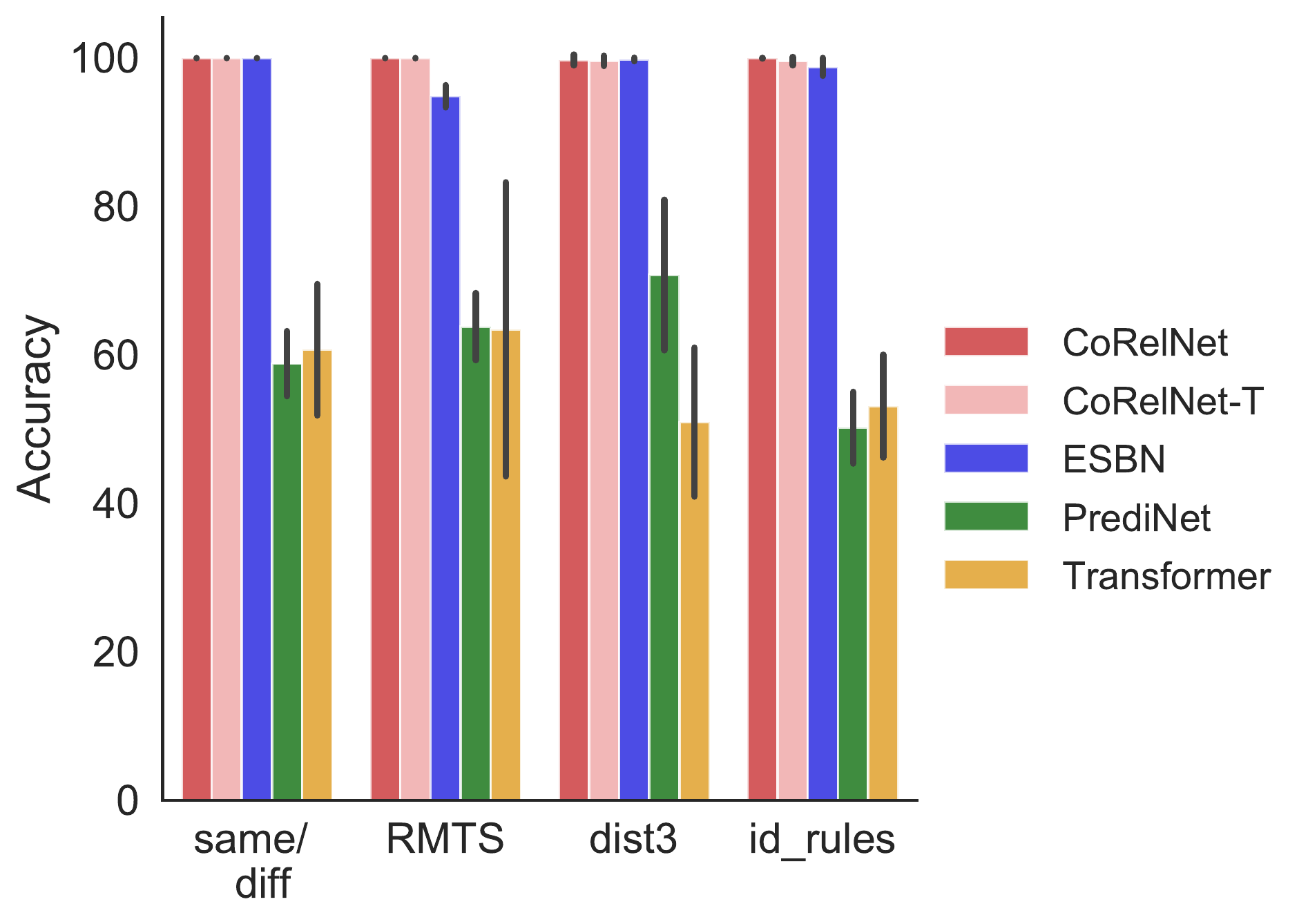}  
\vspace{3mm}
\caption{\small Test accuracies on the relational tasks. Results are displayed for OoD cases for each task. Results reflect test accuracies averaged over 10 seeds. For details see Figure \ref{fig:full_results_basic_tasks} in the appendix.}
\vspace{-12mm}
\label{fig:results_main}
\end{wrapfigure}
We note that given the structure of the relational cognitive tasks, the ground-truth prediction is completely de-coupled from the actual shapes of individual objects, and relies entirely on relations between them. In this section, we analyze the specific inductive biases that allow our models to perform well in OoD settings that respect this assumption.

\begin{figure}
\begin{minipage}{0.32\columnwidth}
    \centering
  \includegraphics[width=\textwidth]{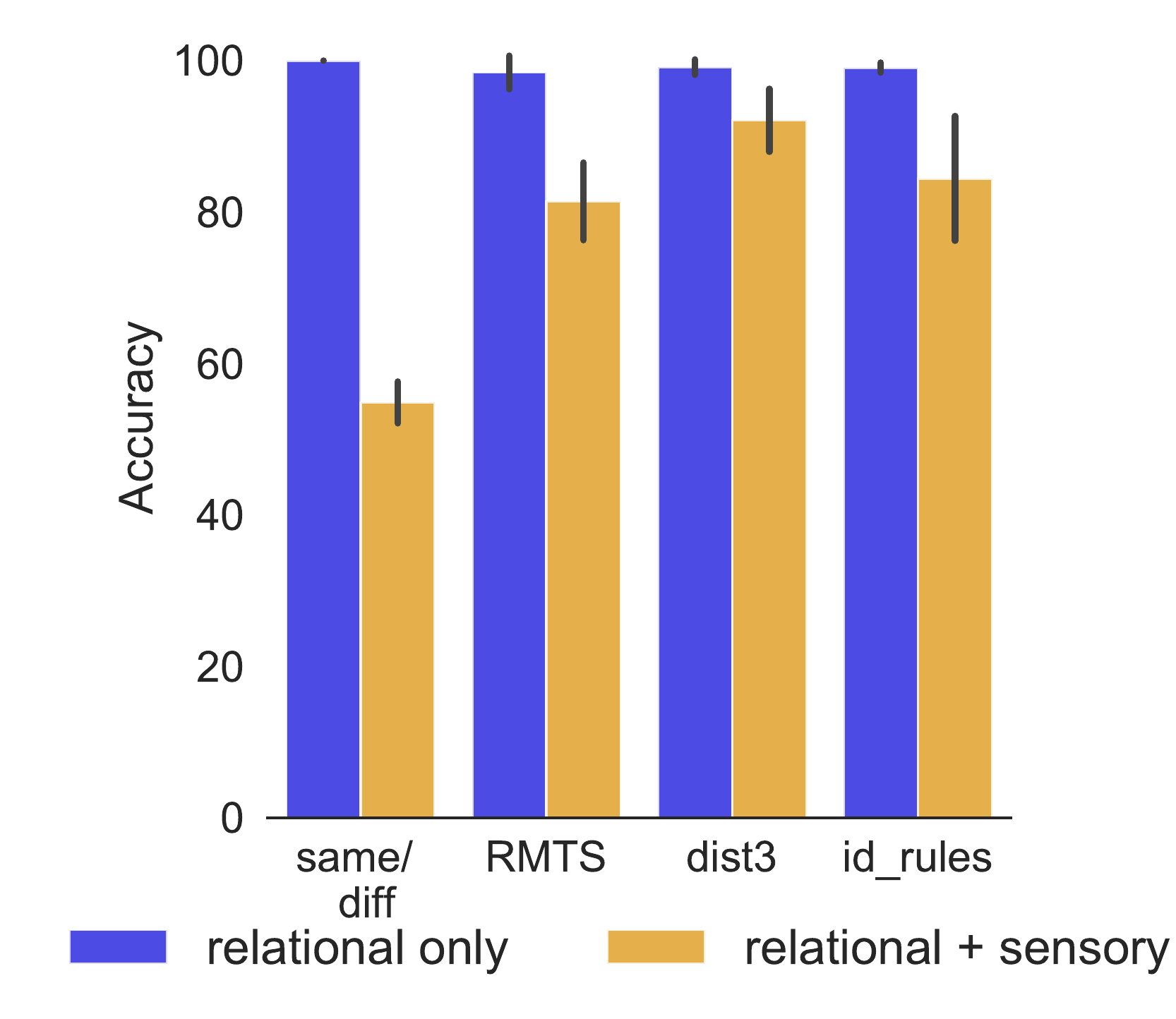}  
\end{minipage}
\begin{minipage}{0.32\columnwidth}
    \centering
  \includegraphics[width=0.9\textwidth]{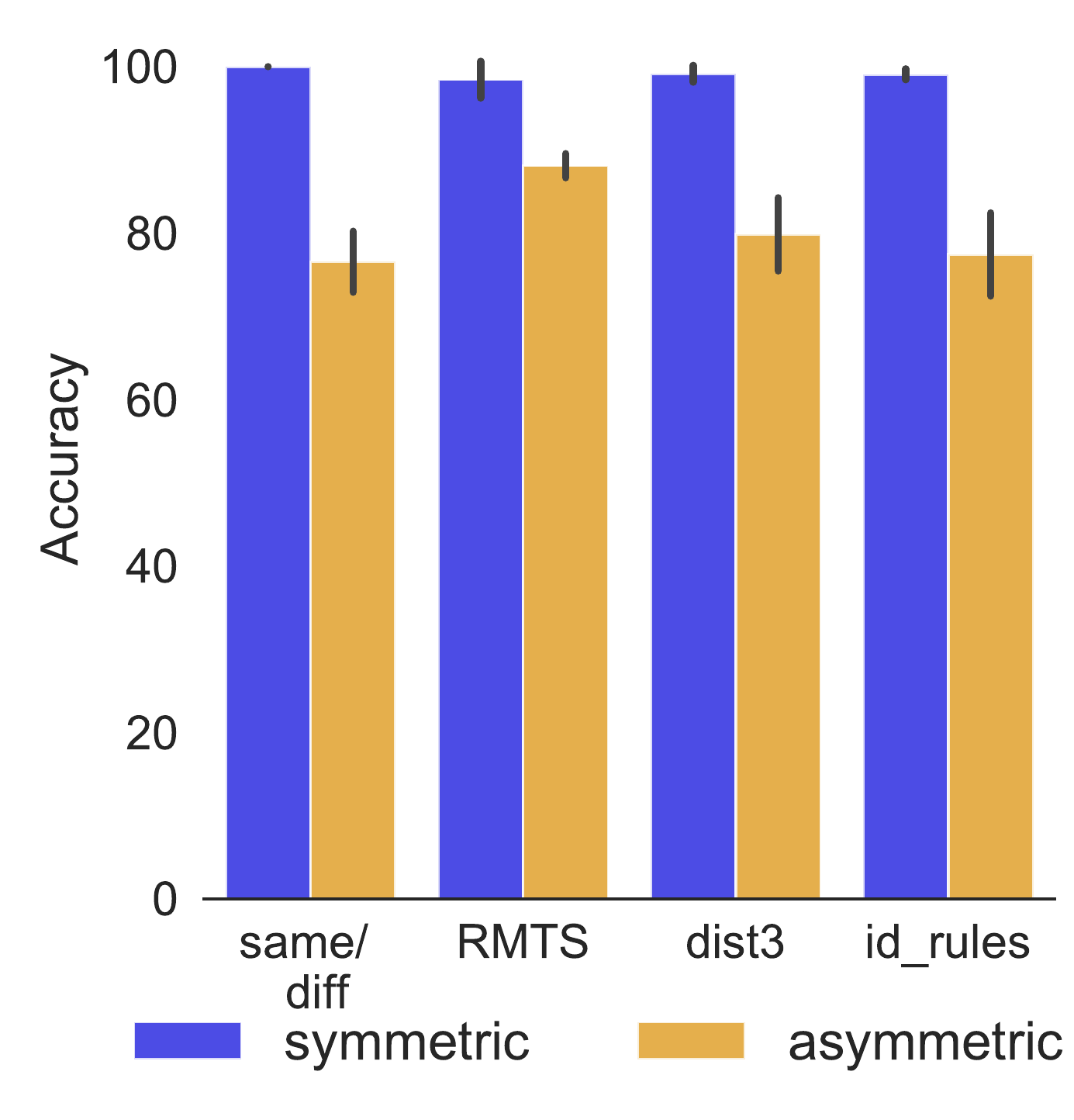}  
\end{minipage}
\begin{minipage}{0.32\columnwidth}
    \centering
  \includegraphics[width=\textwidth]{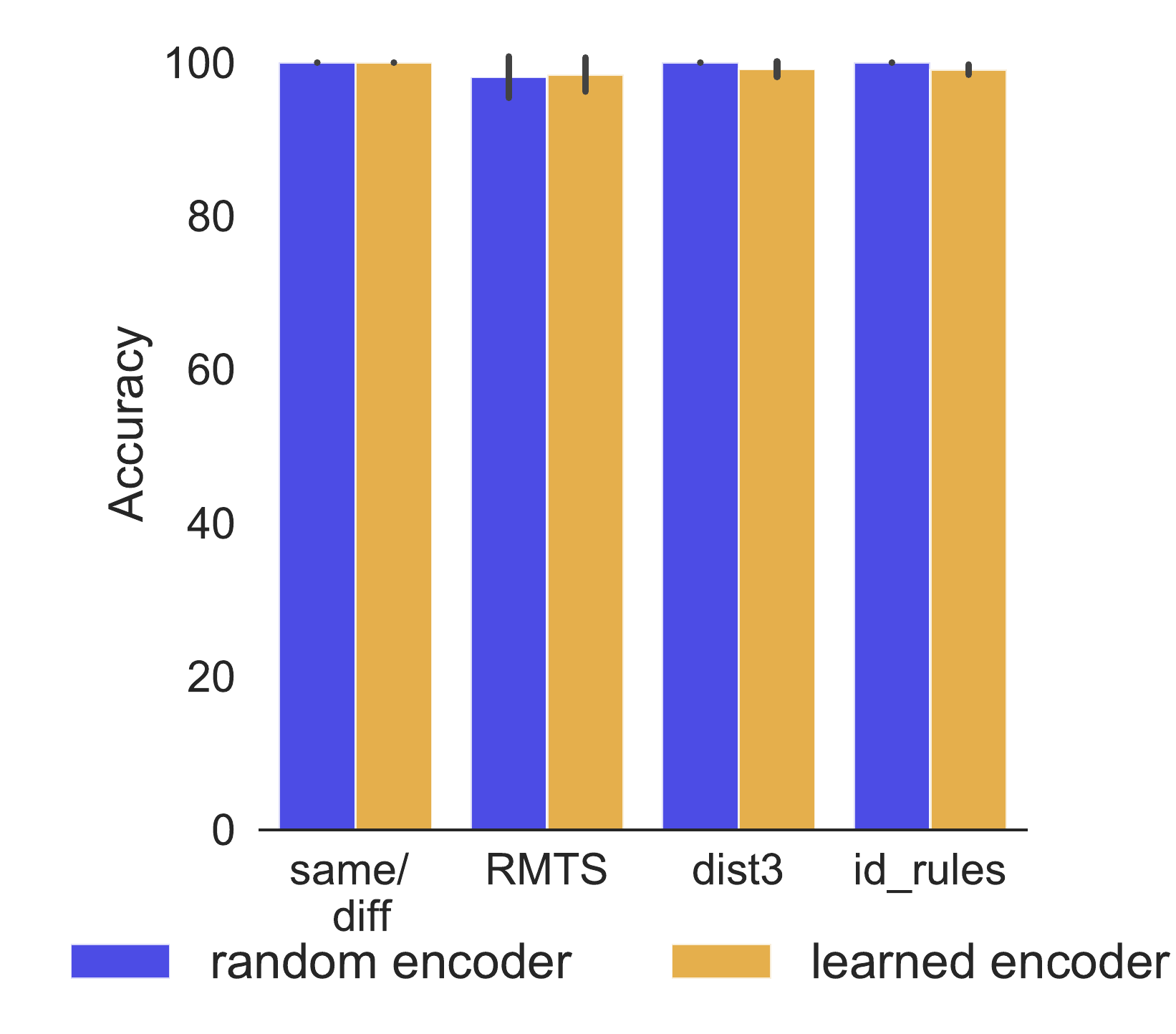}  
\end{minipage}
\caption{\small Test accuracies on the 4 basic relational tasks. Results are displayed for the OOD cases for each task. Results reflect test accuracies averaged over 10 seeds. \textbf{Left.} Results showing average accuracy of relational only models (ESBN, CoRelNet and CoRelNet-T) vs average accuracy of relational + sensory input models (here for all three models (ESBN, CoRelNet and CoRelNet-T) we concatenated the encoded input vectors to the input of the decoder.) For detailed performances see Figure \ref{fig:full_results_concat} in the appendix. \textbf{Middle.} Results showing average accuracy of symmetric models (ESBN, CoRelNet, CoRelNet-T) vs average accuracy of their asymmetric counterparts. For detailed performances see Figure \ref{fig:full_results_symmetry} in the appendix. \textbf{Right.} Results showing average test accuracy of ESBN, CoRelNet and CoRelNet-T with a random encoder vs learned encoder. For the random encoder, the weights of the encoder are randomly initialized, but not updated via backpropagation. For detailed performances see Figure \ref{fig:full_results_freeze} in the appendix.}
\vspace{-4mm}
\label{fig:ablations}
\end{figure}
\looseness=-1
\textbf{All you need is a set of similarity scores.} We first evaluate the hypothesis that the relational information between the objects seen in the input sequence is all we need for OoD generalization on these relational cognitive tasks. 
To this end, we first investigate the importance of disconnection of relational information from the sensory input for OoD generalization.
We hypothesize that if a model learns this function and ignores any information regarding the absolute identity of the objects, then it will be able to generalize well in the OoD settings where the only change performed is the identity of objects considered. This is indeed what we find in experiments, where the models that make predictions solely based on similarity scores between objects (\textit{ESBN}, \textit{CoRelNet} and \textit{CoRelNet-T}) do exceptionally well on OoD generalization, as opposed to models that also rely on the sensory details of the objects (\textit{Transformer}, \textit{LSTM}, \textit{RN}, etc). This is illustrated in Figure \ref{fig:results_main} as well as in the results outlined by \citet{ESBN}. This makes the point that the relational information alone in the form of a $T \times T$ similarity matrix between objects is sufficient to generalize OoD on these tasks, where $T$ is the number of objects in the sequence. 

\begin{figure}[t!]
\begin{subfigure}[c]{0.51\columnwidth}
  \centering
  \includegraphics[width=1.05\textwidth]{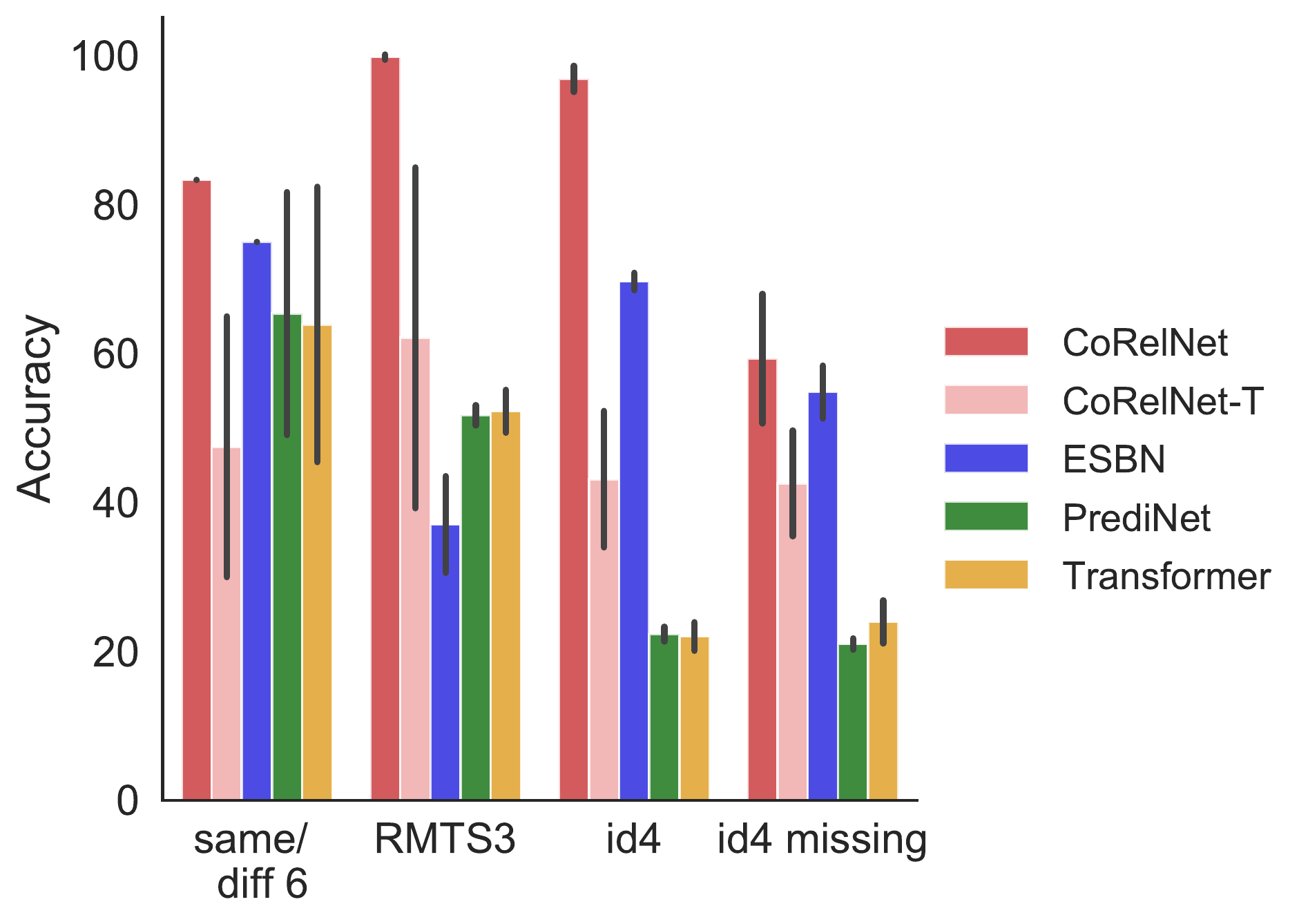}  
\end{subfigure}
\begin{subfigure}[c]{0.47\columnwidth}
  \centering
  \includegraphics[width=0.95\textwidth]{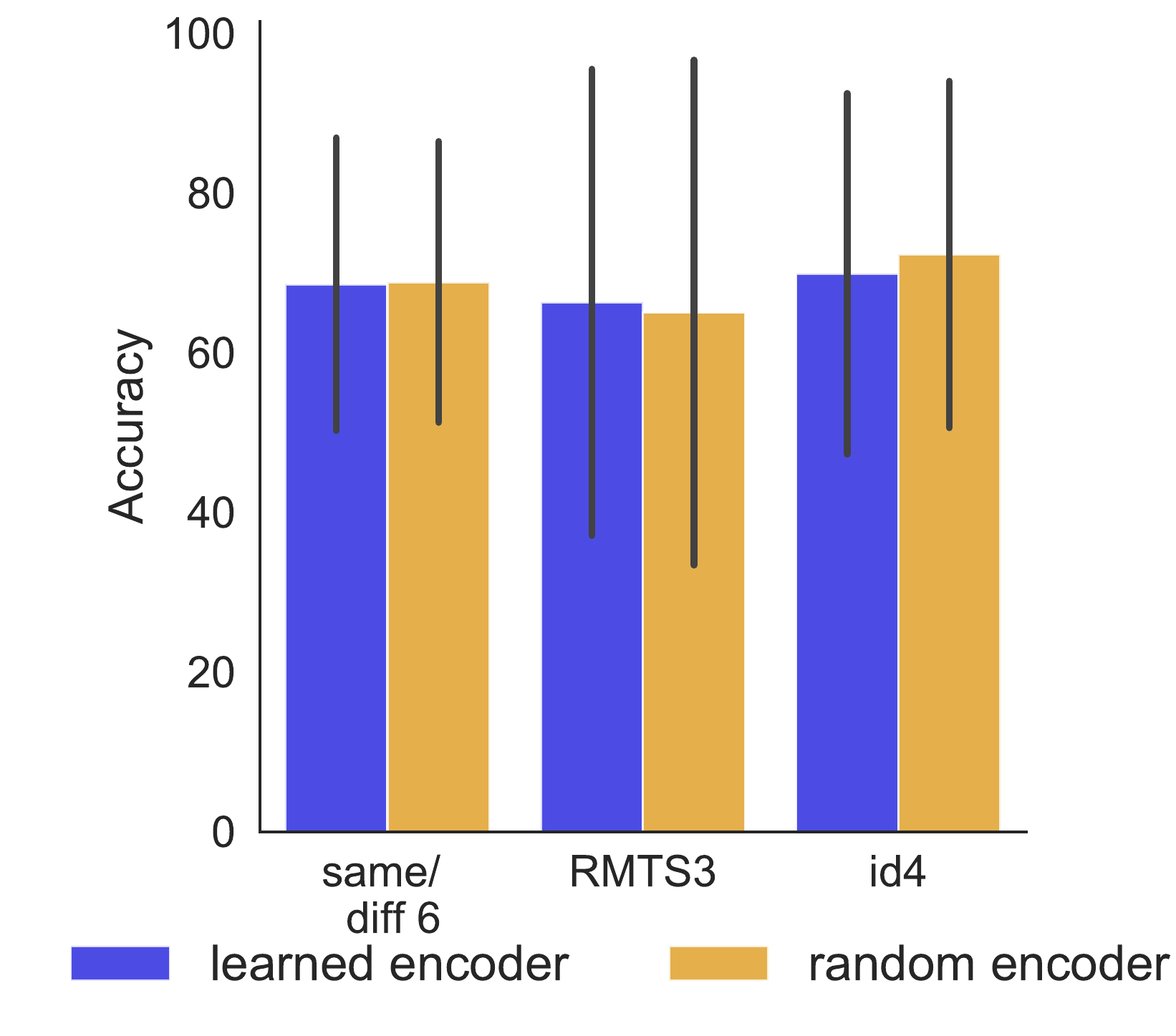}  
\end{subfigure}
\caption{{\small Test accuracies on the relational tasks with unseen relations. Results are for the OOD cases for each task. Results reflect test accuracies averaged over 10 seeds. \textbf{Left.} Overall performance for all 5 models. For full details see Figure \ref{fig:full_results_harder_unseen} in the appendix. \textbf{Right.} Results showing average test accuracy of ESBN, CoRelNet and CoRelNet-T with a random encoder vs learned encoder. For the random encoder, the weights of the encoder are randomly initialized, but not updated via backpropagation. For full details see Figure \ref{fig:full_results_harder_unseen_freeze} in the appendix.}}
\label{fig:results_unseen_main}
\vspace{-4mm}
\end{figure}

Next, we also show experimentally that a a simple MLP decoder, like in \textit{CoRelNet}, outperforms more complex architecture like \textit{ESBN} or \textit{CoRelNet-T} on OoD generalization (also see Figure \ref{fig:results_unseen_main}), assuming the number of objects is fixed. Thus, having stripped away all the additional inductive biases regarding writing keys and recurrence from \textit{ESBN}, we actually see better performance and optimal OoD generalization.

As additional evidence for the hypothesis of this section, we see in Figure \ref{fig:ablations} (left) that the concatenation of sensory information to the already present relational information in input to the classifier stage degrades the OoD generalization capacity of the models. This is because overfitting on the sensory information of objects cannot lead to generalization in the OoD settings, while memorizing the similarity matrix patterns does generalize on unseen new shapes, since the prediction rule as a function of the similarity matrix remains the same in as well as out-of-distribution. 

\textbf{Symmetry.} Having established that learning using the similarity matrix can work well for the tasks at hand, the problem reduces to finding a similarity measure that can decide same-vs-different even for unseen objects. We note that the use of a (symmetric) dot-product leads to an important bias: \textit{it always scores identical objects higher than different objects}. To understand how much the symmetric nature of the dot-product plays a role as a relevant inductive bias, we ablate over an asymmetric version of the dot-product, defined as $(W_1\cdot z_t)^\top (W_2 \cdot z_t)$, where $W_1$ and $W_2$ are learnable matrices, learned via backpropagation. We train \textit{ESBN}, \textit{CoRelNet} and \textit{CoRelNet-T} with symmetric vs asymmetric versions and see that indeed a symmetric dot-product is essential for OoD generalization on these relational tasks (see Figure \ref{fig:ablations} (center)). Note that the symmetric dot-product can be recovered from the asymmetric one but the model is not able to learn it, indicating that it needs to be explicitly encoded as an inductive bias.
\looseness=-1
\textbf{Learning a sensory encoder.} From results presented in the last two paragraphs and Figure \ref{fig:ablations} (left and center), we conclude that if the encoder provides an injective mapping (mapping two distinct objects to two distinct encodings), the symmetric inductive bias of the dot-product allows the model to detect same-vs-different without any encoder training. Given sufficient representation capacity, it becomes hard for a random encoder to lead to a non-injective mapping. This begs the question as to whether the models will maintain their performance even with a randomly initialized and untrained encoder (random encoder)?
Figure \ref{fig:ablations} (right) shows that for the relational cognitive tasks, this is indeed the case, hence, in these tasks, the only part that needs to be trained (besides potentially the parameters of the normalization step) is the decoder (an RNN-based system for ESBN; MLP for CoRelNet, etc.). The job of this decoder is essentially to map patterns in the similarity matrix to final predictions. Note that reducing the data-setting from a sequence of objects with various shapes to a similarity matrix reduces the hypothesis space and incentivizes the decoder to learn the true prediction rule for these set of tasks.

\begin{wrapfigure}{r}{0.5\textwidth}
\vspace{-6mm}
\includegraphics[width=\textwidth]{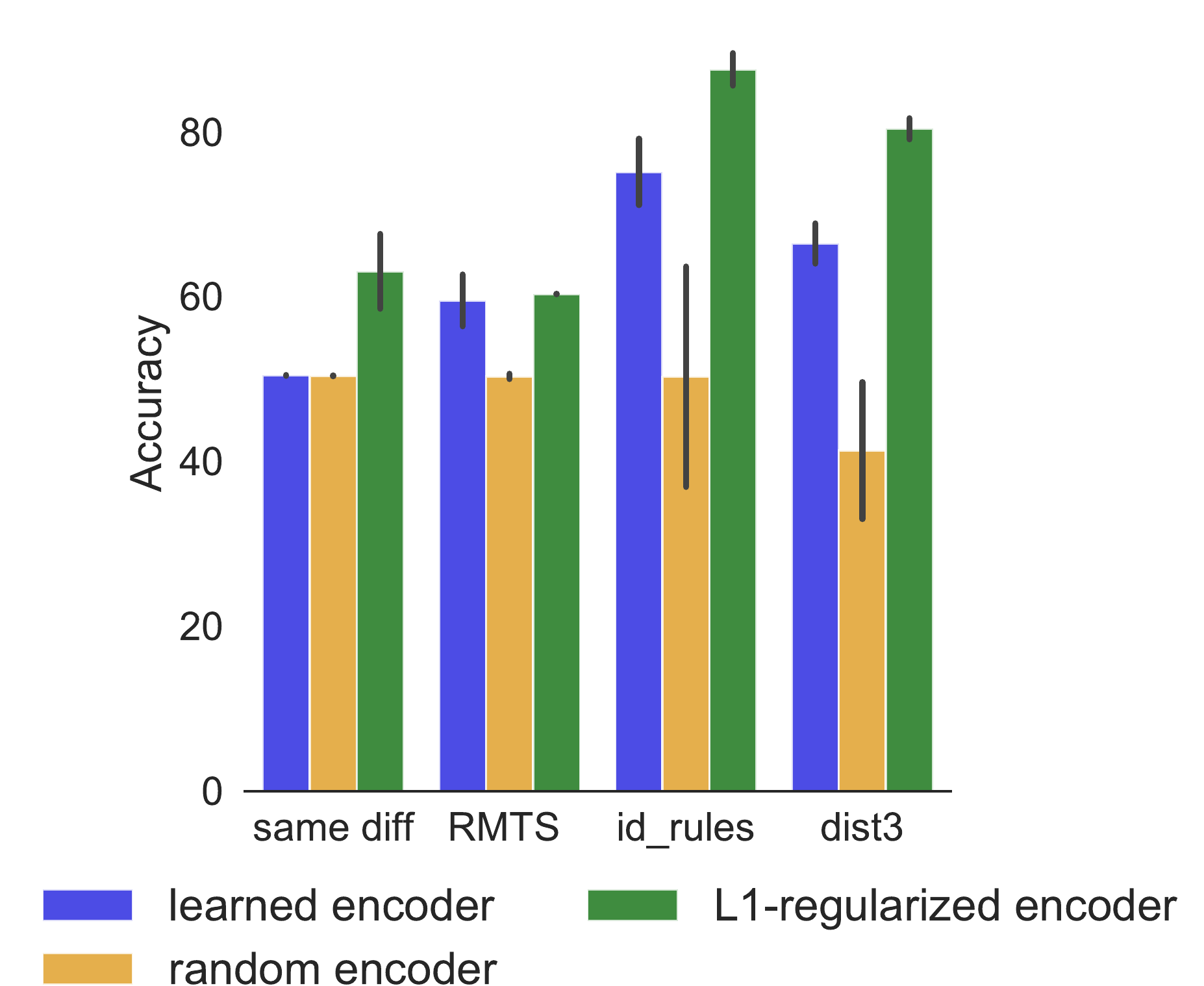}  
\caption{{\small Test accuracies on the relational tasks with spurious features. Results are displayed for the OOD cases, and reflect test accuracy averaged over 10 seeds. The random encoder model class displays the average accuracies of ESBN and \textit{CoRelNet} with random encoder, while the learned encoder model class displays the average accuracies of ESBN and \textit{CoRelNet} with a trained encoder. The L1-regularized include ESBN and \textit{CoRelNet}, both having L1-regularization on learnt representations (with $\lambda=5$ for RMTS and $\lambda=1$ for the other three tasks). For more details see Appendix \ref{app:spurious_details}}.}
\label{fig:results_spurious_main}
\end{wrapfigure}

\looseness=-1
\textbf{Generalization with unseen relations.} Next, we consider a scenario where only a subset of relations are shown during training. We see that \textit{CoRelNet} outperforms ESBN and Transformer on all of these tasks (Figure \ref{fig:results_unseen_main}), with near perfect accuracy on \textit{RMTS 3} and \textit{identity rules 4}, confirming that it is not only able to generalize to unseen objects but also to unseen relations. We hypothesize that the drop in performance of all models in the \textit{identity rules for 4 elements (with missing variations)} task is due to the fact that some of the ``essential base elements" that span the space of all possible abstract relations are left out. We specifically set up this task in order to see a substantial drop in performance. We also note that leaving out either $ABAA$ or $ABAB$ alone is not sufficient for \textit{CoRelNet} to experience a drop in performance, both classes of quadruples need to be removed from the training set. 

\textbf{Generalization with spurious features.} Results in Figure \ref{fig:results_spurious_main} show that random encoder performs poorly (averaged over ESBN and CoRelNet models) in the presence of spurious features. In the relational cognitive tasks, the entire input image can be considered to compute the underlying relational structure, and all the encoder had to accomplish is distinguish two distinct images. In the spurious feature tasks, the encoder not only has to learn what features to ignore on seen shapes but also do so on unseen shapes. Hence a random encoder is not sufficient to accomplish this goal. We also included ESBN and \textit{CoRelNet} with a linear L1-regularization layer after the encoder in order to dampen the effect of spurious features. We see that even a trained encoder struggles on ignoring the color, and often requires strong regularization to do well.

\section{Conclusion}\label{section:conclusion}
We conclude that for purely relational tasks,
the relational information between the objects is all that is needed for good OoD generalization. In particular, for the models we considered, disconnecting the said relational information from the sensory input is an essential inductive bias for OoD generalization on unseen inputs (Figure \ref{fig:ablations} (left)). This is further strengthened by the fact that after removing the effects of encoder-training as well as additional inductive biases in the decoder, we still achieve good OoD generalization as long as final prediction is driven \textit{solely} by relational information.

Further, we see that computing the relational information through a (\textit{symmetric}) dot-product is useful, as it always scores higher on identical objects than on different objects (Figure \ref{fig:ablations} (center)). We also conclude that as long as the encoder is injective and there are no spurious features, even a random encoder can accomplish good OoD generalization not only on unseen objects but often also on unseen relations (Figure \ref{fig:ablations} (right) and Figure \ref{fig:results_unseen_main} (right)). In other words, OoD performance on purely relational tasks without spurious features is a direct reflection of how well the decoder translates the relational information encoded in the similarity matrix to the correct output. It turns out that even a simple MLP decoder does the job. We also note that OoD performance on unseen objects and relations can deteriorate if certain crucial classes of training samples are withheld (refer to Identity rules 4 task with missing examples in Section \ref{sec:harder_unseen_psychophysics_tasks} and Figure \ref{fig:results_unseen_main} (left)). 

In the presence of spurious features, we see that a random encoder no longer does well, as expected since the encoder now needs to selectively encode only a subset of features (also on unseen objects) which are directly relevant for the underlying rule while ignoring the rest. We experimentally show that this is substantially harder as even using a learned encoder is often not enough and requires strong regularizations on the representations (see Figure \ref{fig:results_spurious_main}).

\looseness=-1
Finally, we highlight that the use of network architectures that promote the dichotomy of relational and sensory signals could potentially promote unwanted biases when sub-optimal, as seen in models that were not able to ignore distracting features. The framework and analysis proposed in this work could help better disentangle sensory and relational information to mitigate this potentially problematic impact on society.

\section{Future work}
This work analyses the use of self-attention coefficients or some kind of similarity measure between objects as sole inputs for downstream predictions to implement the inductive bias of separating sensory inputs from relational information. This procedure is easy and flexible to implement in a variety of settings, potentially addressing some limitation of current large scale models that struggle with OoD generalization. We would also like to explore this inductive bias in several architectures with potentially more general similarity measures which not only rely on a notion of "sameness". 

We also see that in some OoD tasks, separating sensory inputs from relational information might be beneficial, but sometimes only by a small margin (Figure \ref{fig:ablations} (left); for instance in the Dist3 task the difference is smallest). Meanwhile, there might be more complex real-world tasks which may require heavier reliance on sensory information. A potential idea would be to explore a flexible task-based competition structure between the use of sensory and relational information, where the use of sensory information for downstream prediction has to undergo a bottleneck due to the competition structure as well as additional task-specific regularization.

Another interesting avenue is to explore settings where multiple abstract rules are at play and need to be individually inferred and combined in a flexible manner, often in the presence of spurious features and out-of-distribution compositions. This is a challenging problem as the encoder has to learn to ignore spurious features for a given rule, which is crucial especially when combined with multiple rules at play. We believe that understanding these settings and distilling a key set of inductive biases that may make artificial systems succeed in these complex domains is an important future work.
\newpage
\begin{ack}
The authors would like to thank Mike Mozer, Matthew Jones, 
Jonathan Cohen and Taylor Webb for insightful discussions. GK acknowledges UNIQUE PhD Excellence Scholarship. SM acknowledges the support of UNIQUE and IVADO Masters Excellence Scholarship towards his research. The authors are grateful to Samsung Electronics Co., Ldt., IVADO, CIFAR Canada AI Chair and Learning in Brains and Machines programs, and the Canada Research Chair Program for support. Authors thank Calcul Qu\'{e}bec and Compute Canada for providing us with the computing resources.
\end{ack}

{
\small
\bibliography{neurips_2022}

\begin{thebibliography}{51}
\providecommand{\natexlab}[1]{#1}
\providecommand{\url}[1]{\texttt{#1}}
\expandafter\ifx\csname urlstyle\endcsname\relax
  \providecommand{\doi}[1]{doi: #1}\else
  \providecommand{\doi}{doi: \begingroup \urlstyle{rm}\Url}\fi

\bibitem[Bahdanau et~al.(2015)Bahdanau, Cho, and Bengio]{Bahdanau2015NeuralMT}
Bahdanau, D., Cho, K., and Bengio, Y.
\newblock Neural machine translation by jointly learning to align and
  translate.
\newblock \emph{CoRR}, abs/1409.0473, 2015.

\bibitem[Barsalou et~al.(2003)Barsalou, {Kyle Simmons}, Barbey, and
  Wilson]{BARSALOU200384}
Barsalou, L.~W., {Kyle Simmons}, W., Barbey, A.~K., and Wilson, C.~D.
\newblock Grounding conceptual knowledge in modality-specific systems.
\newblock \emph{Trends in Cognitive Sciences}, 7\penalty0 (2):\penalty0 84--91,
  2003.
\newblock ISSN 1364-6613.
\newblock \doi{https://doi.org/10.1016/S1364-6613(02)00029-3}.
\newblock URL
  \url{https://www.sciencedirect.com/science/article/pii/S1364661302000293}.

\bibitem[Battaglia et~al.(2018)Battaglia, Hamrick, Bapst, Sanchez-Gonzalez,
  Zambaldi, Malinowski, Tacchetti, Raposo, Santoro, Faulkner,
  et~al.]{battaglia2018relational}
Battaglia, P.~W., Hamrick, J.~B., Bapst, V., Sanchez-Gonzalez, A., Zambaldi,
  V., Malinowski, M., Tacchetti, A., Raposo, D., Santoro, A., Faulkner, R.,
  et~al.
\newblock Relational inductive biases, deep learning, and graph networks.
\newblock \emph{arXiv preprint arXiv:1806.01261}, 2018.

\bibitem[Bengio et~al.(2019)Bengio, Deleu, Rahaman, Ke, Lachapelle, Bilaniuk,
  Goyal, and Pal]{bengio2019meta}
Bengio, Y., Deleu, T., Rahaman, N., Ke, R., Lachapelle, S., Bilaniuk, O.,
  Goyal, A., and Pal, C.
\newblock A meta-transfer objective for learning to disentangle causal
  mechanisms.
\newblock \emph{arXiv preprint arXiv:1901.10912}, 2019.

\bibitem[Brown et~al.(2020)Brown, Mann, Ryder, Subbiah, Kaplan, Dhariwal,
  Neelakantan, Shyam, Sastry, Askell, et~al.]{brown2020language}
Brown, T.~B., Mann, B., Ryder, N., Subbiah, M., Kaplan, J., Dhariwal, P.,
  Neelakantan, A., Shyam, P., Sastry, G., Askell, A., et~al.
\newblock Language models are few-shot learners.
\newblock \emph{arXiv preprint arXiv:2005.14165}, 2020.

\bibitem[Carpenter et~al.(1990)Carpenter, Just, and Shell]{carpenter1990one}
Carpenter, P.~A., Just, M.~A., and Shell, P.
\newblock What one intelligence test measures: a theoretical account of the
  processing in the raven progressive matrices test.
\newblock \emph{Psychological review}, 97\penalty0 (3):\penalty0 404, 1990.

\bibitem[Child et~al.(2019)Child, Gray, Radford, and
  Sutskever]{child2019generating}
Child, R., Gray, S., Radford, A., and Sutskever, I.
\newblock Generating long sequences with sparse transformers.
\newblock \emph{arXiv preprint arXiv:1904.10509}, 2019.

\bibitem[Cohen \& Eichenbaum(1993)Cohen and Eichenbaum]{cohenEichenbaum93}
Cohen, N. and Eichenbaum, H.
\newblock \emph{Memory, amnesia, and the hippocampal system}.
\newblock The MIT Press, 1993.

\bibitem[Dai et~al.(2019)Dai, Yang, Yang, Carbonell, Le, and
  Salakhutdinov]{dai2019transformer}
Dai, Z., Yang, Z., Yang, Y., Carbonell, J., Le, Q.~V., and Salakhutdinov, R.
\newblock Transformer-xl: Attentive language models beyond a fixed-length
  context.
\newblock \emph{arXiv preprint arXiv:1901.02860}, 2019.

\bibitem[Dehghani et~al.(2018)Dehghani, Gouws, Vinyals, Uszkoreit, and
  Kaiser]{universal_tsf}
Dehghani, M., Gouws, S., Vinyals, O., Uszkoreit, J., and Kaiser, L.
\newblock Universal transformers.
\newblock \emph{CoRR}, abs/1807.03819, 2018.
\newblock URL \url{http://arxiv.org/abs/1807.03819}.

\bibitem[Devlin et~al.(2018)Devlin, Chang, Lee, and Toutanova]{devlin2018bert}
Devlin, J., Chang, M.-W., Lee, K., and Toutanova, K.
\newblock Bert: Pre-training of deep bidirectional transformers for language
  understanding.
\newblock \emph{arXiv preprint arXiv:1810.04805}, 2018.

\bibitem[Dosovitskiy et~al.(2020)Dosovitskiy, Beyer, Kolesnikov, Weissenborn,
  Zhai, Unterthiner, Dehghani, Minderer, Heigold, Gelly,
  et~al.]{dosovitskiy2020image}
Dosovitskiy, A., Beyer, L., Kolesnikov, A., Weissenborn, D., Zhai, X.,
  Unterthiner, T., Dehghani, M., Minderer, M., Heigold, G., Gelly, S., et~al.
\newblock An image is worth 16x16 words: Transformers for image recognition at
  scale.
\newblock \emph{arXiv preprint arXiv:2010.11929}, 2020.

\bibitem[Fortunato et~al.(2019)Fortunato, Tan, Faulkner, Hansen, Badia,
  Buttimore, Deck, Leibo, and Blundell]{fortunato2019generalization}
Fortunato, M., Tan, M., Faulkner, R., Hansen, S., Badia, A.~P., Buttimore, G.,
  Deck, C., Leibo, J.~Z., and Blundell, C.
\newblock Generalization of reinforcement learners with working and episodic
  memory.
\newblock In \emph{Advances in Neural Information Processing Systems}, pp.\
  12469--12478, 2019.

\bibitem[Goldstone et~al.(1989)Goldstone, Gentner, and
  Medin]{Goldstone89relationsrelating}
Goldstone, R.~L., Gentner, D., and Medin, D.~L.
\newblock Relations relating relations.
\newblock In \emph{In Proceedings (tf the Eleventh Annual Conference of the
  Cognitive Science Society (pp. 131 - 138}, 1989.

\bibitem[Goodale \& Milner(1992)Goodale and Milner]{goodale1992separate}
Goodale, M.~A. and Milner, A.~D.
\newblock Separate visual pathways for perception and action.
\newblock \emph{Trends in neurosciences}, 15\penalty0 (1):\penalty0 20--25,
  1992.

\bibitem[Graves et~al.(2014)Graves, Wayne, and Danihelka]{graves2014neural}
Graves, A., Wayne, G., and Danihelka, I.
\newblock Neural turing machines.
\newblock \emph{arXiv preprint arXiv:1410.5401}, 2014.

\bibitem[He et~al.(2015)He, Zhang, Ren, and Sun]{he2015delving}
He, K., Zhang, X., Ren, S., and Sun, J.
\newblock Delving deep into rectifiers: Surpassing human-level performance on
  imagenet classification.
\newblock In \emph{Proceedings of the IEEE international conference on computer
  vision}, pp.\  1026--1034, 2015.

\bibitem[Hill et~al.(2020)Hill, Tieleman, von Glehn, Wong, Merzic, and
  Clark]{hill2020grounded}
Hill, F., Tieleman, O., von Glehn, T., Wong, N., Merzic, H., and Clark, S.
\newblock Grounded language learning fast and slow.
\newblock \emph{arXiv preprint arXiv:2009.01719}, 2020.

\bibitem[Hudson \& Manning(2018)Hudson and Manning]{hudson2018compositional}
Hudson, D.~A. and Manning, C.~D.
\newblock Compositional attention networks for machine reasoning.
\newblock \emph{arXiv preprint arXiv:1803.03067}, 2018.

\bibitem[Johnson et~al.(2017)Johnson, Hariharan, Van Der~Maaten, Fei-Fei,
  Lawrence~Zitnick, and Girshick]{johnson2017clevr}
Johnson, J., Hariharan, B., Van Der~Maaten, L., Fei-Fei, L., Lawrence~Zitnick,
  C., and Girshick, R.
\newblock Clevr: A diagnostic dataset for compositional language and elementary
  visual reasoning.
\newblock In \emph{Proceedings of the IEEE conference on computer vision and
  pattern recognition}, pp.\  2901--2910, 2017.

\bibitem[Jones \& Love(2007)Jones and Love]{JONES2007196}
Jones, M. and Love, B.~C.
\newblock Beyond common features: The role of roles in determining similarity.
\newblock \emph{Cognitive Psychology}, 55\penalty0 (3):\penalty0 196--231,
  2007.
\newblock ISSN 0010-0285.
\newblock \doi{https://doi.org/10.1016/j.cogpsych.2006.09.004}.
\newblock URL
  \url{https://www.sciencedirect.com/science/article/pii/S0010028506000715}.

\bibitem[Ke et~al.(2021)Ke, Didolkar, Mittal, Goyal, Lajoie, Bauer, Rezende,
  Mozer, Bengio, and Pal]{ke2021systematic}
Ke, N.~R., Didolkar, A.~R., Mittal, S., Goyal, A., Lajoie, G., Bauer, S.,
  Rezende, D.~J., Mozer, M.~C., Bengio, Y., and Pal, C.
\newblock Systematic evaluation of causal discovery in visual model based
  reinforcement learning.
\newblock \emph{arXiv preprint arXiv:2107.00848}, 2021.

\bibitem[Kim \& Linzen(2020)Kim and Linzen]{kim2020cogs}
Kim, N. and Linzen, T.
\newblock Cogs: A compositional generalization challenge based on semantic
  interpretation.
\newblock \emph{arXiv preprint arXiv:2010.05465}, 2020.

\bibitem[Kim~J(2018)]{not-so-clevr}
Kim~J, Ricci~M, S.~T.
\newblock Not-so-clevr: learning same–different relations strains feedforward
  neural networks.
\newblock \emph{Royal Society}, 2018.
\newblock URL \url{https://doi.org/10.1098/rsfs.2018.0011}.

\bibitem[Kipf et~al.(2018)Kipf, Fetaya, Wang, Welling, and
  Zemel]{kipf2018neural}
Kipf, T., Fetaya, E., Wang, K.-C., Welling, M., and Zemel, R.
\newblock Neural relational inference for interacting systems.
\newblock \emph{arXiv preprint arXiv:1802.04687}, 2018.

\bibitem[Kriete et~al.(2013)Kriete, Noelle, Cohen, and
  O{\textquoteright}Reilly]{Kriete16390}
Kriete, T., Noelle, D.~C., Cohen, J.~D., and O{\textquoteright}Reilly, R.~C.
\newblock Indirection and symbol-like processing in the prefrontal cortex and
  basal ganglia.
\newblock \emph{Proceedings of the National Academy of Sciences}, 110\penalty0
  (41):\penalty0 16390--16395, 2013.
\newblock ISSN 0027-8424.
\newblock \doi{10.1073/pnas.1303547110}.
\newblock URL \url{https://www.pnas.org/content/110/41/16390}.

\bibitem[Locatello et~al.(2020)Locatello, Weissenborn, Unterthiner, Mahendran,
  Heigold, Uszkoreit, Dosovitskiy, and Kipf]{locatello2020object}
Locatello, F., Weissenborn, D., Unterthiner, T., Mahendran, A., Heigold, G.,
  Uszkoreit, J., Dosovitskiy, A., and Kipf, T.
\newblock Object-centric learning with slot attention.
\newblock \emph{arXiv preprint arXiv:2006.15055}, 2020.

\bibitem[Luong et~al.(2015)Luong, Pham, and Manning]{luong2015effective}
Luong, M.-T., Pham, H., and Manning, C.~D.
\newblock Effective approaches to attention-based neural machine translation.
\newblock \emph{arXiv preprint arXiv:1508.04025}, 2015.

\bibitem[Marcus et~al.(1999)Marcus, Vijayan, Rao, and Vishton]{marcus1999rule}
Marcus, G.~F., Vijayan, S., Rao, S.~B., and Vishton, P.~M.
\newblock Rule learning by seven-month-old infants.
\newblock \emph{Science}, 283\penalty0 (5398):\penalty0 77--80, 1999.

\bibitem[Mittal et~al.(2020)Mittal, Lamb, Goyal, Voleti, Shanahan, Lajoie,
  Mozer, and Bengio]{mittal2020learning}
Mittal, S., Lamb, A., Goyal, A., Voleti, V., Shanahan, M., Lajoie, G., Mozer,
  M., and Bengio, Y.
\newblock Learning to combine top-down and bottom-up signals in recurrent
  neural networks with attention over modules.
\newblock In \emph{International Conference on Machine Learning}, pp.\
  6972--6986. PMLR, 2020.

\bibitem[Mittal et~al.(2021)Mittal, Raparthy, Rish, Bengio, and
  Lajoie]{mittal2021compositional}
Mittal, S., Raparthy, S.~C., Rish, I., Bengio, Y., and Lajoie, G.
\newblock Compositional attention: Disentangling search and retrieval.
\newblock \emph{arXiv preprint arXiv:2110.09419}, 2021.

\bibitem[Newman et~al.(2020)Newman, Hewitt, Liang, and Manning]{newman2020eos}
Newman, B., Hewitt, J., Liang, P., and Manning, C.~D.
\newblock The eos decision and length extrapolation.
\newblock \emph{arXiv preprint arXiv:2010.07174}, 2020.

\bibitem[Nogueira et~al.(2021)Nogueira, Jiang, and
  Lin]{nogueira2021investigating}
Nogueira, R., Jiang, Z., and Lin, J.
\newblock Investigating the limitations of transformers with simple arithmetic
  tasks.
\newblock \emph{arXiv preprint arXiv:2102.13019}, 2021.

\bibitem[Oord et~al.(2016)Oord, Dieleman, Zen, Simonyan, Vinyals, Graves,
  Kalchbrenner, Senior, and Kavukcuoglu]{oord2016wavenet}
Oord, A. v.~d., Dieleman, S., Zen, H., Simonyan, K., Vinyals, O., Graves, A.,
  Kalchbrenner, N., Senior, A., and Kavukcuoglu, K.
\newblock Wavenet: A generative model for raw audio.
\newblock \emph{arXiv preprint arXiv:1609.03499}, 2016.

\bibitem[Parmar et~al.(2018)Parmar, Vaswani, Uszkoreit, Kaiser, Shazeer, and
  Ku]{image-transformer}
Parmar, N., Vaswani, A., Uszkoreit, J., Kaiser, L., Shazeer, N., and Ku, A.
\newblock Image transformer.
\newblock \emph{International Conference on Machine Learning}, 2018.
\newblock URL \url{http://arxiv.org/abs/1802.05751}.

\bibitem[Pratap et~al.(2019)Pratap, Hannun, Xu, Cai, Kahn, Synnaeve,
  Liptchinsky, and Collobert]{pratap2019wav2letter++}
Pratap, V., Hannun, A., Xu, Q., Cai, J., Kahn, J., Synnaeve, G., Liptchinsky,
  V., and Collobert, R.
\newblock Wav2letter++: A fast open-source speech recognition system.
\newblock In \emph{ICASSP 2019-2019 IEEE International Conference on Acoustics,
  Speech and Signal Processing (ICASSP)}, pp.\  6460--6464. IEEE, 2019.

\bibitem[Pritzel et~al.(2017)Pritzel, Uria, Srinivasan, Puigdomenech, Vinyals,
  Hassabis, Wierstra, and Blundell]{pritzel2017neural}
Pritzel, A., Uria, B., Srinivasan, S., Puigdomenech, A., Vinyals, O., Hassabis,
  D., Wierstra, D., and Blundell, C.
\newblock Neural episodic control.
\newblock \emph{arXiv preprint arXiv:1703.01988}, 2017.

\bibitem[Raven \& Court(1938)Raven and Court]{raven1938raven}
Raven, J.~C. and Court, J.
\newblock \emph{Raven's progressive matrices}.
\newblock Western Psychological Services Los Angeles, CA, 1938.

\bibitem[Santoro et~al.(2017)Santoro, Raposo, Barrett, Malinowski, Pascanu,
  Battaglia, and Lillicrap]{santoro2017simple}
Santoro, A., Raposo, D., Barrett, D.~G., Malinowski, M., Pascanu, R.,
  Battaglia, P., and Lillicrap, T.
\newblock A simple neural network module for relational reasoning.
\newblock In \emph{Advances in neural information processing systems}, pp.\
  4967--4976, 2017.

\bibitem[Santoro et~al.(2018)Santoro, Faulkner, Raposo, Rae, Chrzanowski,
  Weber, Wierstra, Vinyals, Pascanu, and Lillicrap]{santoro2018relational}
Santoro, A., Faulkner, R., Raposo, D., Rae, J., Chrzanowski, M., Weber, T.,
  Wierstra, D., Vinyals, O., Pascanu, R., and Lillicrap, T.
\newblock Relational recurrent neural networks.
\newblock \emph{arXiv preprint arXiv:1806.01822}, 2018.

\bibitem[Scarselli et~al.(2008)Scarselli, Gori, Tsoi, Hagenbuchner, and
  Monfardini]{scarselli2008graph}
Scarselli, F., Gori, M., Tsoi, A.~C., Hagenbuchner, M., and Monfardini, G.
\newblock The graph neural network model.
\newblock \emph{IEEE transactions on neural networks}, 20\penalty0
  (1):\penalty0 61--80, 2008.

\bibitem[Sedda \& Scarpina(2012)Sedda and Scarpina]{sedda2012dorsal}
Sedda, A. and Scarpina, F.
\newblock Dorsal and ventral streams across sensory modalities.
\newblock \emph{Neuroscience bulletin}, 28\penalty0 (3):\penalty0 291--300,
  2012.

\bibitem[Shanahan et~al.(2019)Shanahan, Nikiforou, Creswell, Kaplanis, Barrett,
  and Garnelo]{shanahan2019explicitly}
Shanahan, M., Nikiforou, K., Creswell, A., Kaplanis, C., Barrett, D., and
  Garnelo, M.
\newblock An explicitly relational neural network architecture.
\newblock \emph{arXiv preprint arXiv:1905.10307}, 2019.

\bibitem[Tolman(1948)]{tolman48}
Tolman, E.
\newblock Cognitive maps in rats and men.
\newblock \emph{Psychological Review}, 55(4):\penalty0 189--208, 1948.

\bibitem[Vaswani et~al.(2017)Vaswani, Shazeer, Parmar, Uszkoreit, Jones, Gomez,
  Kaiser, and Polosukhin]{vaswani2017attention}
Vaswani, A., Shazeer, N., Parmar, N., Uszkoreit, J., Jones, L., Gomez, A.~N.,
  Kaiser, {\L}., and Polosukhin, I.
\newblock Attention is all you need.
\newblock In \emph{Advances in neural information processing systems}, pp.\
  5998--6008, 2017.

\bibitem[Veli{\v{c}}kovi{\'c} et~al.(2017)Veli{\v{c}}kovi{\'c}, Cucurull,
  Casanova, Romero, Lio, and Bengio]{velivckovic2017graph}
Veli{\v{c}}kovi{\'c}, P., Cucurull, G., Casanova, A., Romero, A., Lio, P., and
  Bengio, Y.
\newblock Graph attention networks.
\newblock \emph{arXiv preprint arXiv:1710.10903}, 2017.

\bibitem[Webb et~al.(2020)Webb, Dulberg, Frankland, Petrov, O'Reilly, and
  Cohen]{TCN}
Webb, T.~W., Dulberg, Z., Frankland, S.~M., Petrov, A.~A., O'Reilly, R.~C., and
  Cohen, J.~D.
\newblock Learning representations that support extrapolation.
\newblock \emph{Proceedings of the 37th International Conference on Machine
  Learning}, 2020.
\newblock URL \url{https://arxiv.org/abs/2007.05059}.

\bibitem[Webb et~al.(2021)Webb, Sinha, and Cohen]{ESBN}
Webb, T.~W., Sinha, I., and Cohen, J.~D.
\newblock Emergent symbols through binding in external memory.
\newblock \emph{The International Conference on Learning Representations
  (ICLR)}, 2021.
\newblock URL \url{https://arxiv.org/abs/2012.14601}.

\bibitem[Whittington et~al.(2019)Whittington, Muller, Mark, Chen, Barry,
  Burgess, and Behrens]{whittington2019tolman}
Whittington, J.~C., Muller, T.~H., Mark, S., Chen, G., Barry, C., Burgess, N.,
  and Behrens, T.~E.
\newblock The tolman-eichenbaum machine: Unifying space and relational memory
  through generalisation in the hippocampal formation.
\newblock \emph{BioRxiv}, pp.\  770495, 2019.

\bibitem[Yi et~al.(2019)Yi, Gan, Li, Kohli, Wu, Torralba, and
  Tenenbaum]{yi2019clevrer}
Yi, K., Gan, C., Li, Y., Kohli, P., Wu, J., Torralba, A., and Tenenbaum, J.~B.
\newblock Clevrer: Collision events for video representation and reasoning.
\newblock \emph{arXiv preprint arXiv:1910.01442}, 2019.

\bibitem[Zhang et~al.(2019)Zhang, Gao, Jia, Zhu, and Zhu]{zhang2019raven}
Zhang, C., Gao, F., Jia, B., Zhu, Y., and Zhu, S.-C.
\newblock Raven: A dataset for relational and analogical visual reasoning.
\newblock In \emph{Proceedings of the IEEE Conference on Computer Vision and
  Pattern Recognition}, pp.\  5317--5327, 2019.

\end{thebibliography}
\bibliographystyle{neurips_2022}
}

\clearpage
\appendix

\section{Hyperparameters}\label{app:hyperparameters}

Table~\ref{table:parameters} shows the default hyperparameters used for the experiments reported in the main text. All other hyperparamaters, including those of all other models are exactly the same as in \cite{ESBN}.

\begin{table}[ht]
  \caption{Default hyperparameters}
  \label{table:parameters}
  \centering
  \begin{tabular}{ll}
    \toprule
    Parameter                              & Value \\
    \toprule
    Input images size for relational games tasks                     & $36 \times 36 \times 3$ \\
    Iterations for relational games tasks                     & $2500$ \\
    Input images size for relational cognitive tasks                     & $3
    2\times 32 \times 3$ \\
    Iterations for relational cognitive tasks                    & $5000$ \\
    Runs per experiment                    & $10$ \\
    Optimiser                              & Gradient descent \\
    Learning rate                          & $5e-4$ \\
    \midrule
    Encoder output dimension              & $128$ \\
    Encoder non-linearity              & ReLU \\
    Encoder CNN stride              & $2$ \\
    Encoder CNN padding              & $1$ \\
    Encoder CNN first layer input channels              & $3$ \\
    Encoder CNN output channels              & $32$ \\
    Encoder CNN kernel size              & $4$ \\
    Encoder nb. conv layers              & $3$ \\
    Encoder fully connected layer hidden dim              & $256$ \\
    \midrule
    CoRelNet decoder              & fully connected layer with $1$ hidden layer \\
    CoRelNet decoder hidden layer dim             & $256$ \\
    CoRelNet normalization type         & contextnorm \\
    \midrule
    CoRelNet-T heads              & $8$ \\
    CoRelNet-T transformer dim              & $512$ \\
    CoRelNet-T query dim              & $8$ \\
    CoRelNet-T positional encoding dim              & $8$ \\
    CoRelNet-T transformer nb. layers          & $1$ \\
    CoRelNet-T normalization type         & contextnorm \\
    \bottomrule
  \end{tabular}
\end{table}

\section{Relational Games - additional results}\label{app:relational_games}
Table \ref{table:predinet_basic_results_other} displays additional baselines.
\begin{table}[ht!]
\renewcommand{\arraystretch}{1.}
\centering
\resizebox{\columnwidth}{!}{
\begin{tabular}{ c  c | c c | c c c c} 
\toprule
\textit{Task} & \textit{Test Set} & \textit{CoRelNet} & \textit{CoRelNet-T} & \textit{MNM} & \textit{LSTM} & \textit{NTM} & \textit{RN}\\
\midrule
\multirow{2}{4em}{same} & Hex. & \g{94.7}{5.4} & \highlight{\g{98.7}{0.5}} & \g{97.3}{1.5} & \g{96.4}{1.6} & \g{97.2}{1.7} & \g{92.9}{2.8}\\ 
& Str. & \g{90.4}{9.0} & \highlight{\g{98.4}{0.6}} & \g{95.5}{2.0} & \g{90.3}{6.4} &  \g{93.4}{5.3} &  \g{83.7}{10.2} \\ 
\midrule
\multirow{2}{4em}{between} & Hex. & \highlight{\g{96.6}{2.2}} & \g{91.9}{3.3} & \g{93.6}{2.3} & \g{94.9}{3.3} & \g{96.3 }{0.7} & \g{73.6}{8.2}\\ 
& Str. & \g{93.1}{5.4} & \g{87.0}{4.7} & \g{90.2}{5.2} & \g{91.7}{4.5} & \highlight{\g{93.5}{2.5}} & \g{54.5}{5.7}\\ 
\midrule
\multirow{2}{4em}{occurs} & Hex. & \highlight{\g{96.2}{2.2}} & \g{91.6}{4.6} & \g{84.9}{2.8} & \g{92.2}{2.4} & \g{93.5}{8.1} & \g{71.2}{6.6}\\ 
& Str. & \highlight{\g{88.7}{5.3}} & \g{79.3}{12.1} & \g{77.0}{4.5} & \g{79.8}{10.0}  &  \g{87.1}{11.0} & \g{54.6}{4.0}\\ 
\midrule
\multirow{2}{4em}{xoccurs} & Hex. & \highlight{\g{92.2}{6.4}} & \g{91.7}{6.9} & \g{73.8}{6.6} & \g{79.1}{2.8} & \g{84.2}{5.0} & \g{65.7}{5.5}\\ 
& Str. & \g{83.6}{10.9} & \highlight{\g{85.4}{6.4}} & \g{70.6}{6.2} & \g{77.2}{5.9} & \g{81.7}{5.8} & \g{61.7}{7.4}\\ 
\midrule
\multirow{2}{4em}{row matching} & Hex. & \highlight{\g{97.7}{0.8}} & \g{95.4}{5.1} & \g{49.9}{0.3} & \g{50.1}{0.6} & \g{50.2}{0.5} & \g{50.5}{0.3} \\ 
& Str. & \highlight{\g{94.8}{1.3}} & \g{90.5}{5.2} & \g{49.8}{0.6} & \g{50.2}{0.5} &  \g{49.9}{0.5} &  \g{50.4}{0.4} \\ 
\midrule
col./shape & Hex. & \g{47.2}{3.7} & \g{49.6}{0.8} & \g{75.7}{12.1} & \g{80.1}{7.0} &  \highlight{\g{88.0}{2.4}} &  \g{77.5}{9.2} \\  
\midrule
left-of & Hex. & \highlight{\g{99.2}{0.7}} & \g{97.6}{1.2} & \g{97.5}{1.2} & \g{97.2}{1.2} & \g{97.5}{0.9} & \g{51.0}{2.1}\\ 
\bottomrule
\end{tabular}
}
\caption{{\small Out-Of-Distribution test accuracies for the Relational Game tasks on the two held-out sets "Hexominoes" (Hex.) and "Stripes" (Str.). Results reflect test accuracies averaged over 10 seeds.}}
    \label{table:predinet_basic_results_other}  
\end{table}

\section{Identity rules 4 (flipped version) results}\label{app: idrules4_flipped}
\begin{figure}[t!]
    \centering
    \includegraphics[width=0.5\columnwidth]{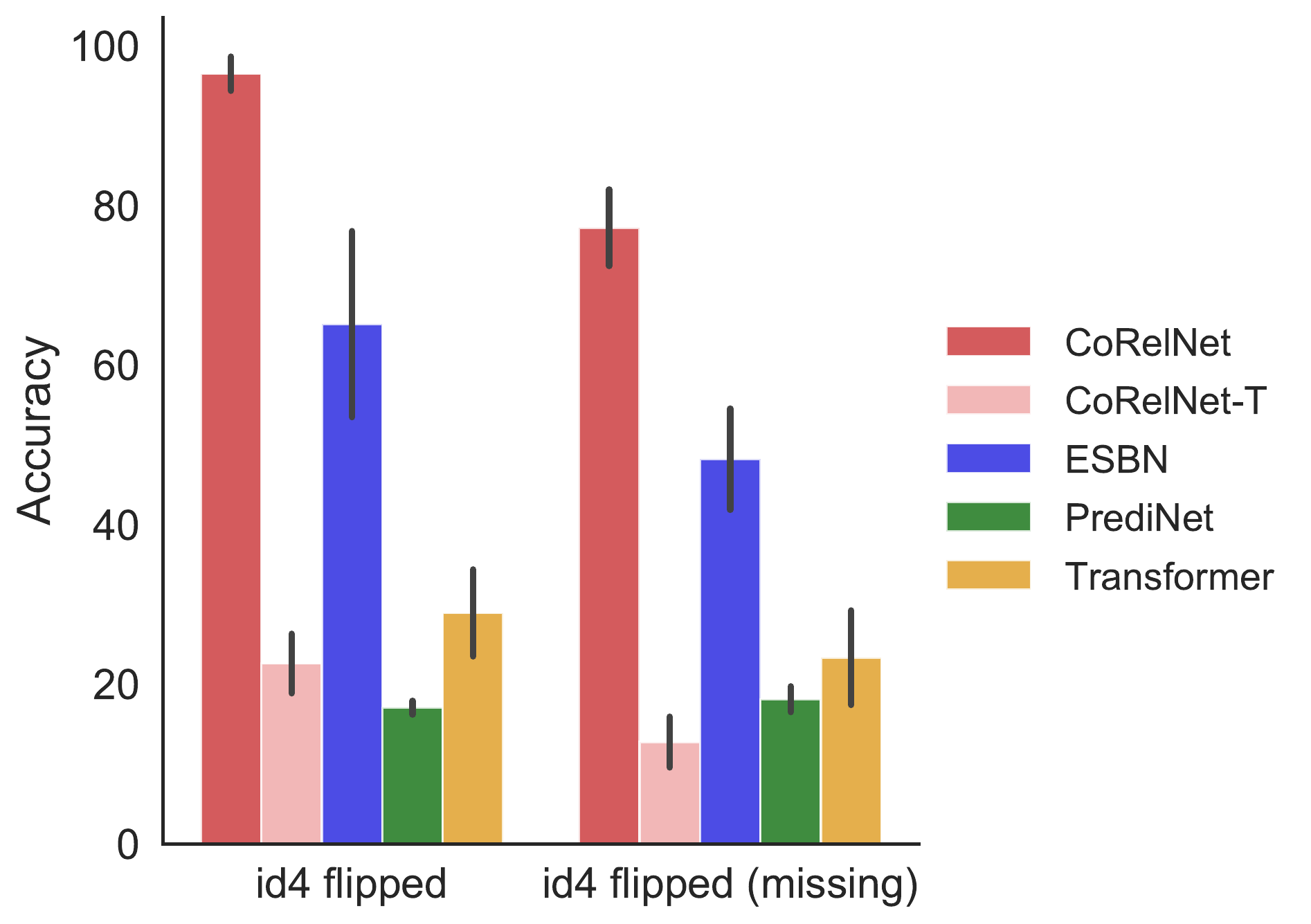}
    \caption{Results for \textit{identity rules 4 [flipped]} and \textit{identity rules 4 [flipped] (with missing variations)} for $m = 94$, which is the most extreme OOD case. Here $m$ denotes the number of shapes not seen during training. There are a total of 100 shapes, and $m=94$ here means $6$ shapes are shown during training, and testing involves only the $94$ unseen shapes. The test result accuracies are averaged over $10$ random seeds.}
    \label{fig:id_rules4_flipped}
\end{figure}

In the section, we briefly outline a version of the \textit{identity rules 4} task, where training and test set have been swapped, which we call \textit{identity rules 4 [flipped]}. In this task, we only train on quadruples with 3 distinct images (only including quadruples of the form $ABCA$, $BACA$ and $BCAA$), while testing only on quadruples with at most two distinct images. We also include a version of this task, with missing variations, where we excluded all quadruples of the form $ABCA$ from the training set. For results, see Figure \ref{fig:id_rules4_flipped}. 
\section{Distribution-of-N results}\label{app: distN_results}

\begin{figure}[t!]
    \centering
    \includegraphics[width=0.7\columnwidth]{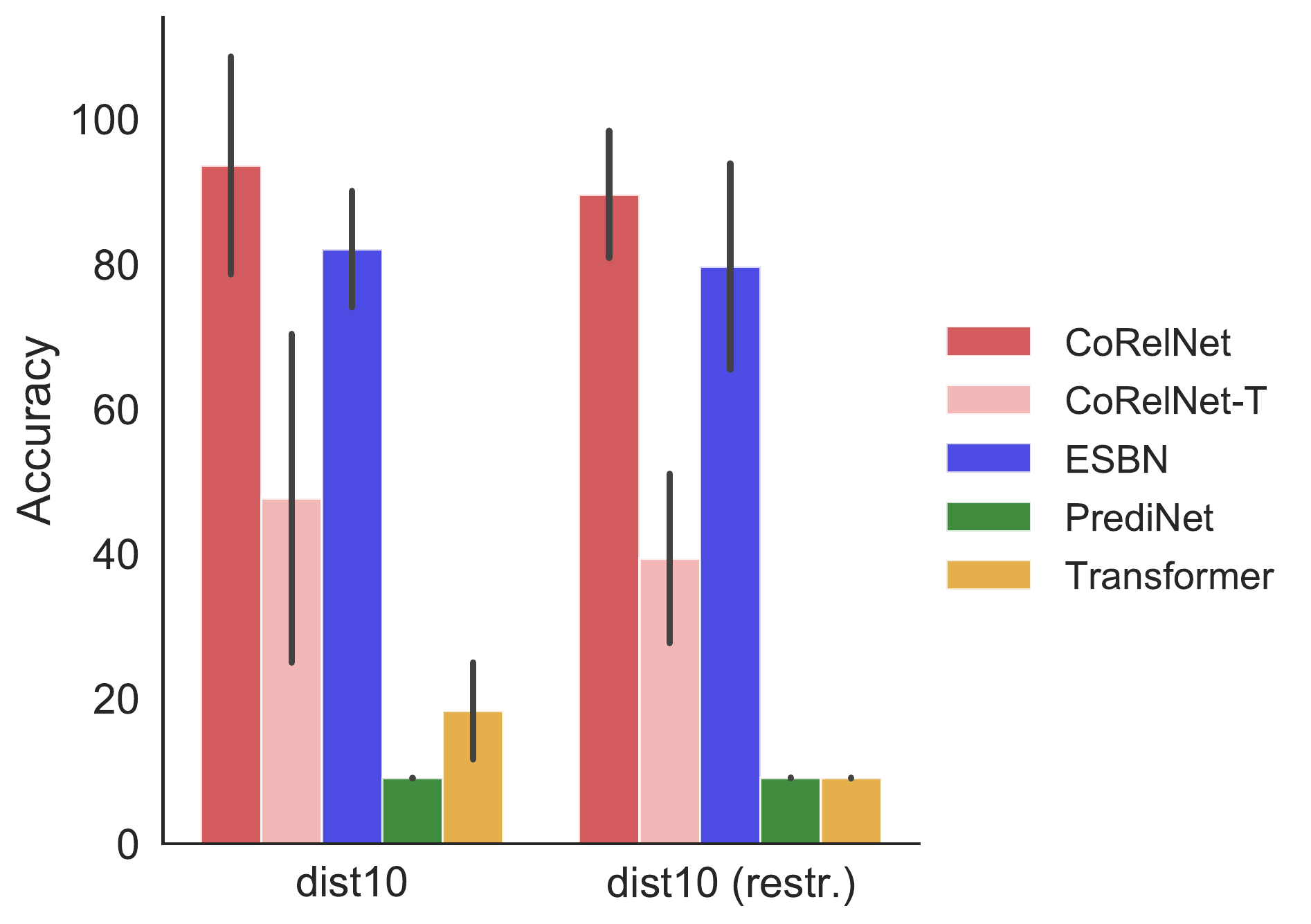}
    \caption{The test result accuracies for \textit{distribution-of-10} and \textit{distribution-of-10} (restricted - with unseen permutations during testing) for the most extreme OOD case ($m=89$). Here $m$ denotes the number of shapes not seen during training. There are a total of 100 shapes, and $m=89$ here means $11$ shapes are shown during training, and testing involves only the $89$ unseen shapes. The test result accuracies are averaged over $10$ random seeds.}
    \label{fig:distN}
\end{figure}

In this section we are going to look at extensions of the \textit{distribution-of-three} task: we will be looking at distributions of $10$ elements, as well as a variation where the set of permutations shown in the training set is restricted and the test set is constructed of permutations not shown during training. We recall that in \textit{distribution-of-three}, the second row in Figure \ref{fig:task_figures} (top left) is a permutation of the first row. The last element of the second row is "removed", and needs to be inferred from the options in the last row, which contains all images shown in the first row and one additional element. In \textit{distribution-of-$N$}, we imagine $N$ distinct images in the first row, and a permutation thereof in the second row. Again, the last element of the second row is "removed" and needs to be inferred from the $N+1$ (the $N$ images shown in the first row and one additional element) options shown in the last row. For the variations of \textit{distribution-of-$N$}, for which we want to only include unseen relations in the test set, we are going to restrict the training set to only contain permutations (sending the first row to the second row) preserving the set of the first $N-3$ elements, while all the other remaining permutations are used to construct the test set. Hence test and training set have disjoint sets of underlying permutations, and thus relations. For results see Figure \ref{fig:distN}. 

\section{Details on relational tasks with spurious features}\label{app:spurious_details}

\begin{figure}[t!]
    \centering
    \includegraphics[width=0.7\columnwidth]{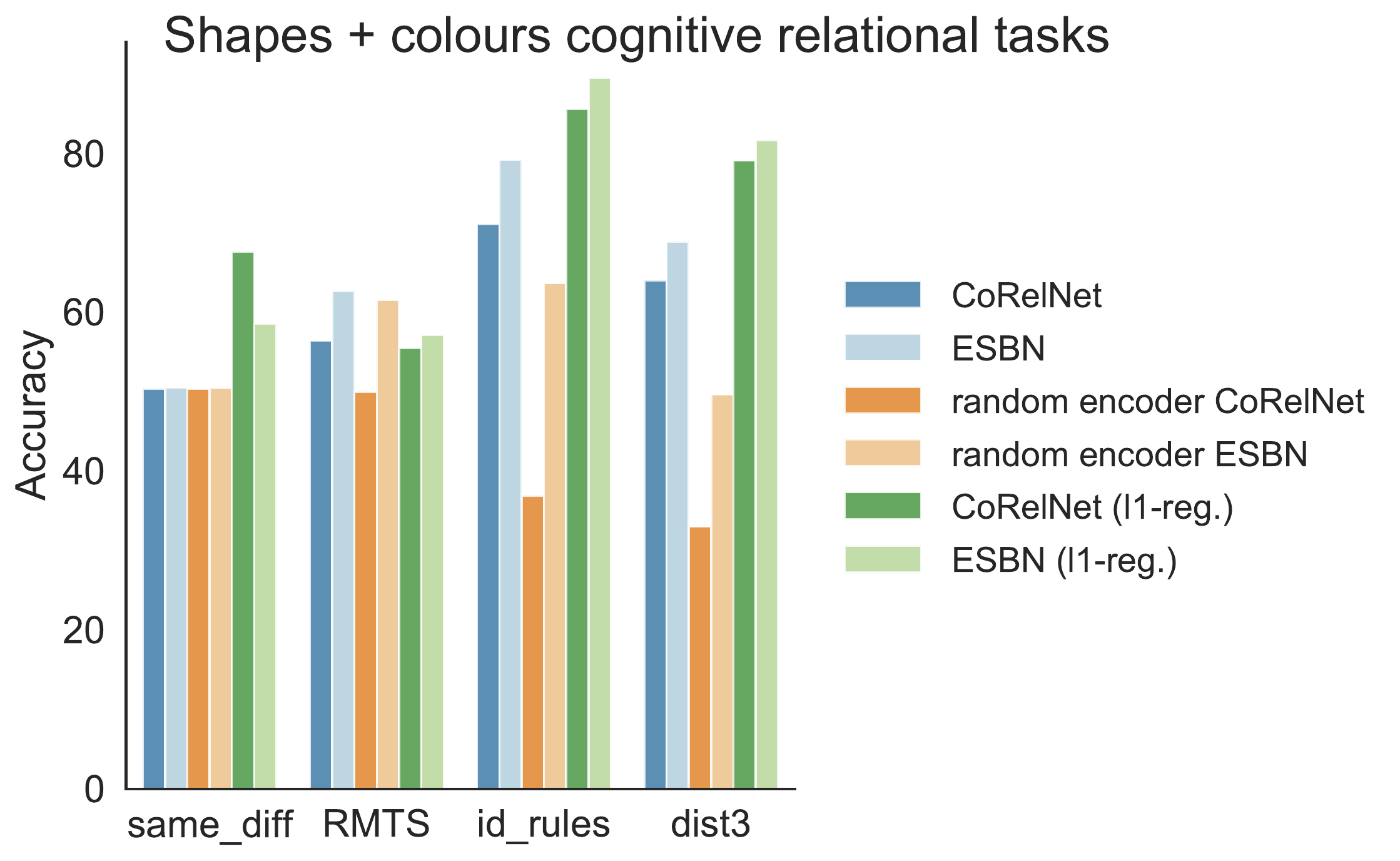}
    \caption{Figure displaying average test performance across 10 seeds for the respective models displayed in the plot. For RMTS regularization coefficient $\lambda=5$ was used for both models, otherwise $\lambda=1$ was used.}
    \label{fig:spurious_detailed_plot}
\end{figure}

Please see Figure \ref{fig:spurious_detailed_plot}.

\section{Spurious features in separated inputs}\label{app: csep_results}
\begin{figure}[t!]
\centering
\begin{subfigure}[c]{0.24\columnwidth}
  \centering
  \includegraphics[width=0.7\textwidth]{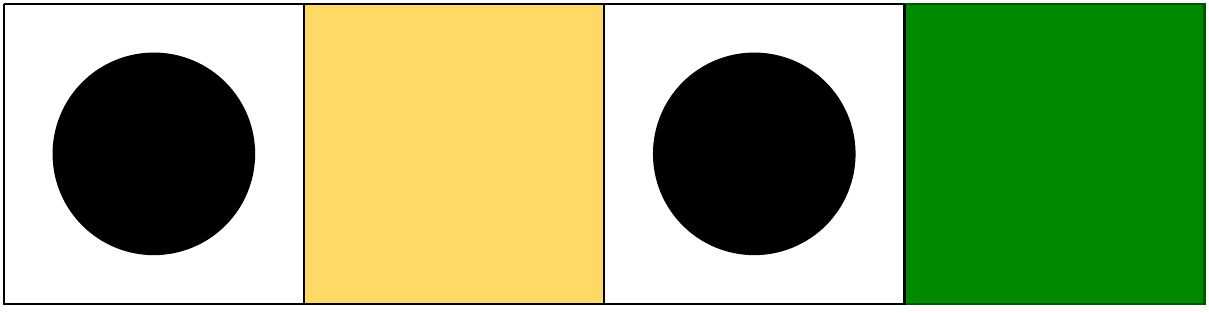}  
  \label{fig:same_diff_csep_task}
\end{subfigure}
\begin{subfigure}[c]{0.75\columnwidth}
  \centering
  \includegraphics[width=\textwidth]{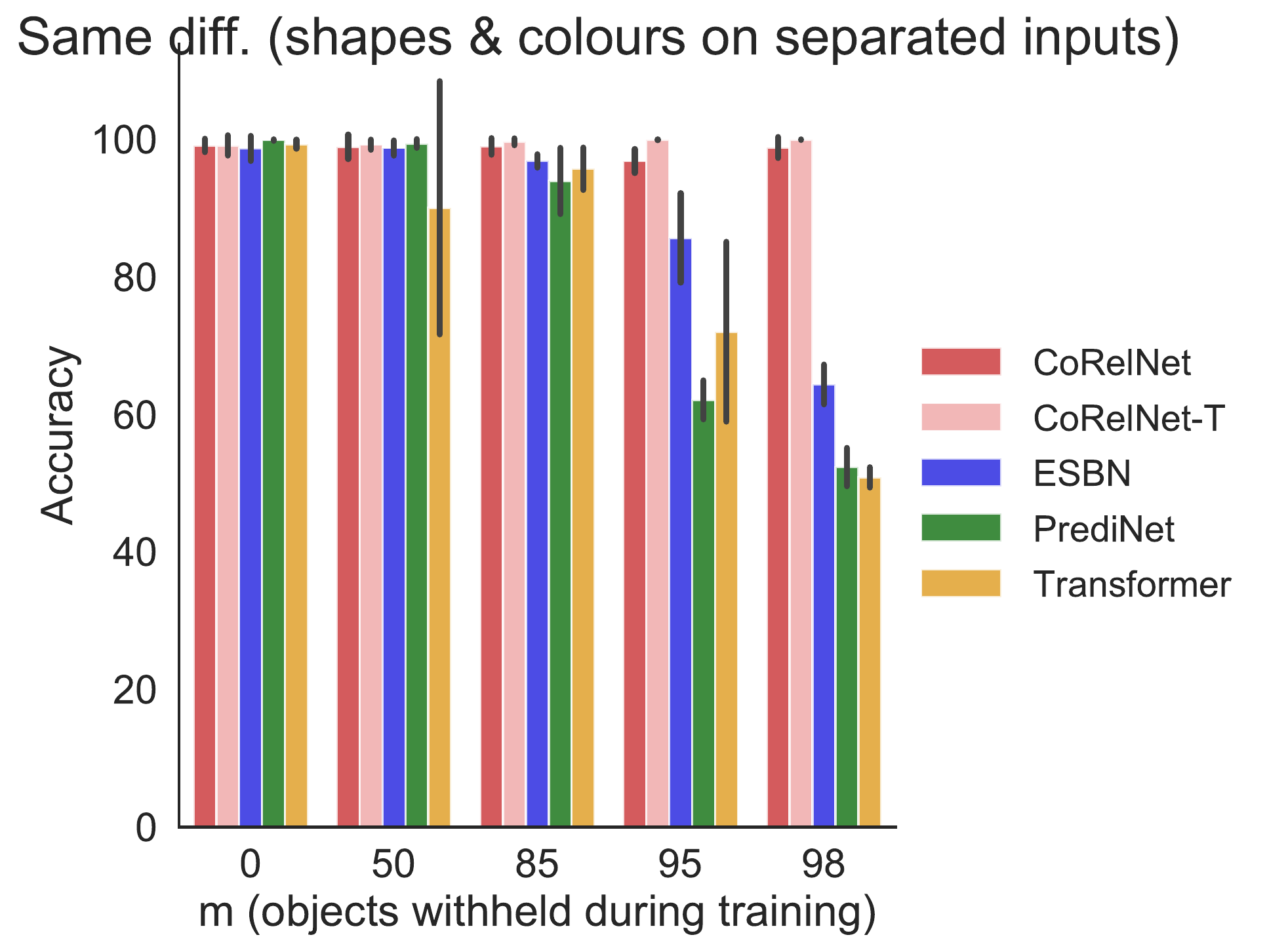}  
  \label{fig:results_csep}
\end{subfigure}
\caption{\textbf{Left.} Task illustration: example for same different task (shapes+colours) on separated inputs. Inputs are given in sequential order shapes, colour, shapes, colour (4 inputs). The task consists in determining whether the two shapes are the same or different irrespective of the colours shown in the sequence. Here the answer would be "same", since both shapes are identical. \textbf{Right.} OoD test accuracy for same different (shapes+colours) on separated inputs. Here the values on the $x$ axis denote the holdout value $m$, for shapes not shown during training. There are a total of 100 shapes, and $m=98$ here means $2$ shapes are shown during training, and testing involves only the $98$ unseen shapes. The set of 100 colours used in training and testing regimes are the same. The case $m=0$ corresponds to the in-distribution case, where the same 100 shapes are shown at testing and training. The test result accuracies are averaged over $10$ random seeds. We can see that for increasing values for the shape holdout parameter $m$ the testing accuracies of ESBN and Transformer are degrading, while CoRelNet maintains close to perfect accuracy. }
\label{fig:csep}
\end{figure}

In this section we investigate a variation of the same different discrimination task with shapes and colours as outlined in Section \ref{sec:harder_unseen_psychophysics_tasks}. Instead of presenting colour and shape in the same image we provide shapes and colours alternatively in different images. Hence we expect that the model can more easily discern the spurious features, as the heavy lifting for ignoring spurious features no longer needs to be done on the level of the encoder but can instead be done on the level of the relational matrix. This is where we believe our model CoRelNet has an advantage over ESBN, as it gets to see the whole similarity matrix all at once, and since spurious features are position dependent, it can decide which positions of the similarity matrix it should consistently ignore. See Figure \ref{fig:csep} for details.

\newpage
\section{Full plots} \label{app:full_plots}

\begin{figure}[ht]
\centering
\begin{subfigure}{0.49\textwidth}
  \centering
  \includegraphics[width=\linewidth]{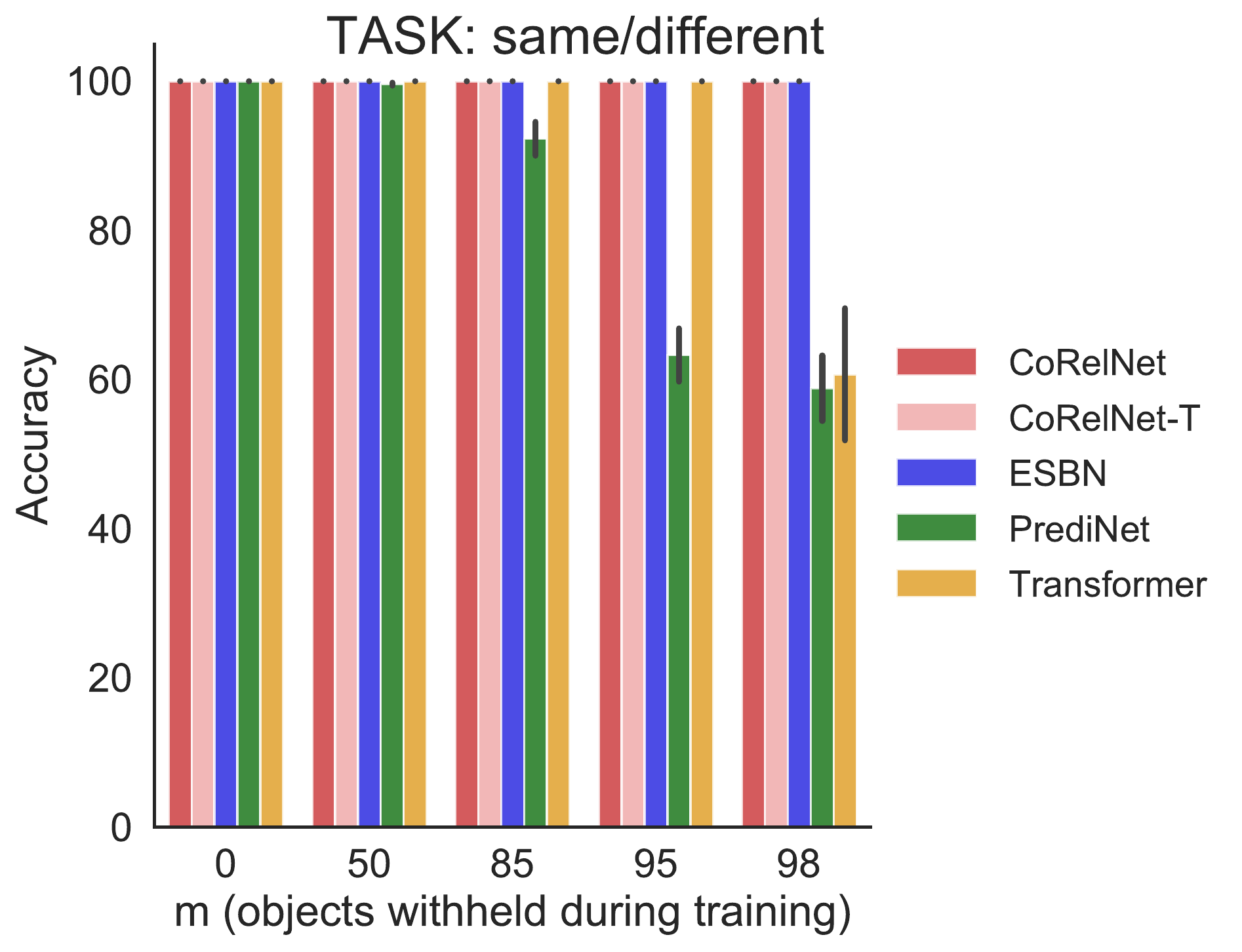}  
  \label{fig:full_results_same_diff}
\end{subfigure}
\begin{subfigure}{0.49\textwidth}
  \centering
  \includegraphics[width=\linewidth]{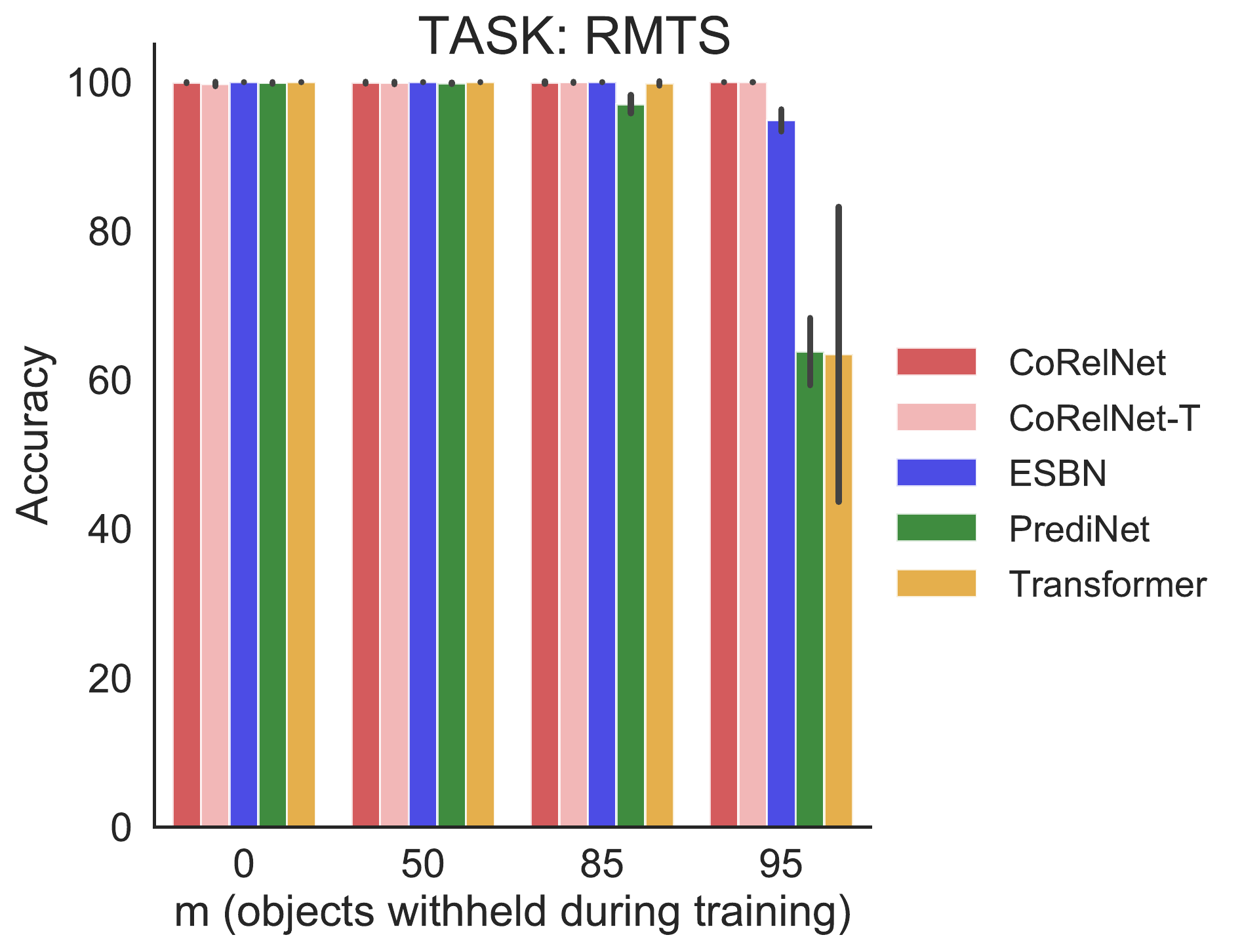}  
  \label{fig:full_results_RMTS}
\end{subfigure}
\newline
\begin{subfigure}{.49\textwidth}
  \centering
  \includegraphics[width=\linewidth]{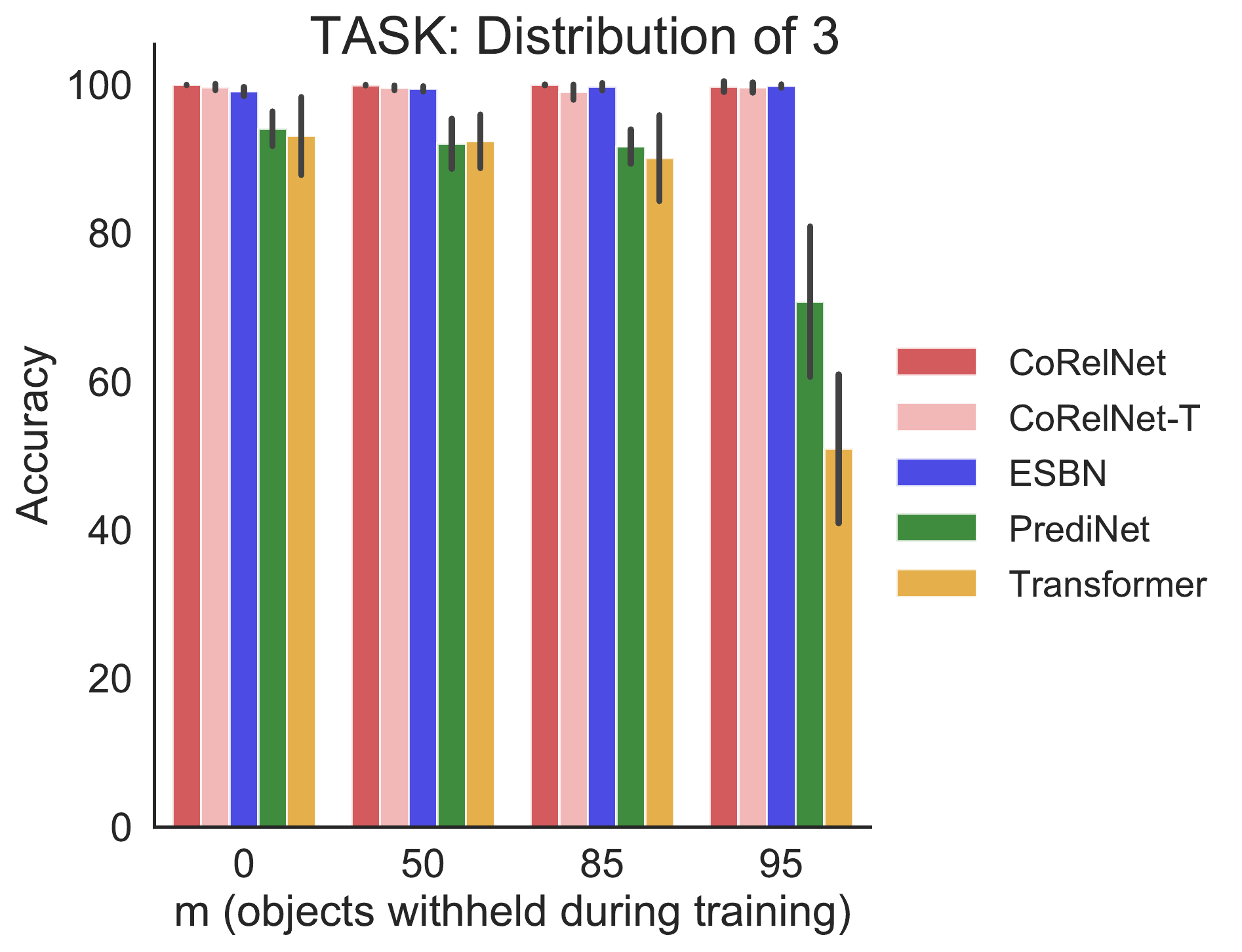}  
  \label{fig:full_results_dist3}
\end{subfigure}
\begin{subfigure}{.49\textwidth}
  \centering
  \includegraphics[width=\linewidth]{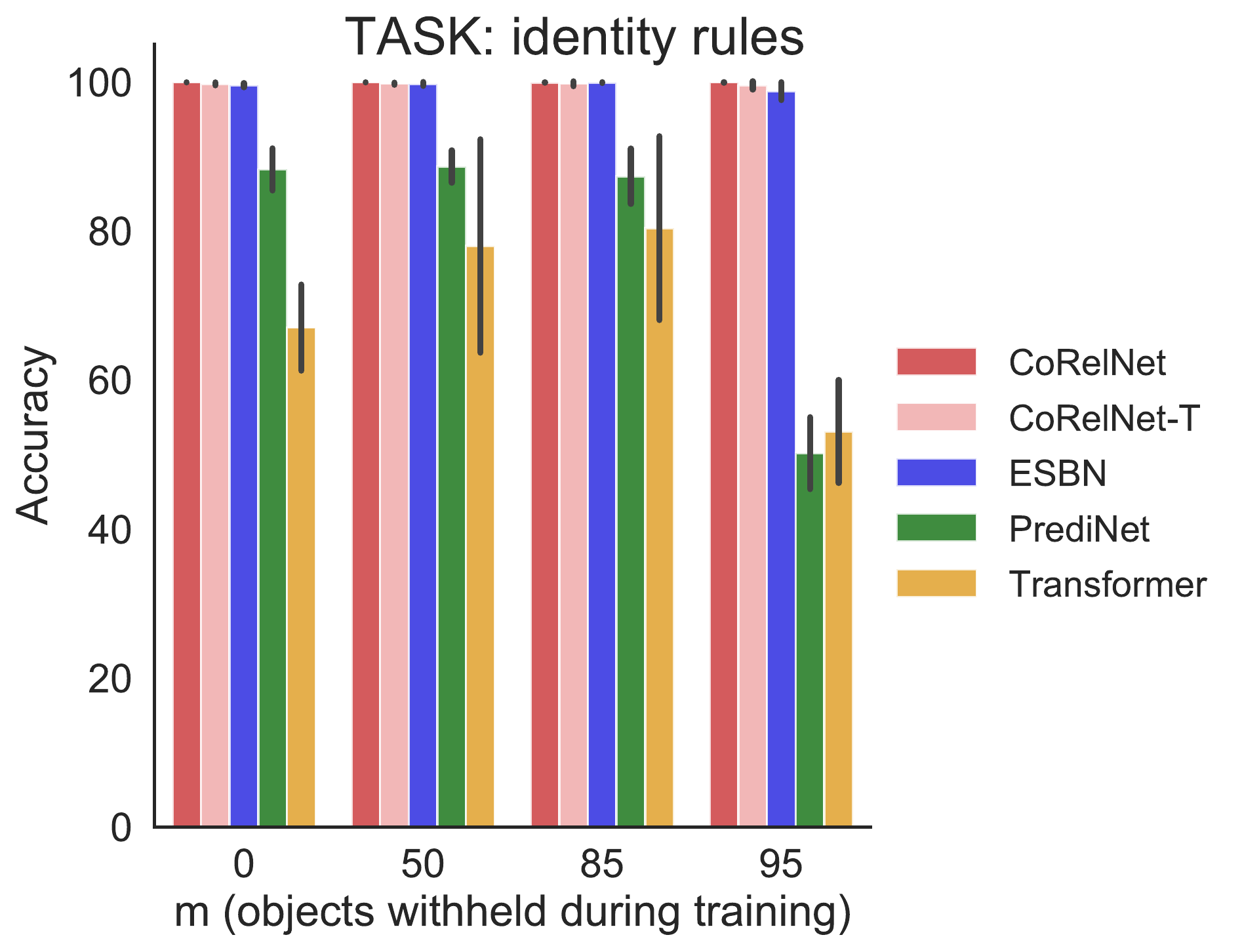}  
  \label{fig:full_results_identity_rules}
\end{subfigure}
\caption{Full detailed test accuracy results for the four basic relational tasks, across the full range of values for $m$ (the number of heldout shapes during training only shown at testing, displayed on the $x$-axis). There are total of $\displaystyle{n} = 100$ shapes, hence $100-m$ of those are shown during training, and the test set consists only of the other $m$ shapes. The most extreme OoD case corresponds to $m=98$ for the same different task (98\% of the possible combinations of shapes are in the test set and not shown during training), and $m=95$ for the 3 other tasks. The case $m=0$ corresponds to the in-distribution case, where the same 100 shapes are shown at testing and training. The test result accuracies are averaged over $10$ random seeds.}
\label{fig:full_results_basic_tasks}
\end{figure}

\begin{figure}[ht]
\centering
\begin{subfigure}{0.49\textwidth}
  \centering
  \includegraphics[width=\linewidth]{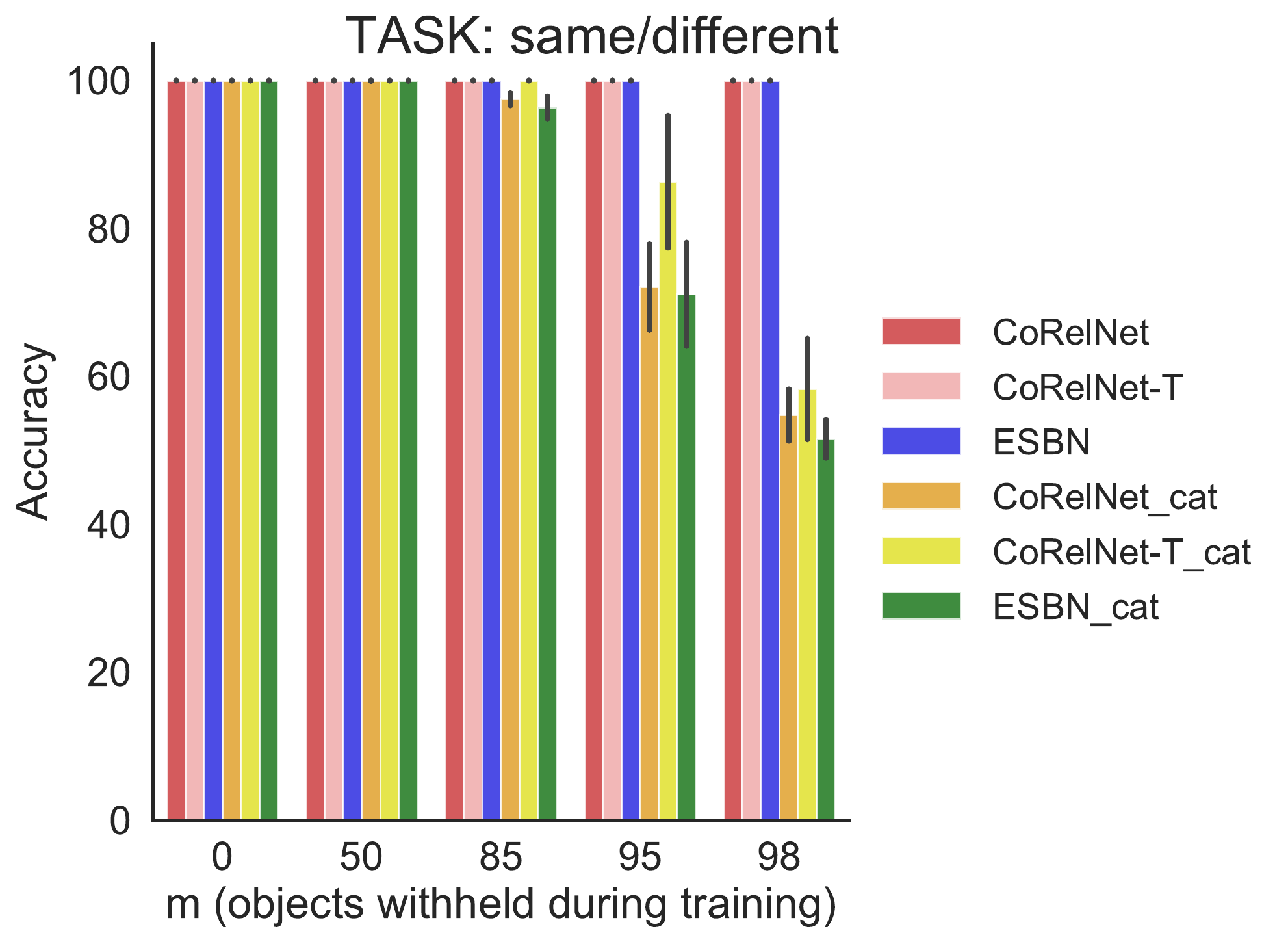}  
  \label{fig:cat_results_same_diff}
\end{subfigure}
\begin{subfigure}{0.49\textwidth}
  \centering
  \includegraphics[width=\linewidth]{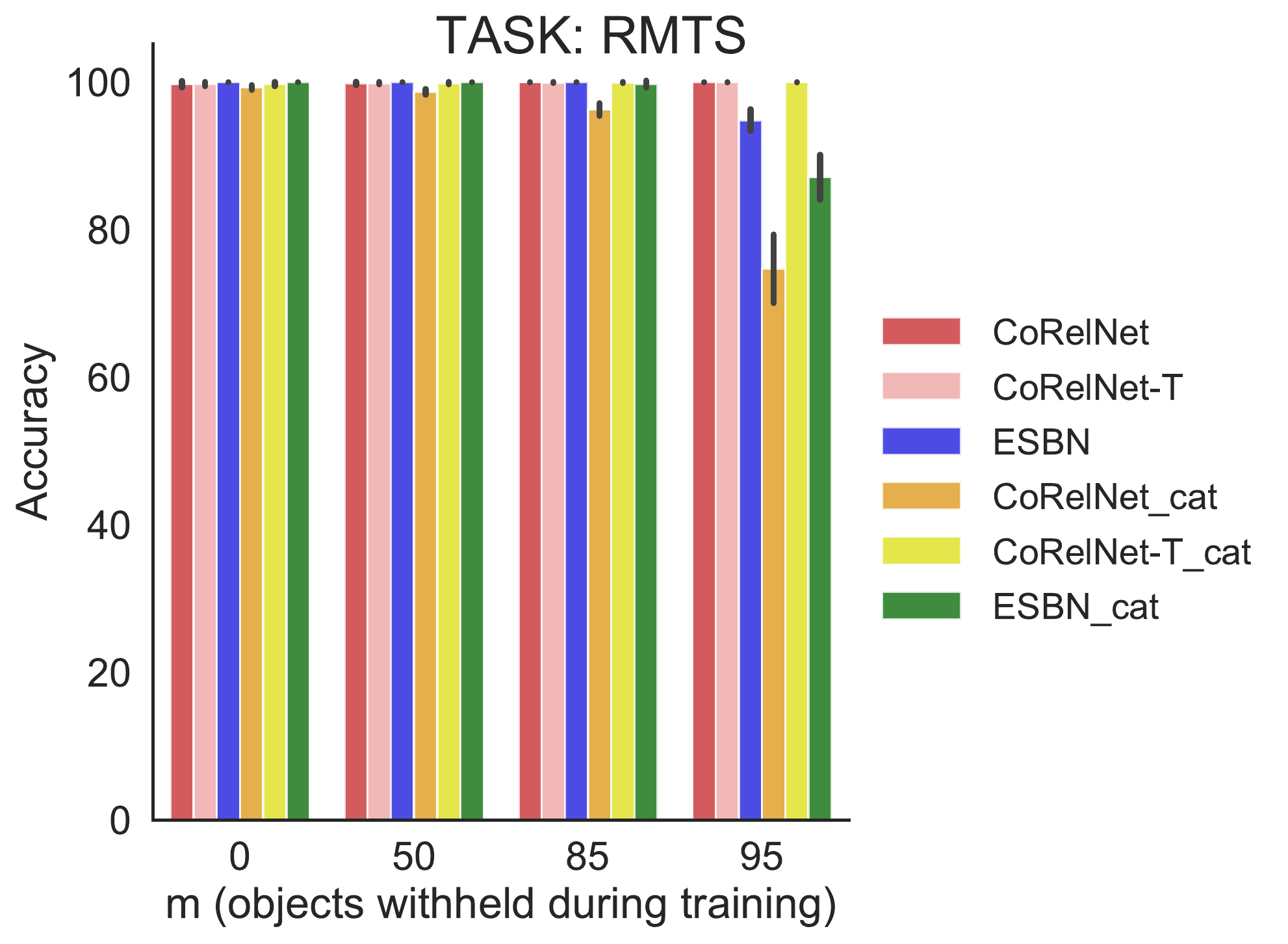}  
  \label{fig:cat_results_RMTS}
\end{subfigure}
\newline
\begin{subfigure}{.49\textwidth}
  \centering
  \includegraphics[width=\linewidth]{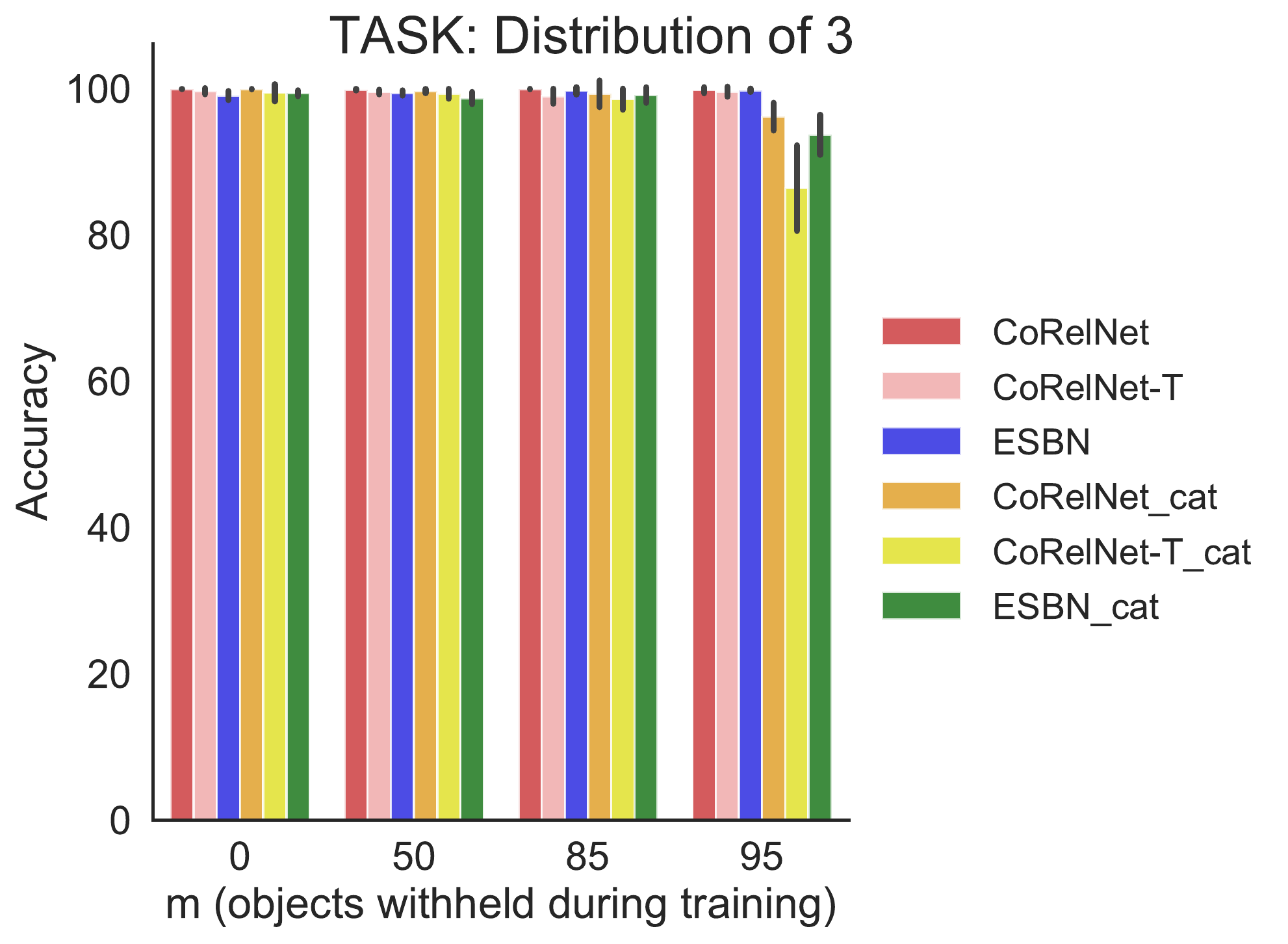}  
  \label{fig:cat_results_dist3}
\end{subfigure}
\begin{subfigure}{.49\textwidth}
  \centering
  \includegraphics[width=\linewidth]{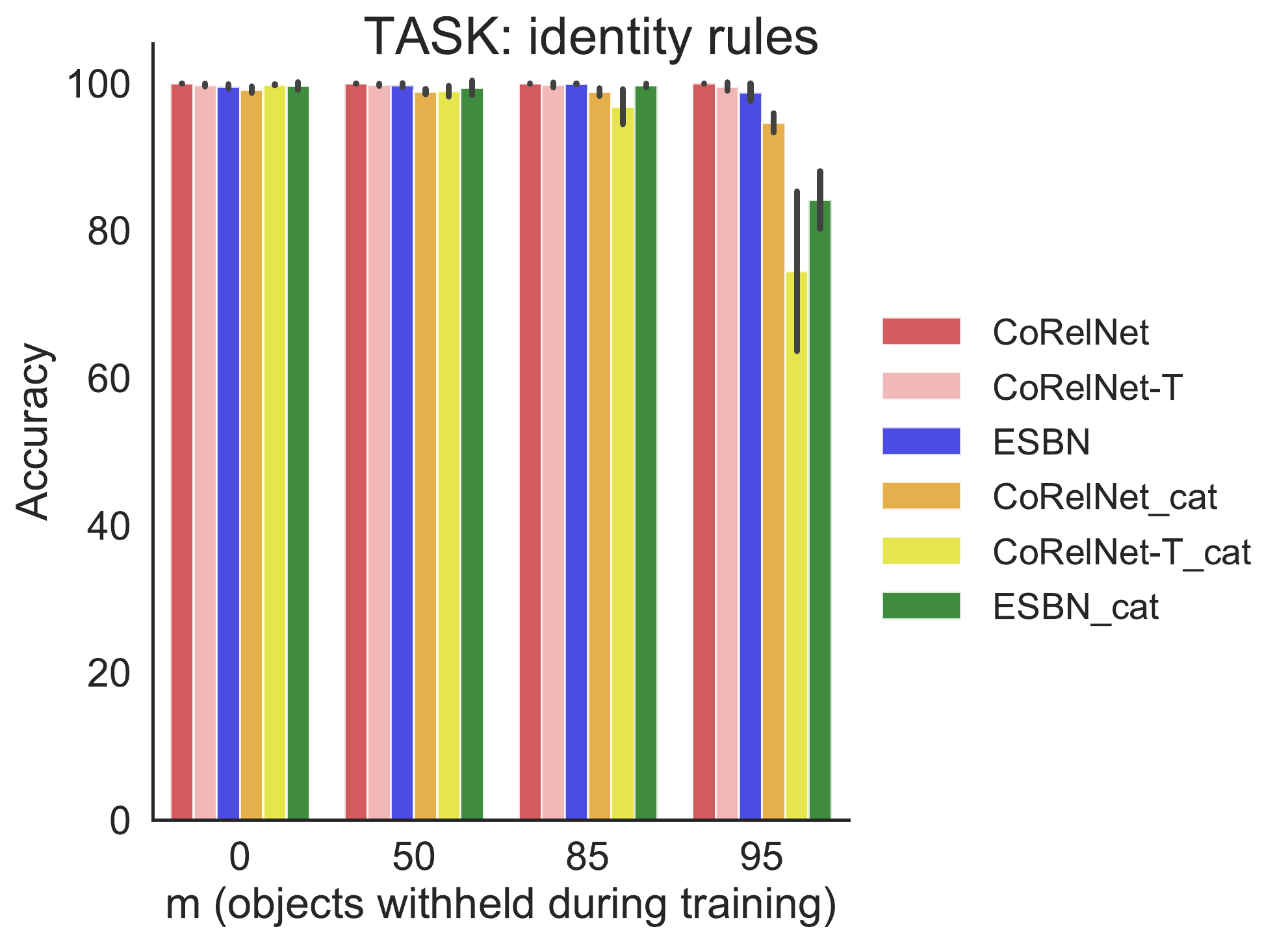}  
  \label{fig:cat_results_identity_rules}
\end{subfigure}
\caption{Full concatenation plots. Full detailed test accuracy results on the four basic relational tasks, across the full range of values for $m$ (the number of heldout shapes during training, displayed on the $x$-axis). There are total of $\displaystyle{n} = 100$ shapes, hence $100-m$ of those are shown during training, and the test set consists only of the other $m$ shapes. The case $m=0$ corresponds to the in-distribution case, where the same 100 shapes are shown at testing and training. Here the suffix "cat" stands for concatenating the encoded input on top of the input of the decoder. The test result accuracies are averaged over $10$ random seeds.}
\label{fig:full_results_concat}
\end{figure}

\begin{figure}[ht]
\centering
\begin{subfigure}{0.49\textwidth}
  \centering
  \includegraphics[width=\linewidth]{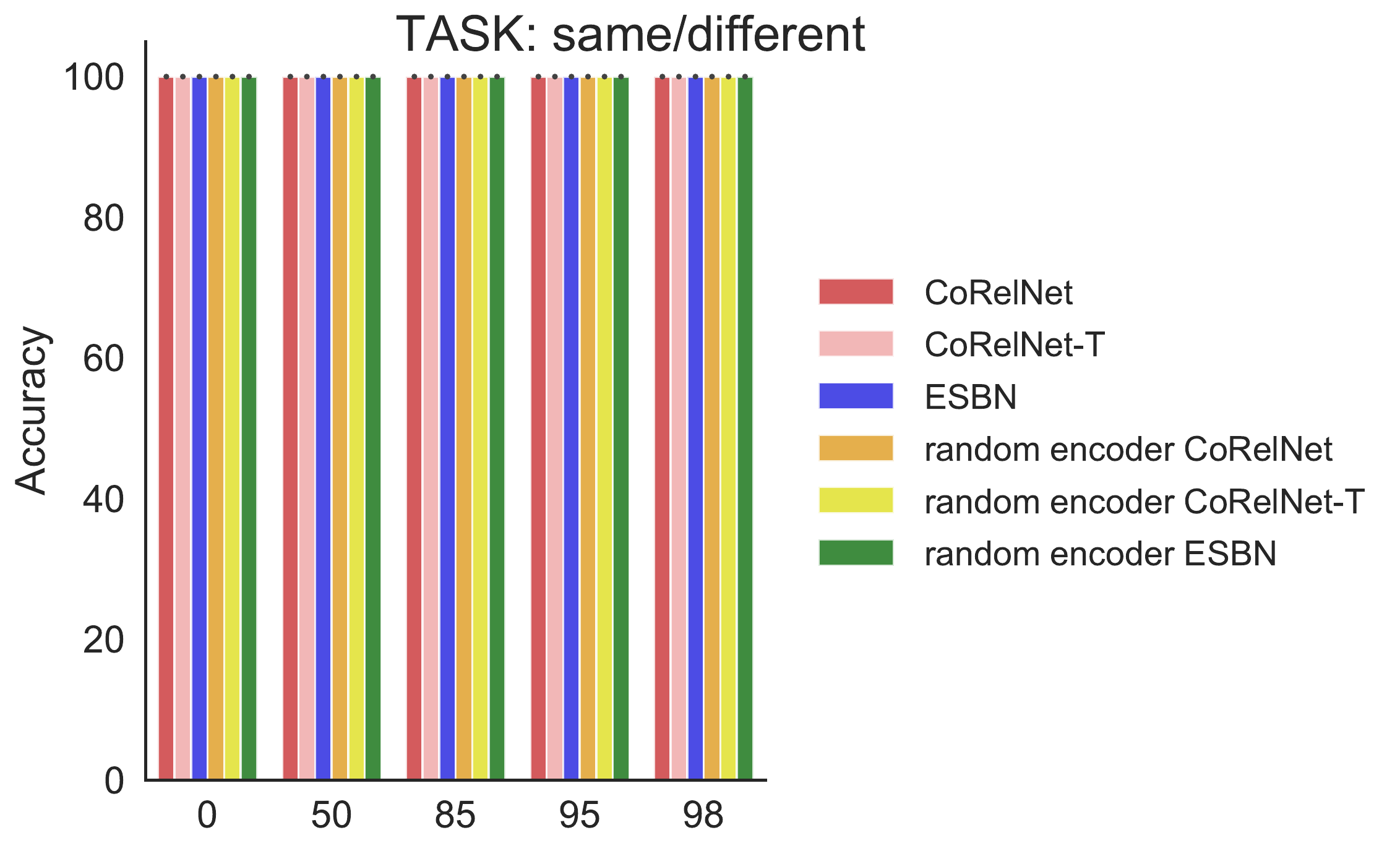}  
  \label{fig:random_results_same_diff}
\end{subfigure}
\begin{subfigure}{0.49\textwidth}
  \centering
  \includegraphics[width=\linewidth]{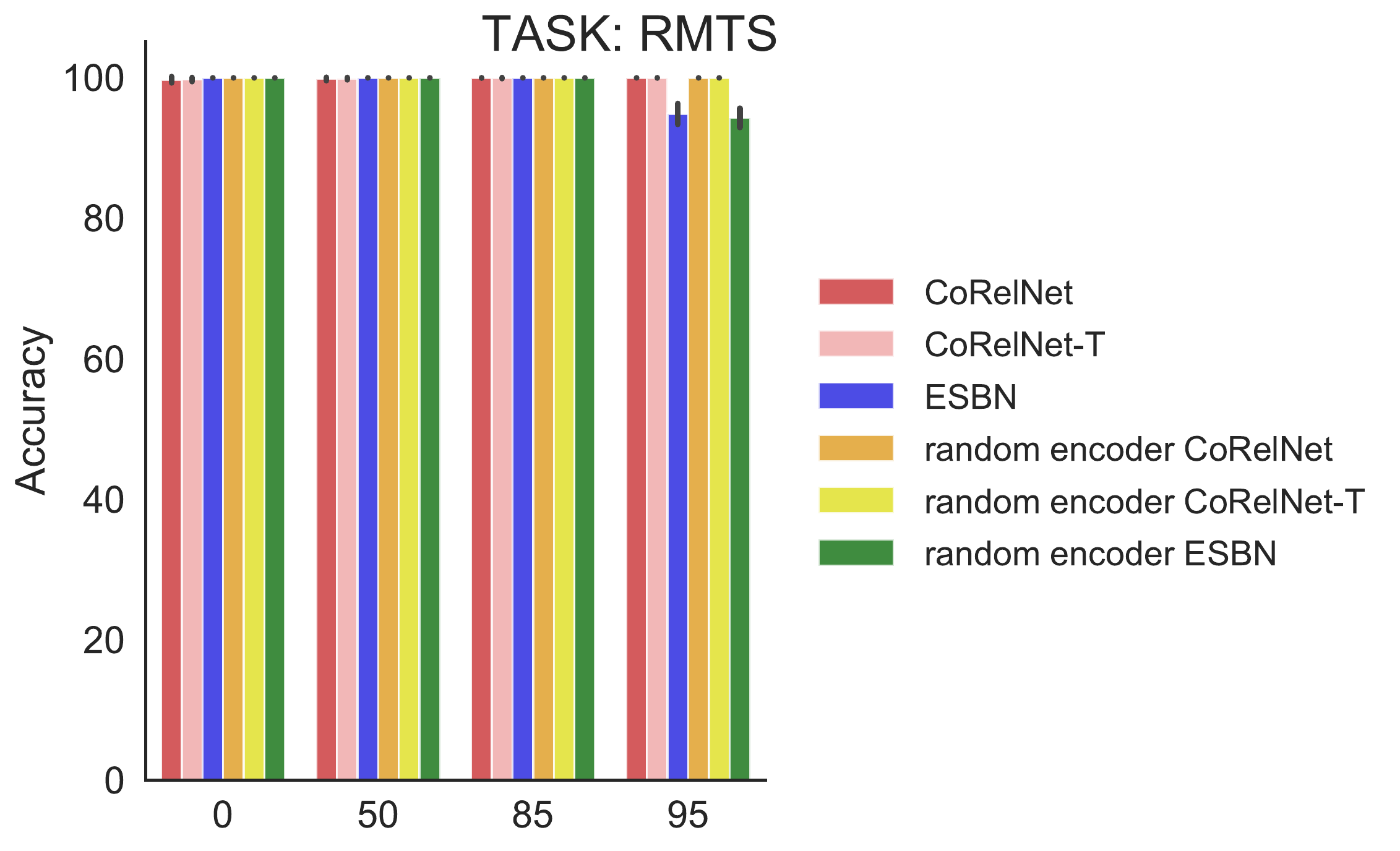}  
  \label{fig:random_results_RMTS}
\end{subfigure}
\newline
\begin{subfigure}{.49\textwidth}
  \centering
  \includegraphics[width=\linewidth]{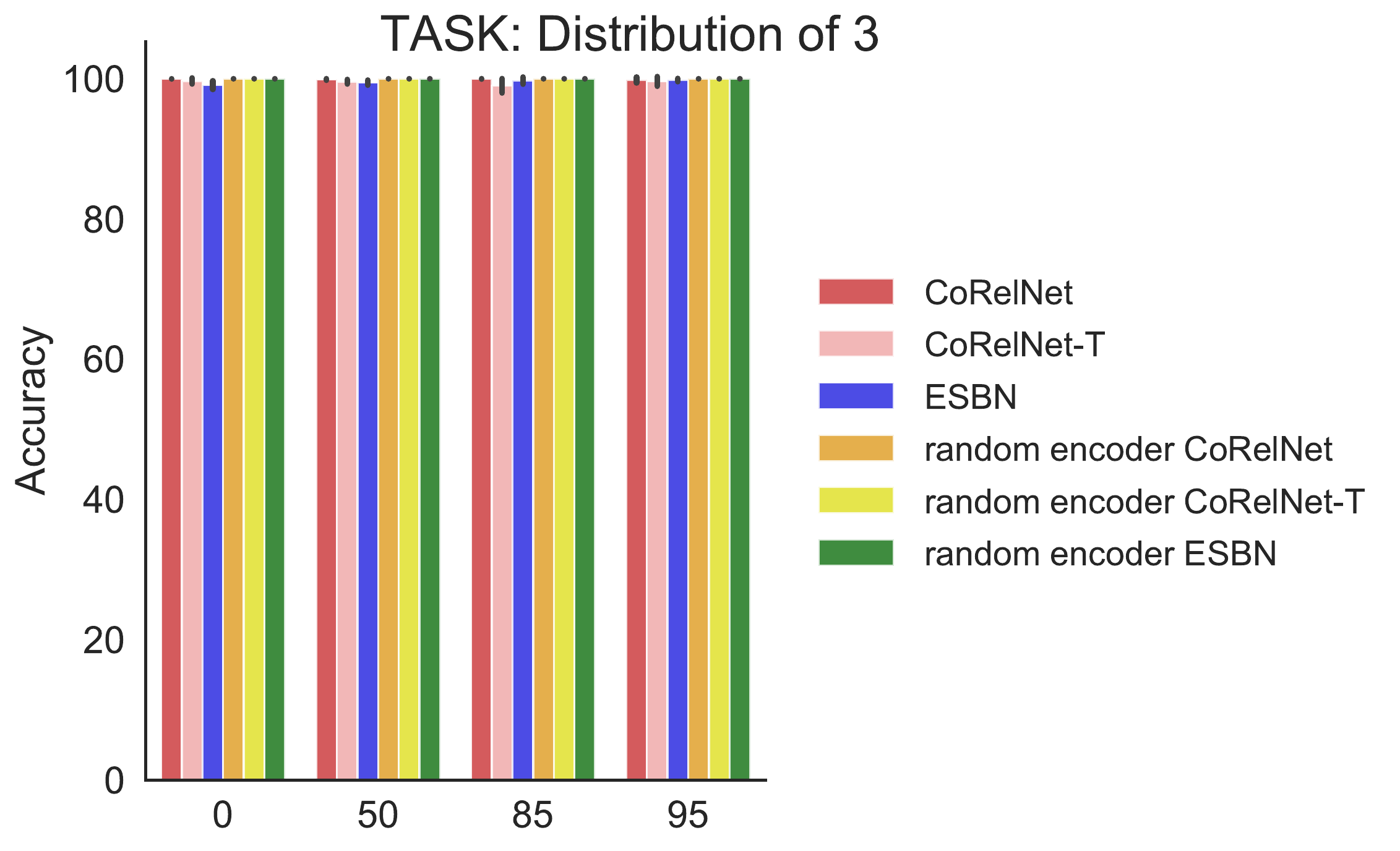}  
  \label{fig:random_results_dist3}
\end{subfigure}
\begin{subfigure}{.49\textwidth}
  \centering
  \includegraphics[width=\linewidth]{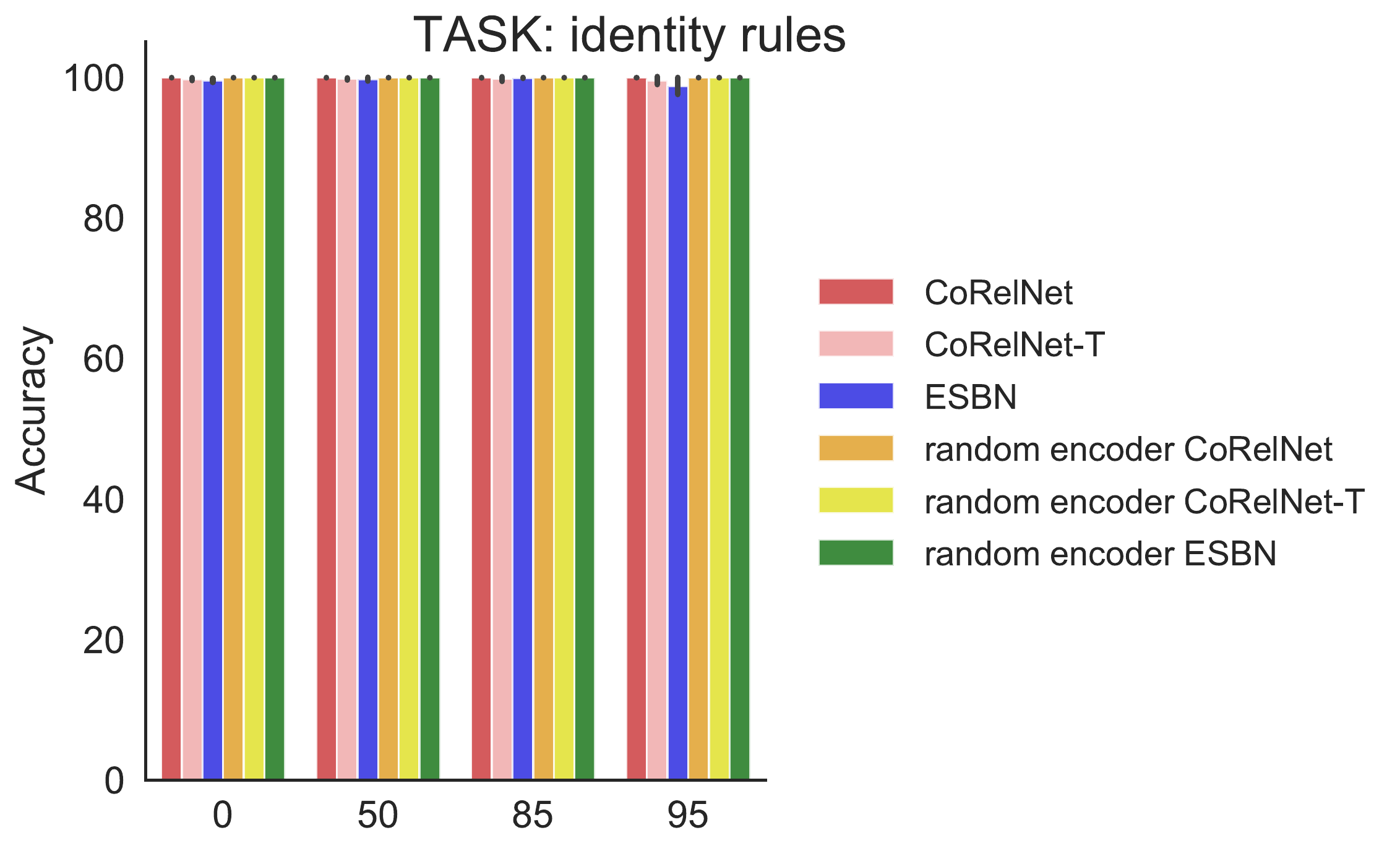} 
  \label{fig:random_results_identity_rules}
\end{subfigure}
\caption{Full random encoder plots. Full detailed test accuracy results on the four basic relational tasks, across the full range of values for $m$ (the number of heldout shapes during training, displayed on the $x$-axis). There are total of $\displaystyle{n} = 100$ shapes, hence $100-m$ of those are shown during training, and the test set consists only of the other $m$ shapes. The case $m=0$ corresponds to the in-distribution case, where the same 100 shapes are shown at testing and training. Here the prefix "random encoder" stands for randomly initializing the encoder but not updating it via backpropagation during training. The test result accuracies are averaged over $10$ random seeds.}
\label{fig:full_results_freeze}
\end{figure}

\begin{figure}[ht]
\centering
\begin{subfigure}{0.49\textwidth}
  \centering
  \includegraphics[width=\linewidth]{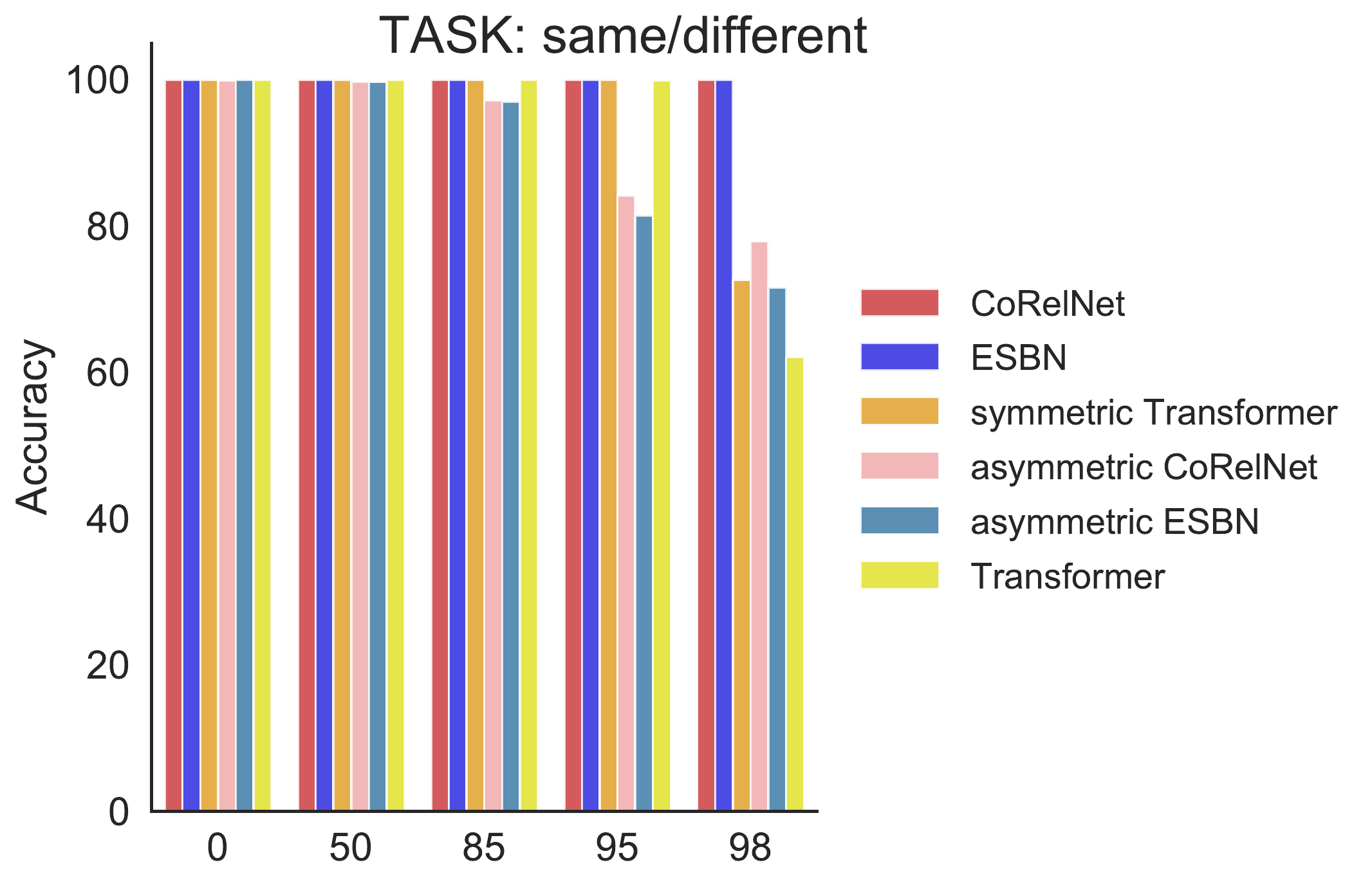}  
  \label{fig:full_symmetry_same_diff}
\end{subfigure}
\begin{subfigure}{0.49\textwidth}
  \centering
  \includegraphics[width=\linewidth]{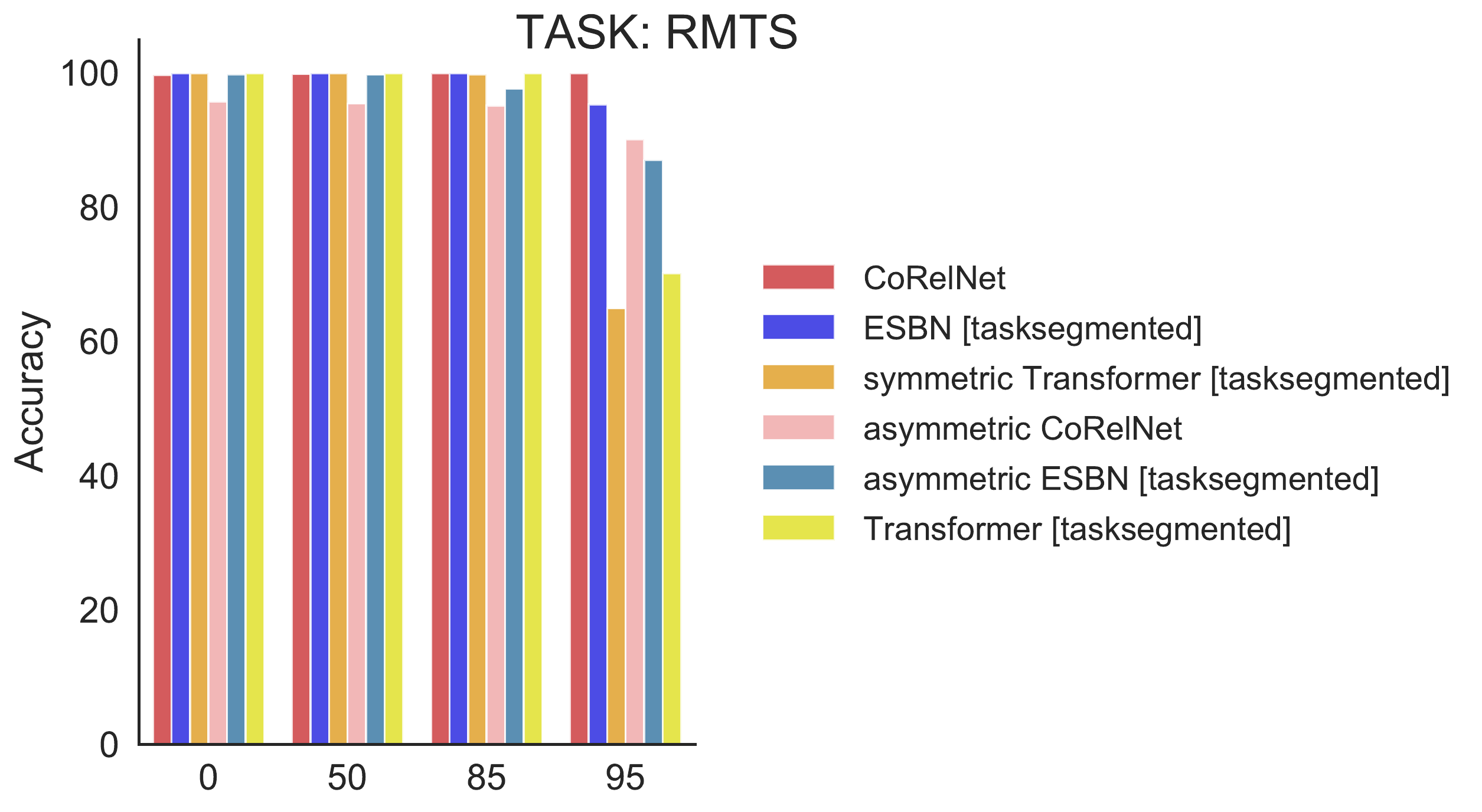}  
  \label{fig:full_symmetry_RMTS}
\end{subfigure}
\newline
\begin{subfigure}{.49\textwidth}
  \centering
  \includegraphics[width=\linewidth]{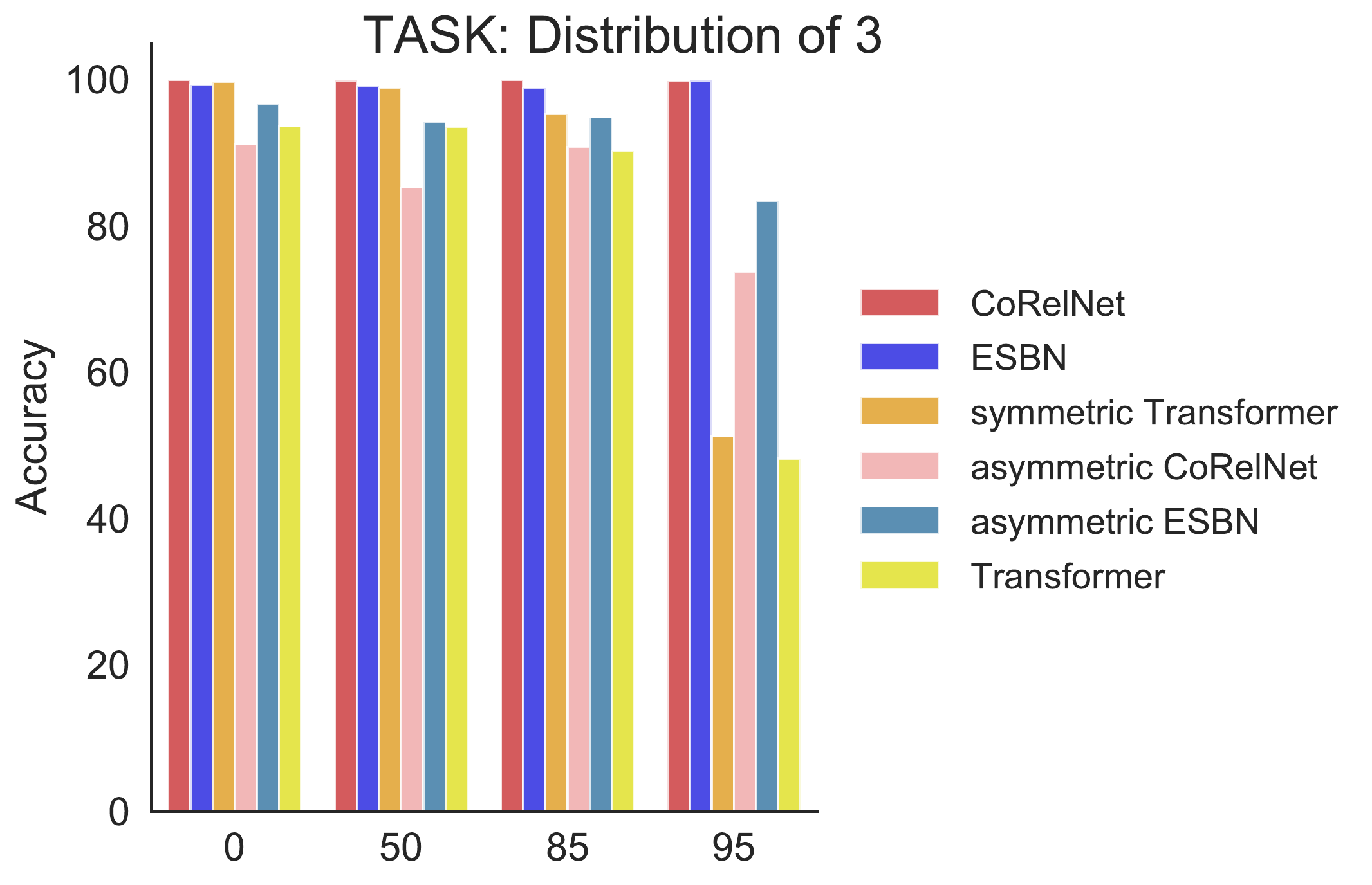}  
  \label{fig:full_symmetry_dist3}
\end{subfigure}
\begin{subfigure}{.49\textwidth}
  \centering
  \includegraphics[width=\linewidth]{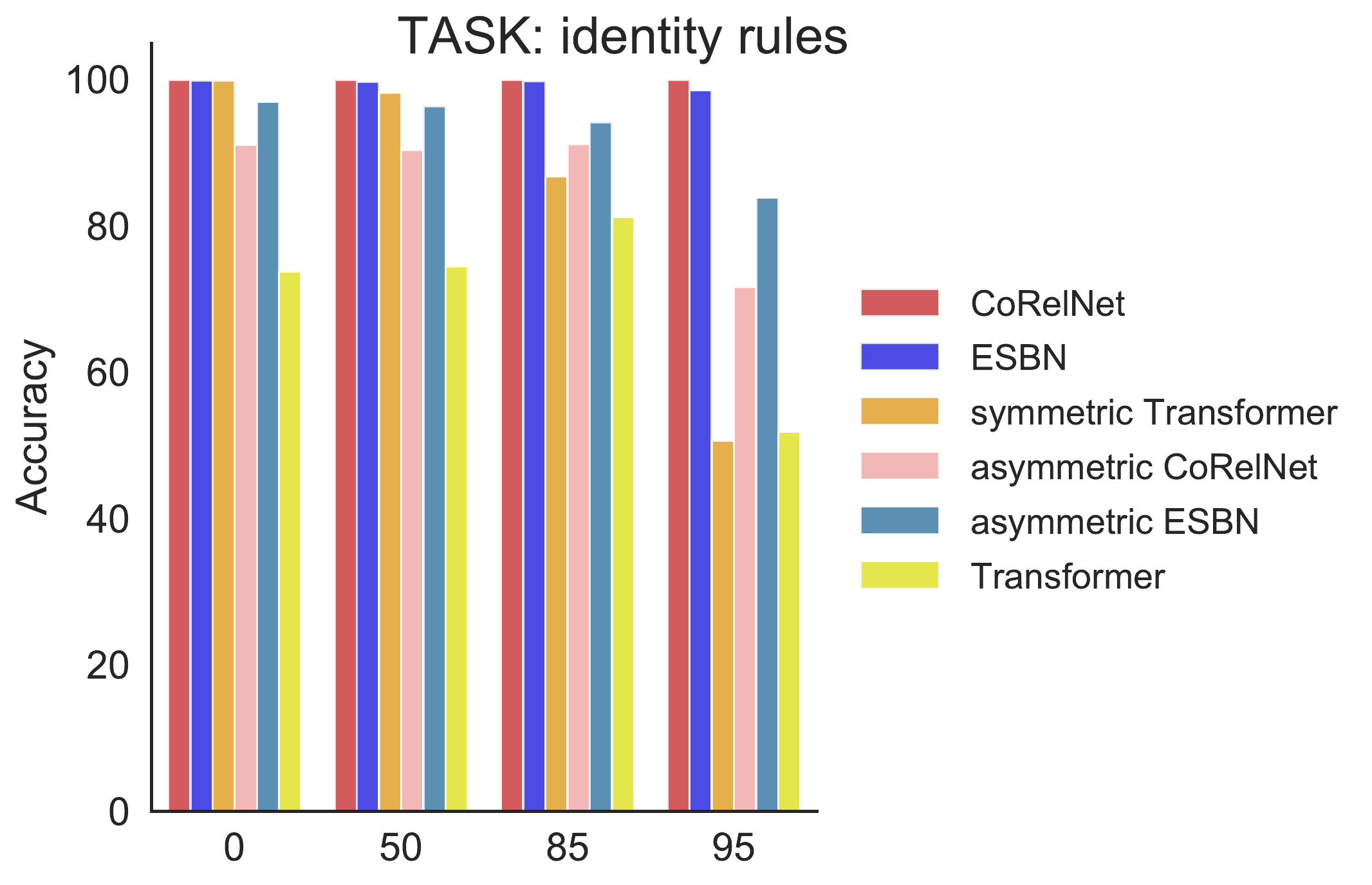} 
  \label{fig:full_symmetry_identity_rules}
\end{subfigure}
\caption{Full symmetry plots. Full detailed test accuracy results on the four basic relational tasks, across the full range of values for $m$ (the number of heldout shapes during training, displayed on the $x$-axis). There are total of $\displaystyle{n} = 100$ shapes, hence $100-m$ of those are shown during training, and the test set consists only of the other $m$ shapes. The case $m=0$ corresponds to the in-distribution case, where the same 100 shapes are shown at testing and training. Here the prefix "asymmetric" stands for replacing the symmetric dot-product $z_t^\top z_t$ by the asymmetric $(W_1 \cdot z_t)^\top (W_2 \cdot z_t)$, whenever self-attention is performed. Similarly, the prefix "symmetric" stands for replacing the 'asymmetric' dot-product $(Q \cdot z_t)^\top (K \cdot z_t)$ by the symmetric $(Q\cdot z_t)^\top (Q \cdot z_t)$ counterpart, whenever self-attention is performed. However in our analysis in the main text for symmetric vs asymmetric, we did not include symmetric Transformer and Transformer, since they are not built upon the same inductive bias. The test result accuracies are averaged over $10$ random seeds.}
\label{fig:full_results_symmetry}
\end{figure}

\begin{figure}[ht]
\centering
\begin{subfigure}{0.49\textwidth}
  \centering
  \includegraphics[width=\linewidth]{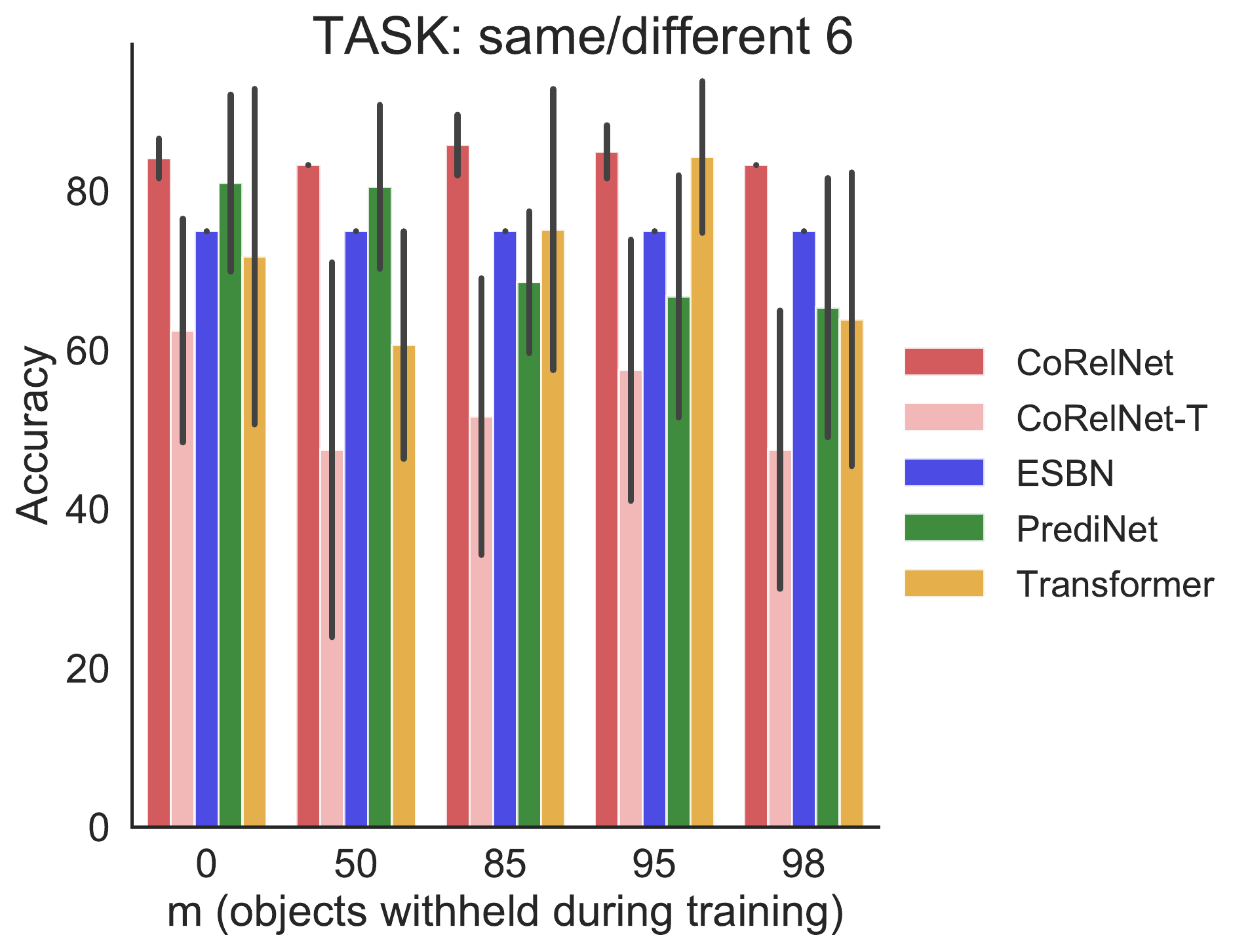}  
  \label{fig:full_results_same_diff_6}
\end{subfigure}
\begin{subfigure}{0.49\textwidth}
  \centering
  \includegraphics[width=\linewidth]{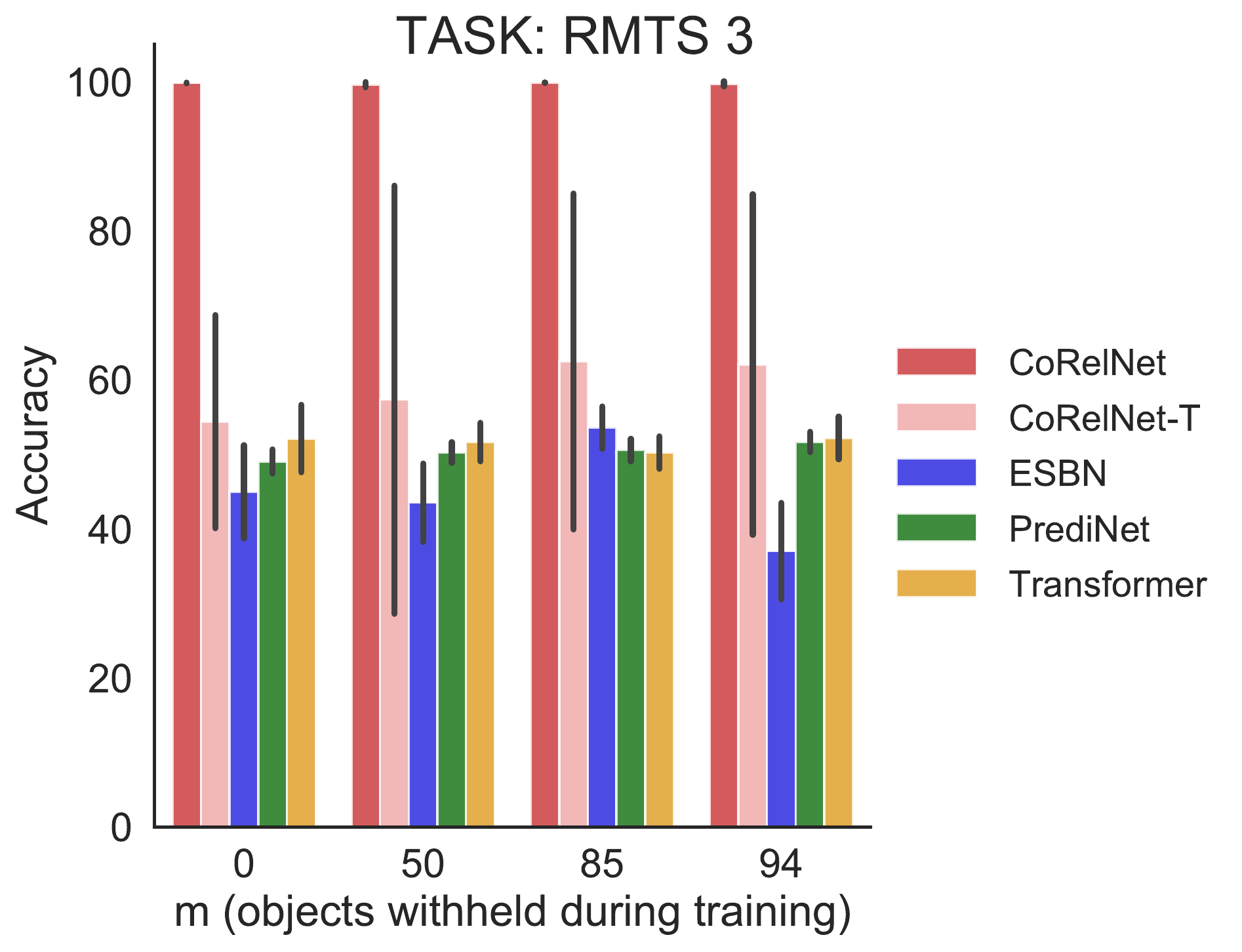}  
  \label{fig:full_results_RMTS3}
\end{subfigure}
\newline
\begin{subfigure}{.49\textwidth}
  \centering
  \includegraphics[width=\linewidth]{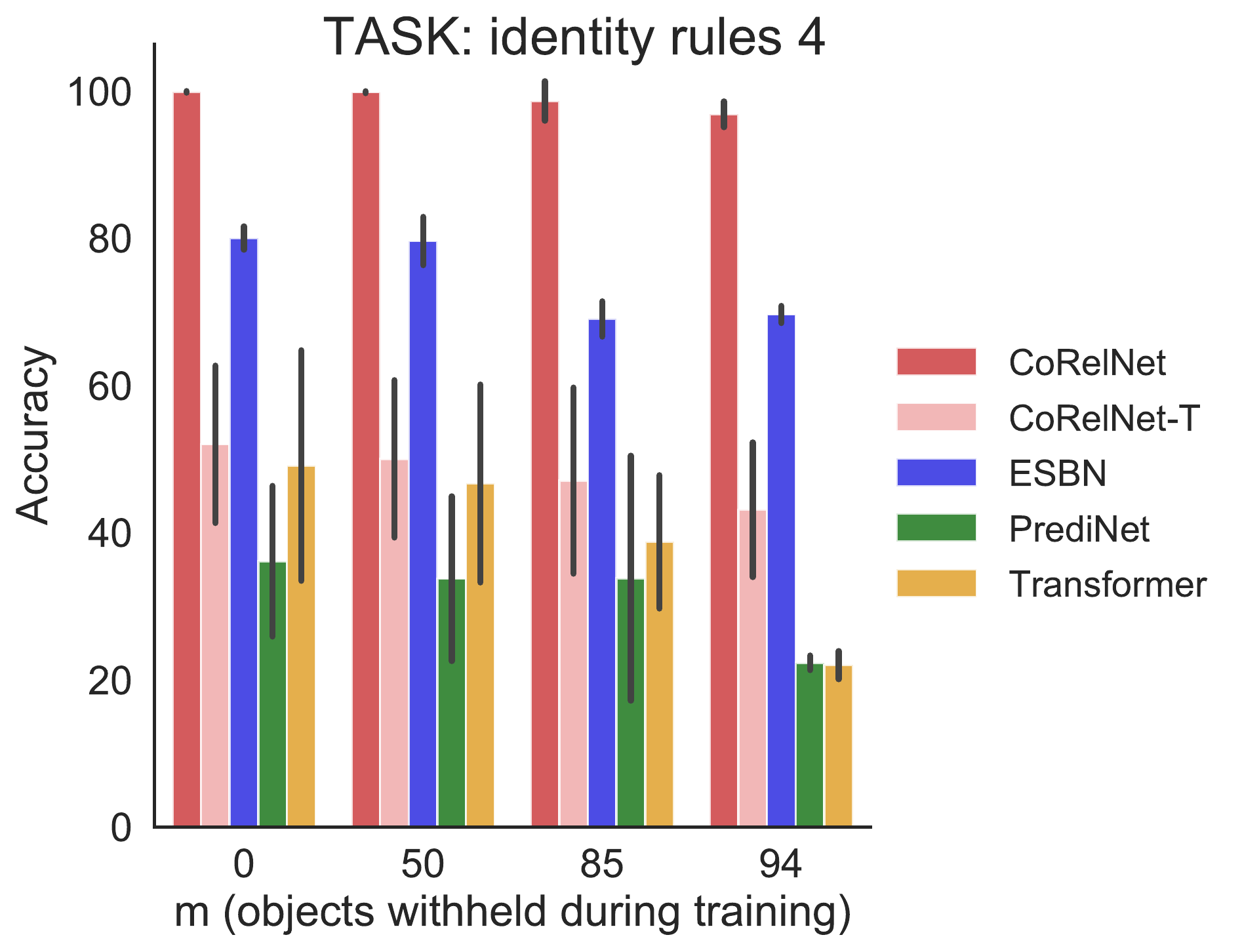}  
  \label{fig:full_results_identity_rules4}
\end{subfigure}
\begin{subfigure}{.49\textwidth}
  \centering
  \includegraphics[width=\linewidth]{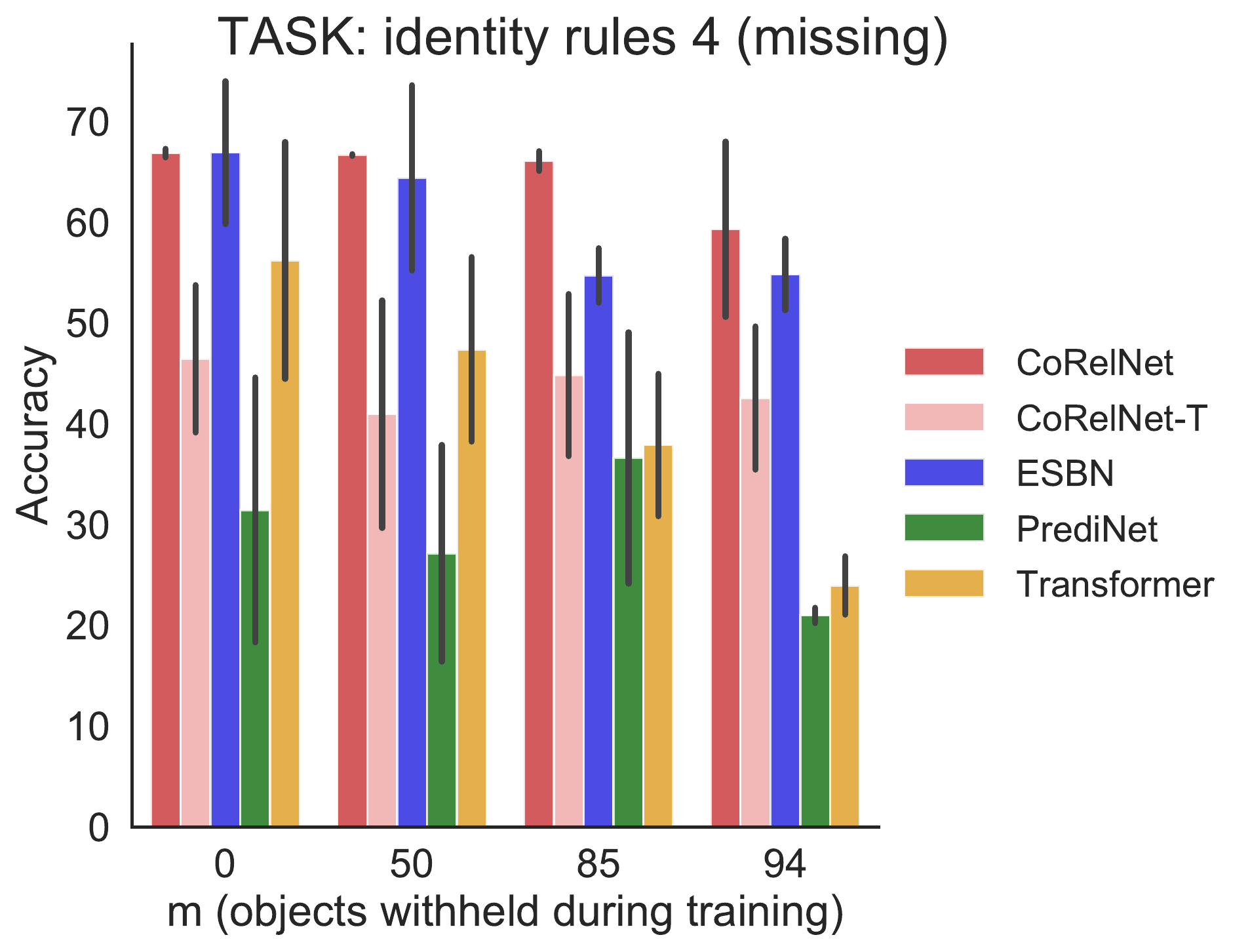}  
  \label{fig:full_results_identity_rules4_missing}
\end{subfigure}
\caption{Full detailed test accuracy results on the harder relational tasks with unseen relations, across the full range of values for $m$ (the number of heldout shapes during training, displayed on the $x$-axis). There are total of $\displaystyle{n} = 100$ shapes, hence $100-m$ of those are shown during training, and the test set consists only of the other $m$ shapes. The case $m=98$ corresponds to the most extreme OoD case for same/different 6, and $m=94$ for the other three tasks. The case $m=0$ corresponds to the in-distribution case, where the same 100 shapes are shown at testing and training. The test result accuracies are averaged over $10$ random seeds. }
\label{fig:full_results_harder_unseen}
\end{figure}

\begin{figure}[ht]
\centering
\begin{subfigure}{0.49\textwidth}
  \centering
  \includegraphics[width=\linewidth]{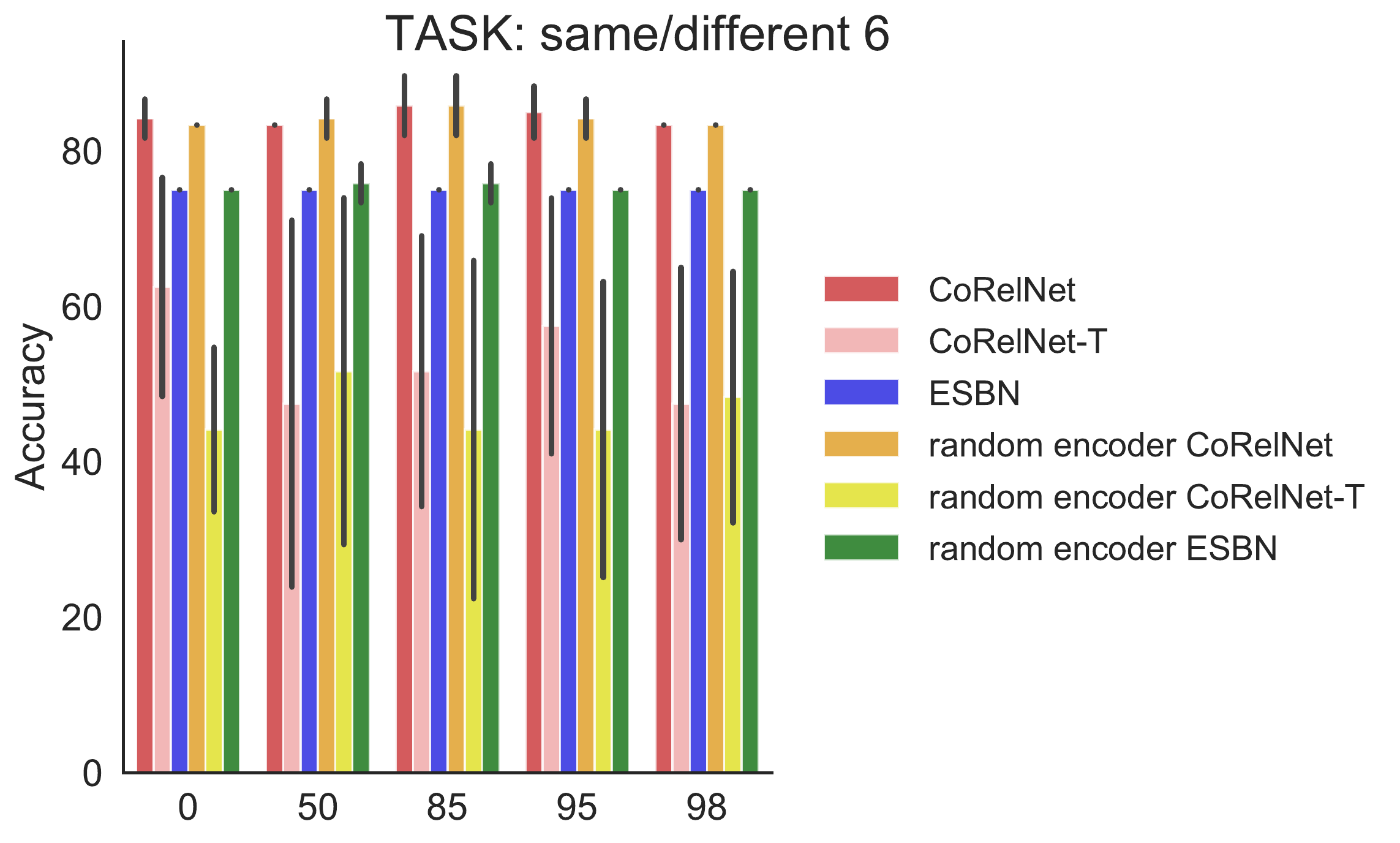}
  \label{fig:random_results_same_diff6}
\end{subfigure}
\begin{subfigure}{0.49\textwidth}
  \centering
  \includegraphics[width=\linewidth]{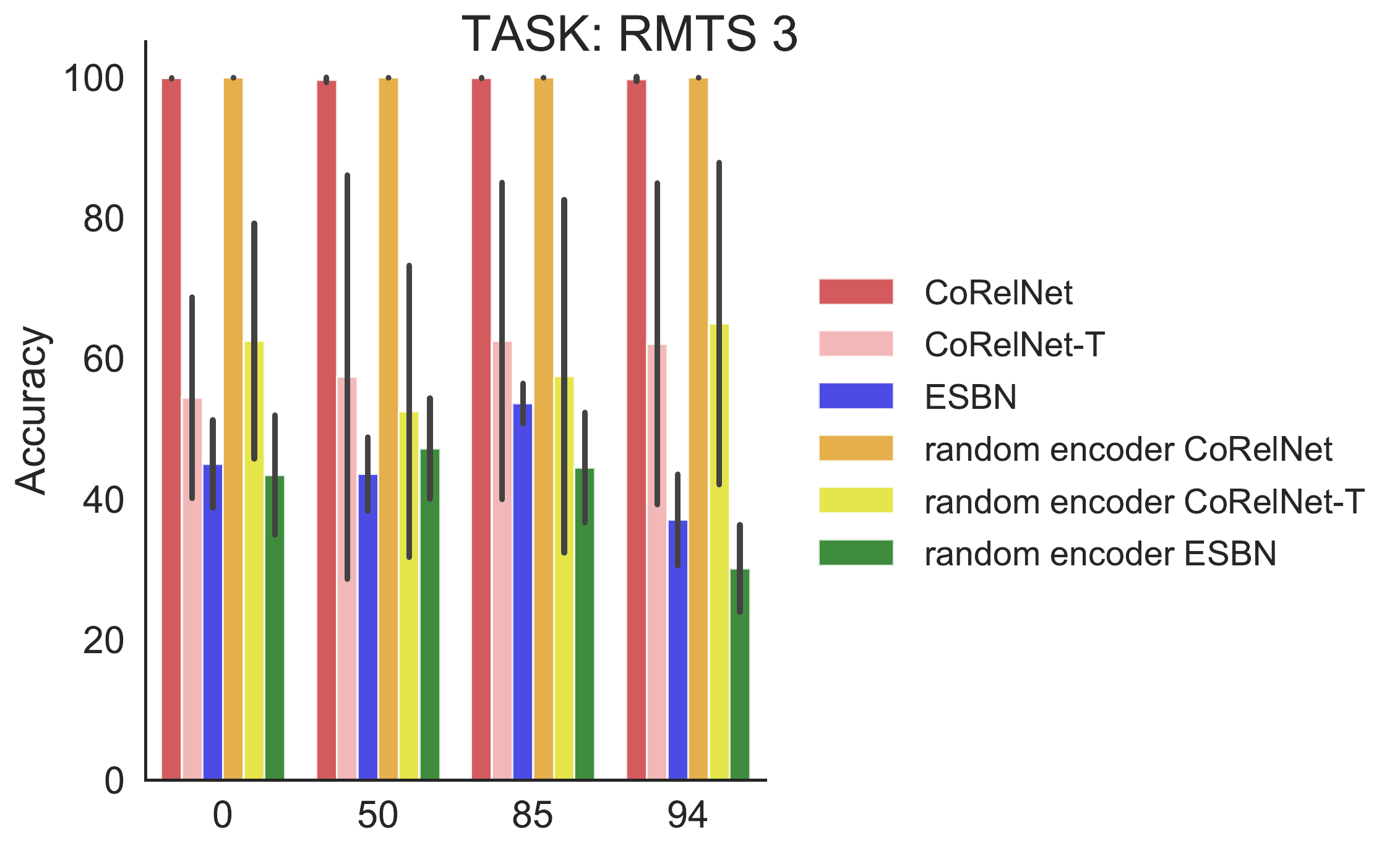}  
  \label{fig:random_results_rmts3}
\end{subfigure}
\newline
\begin{subfigure}{0.49\textwidth}
  \centering
  \includegraphics[width=\linewidth]{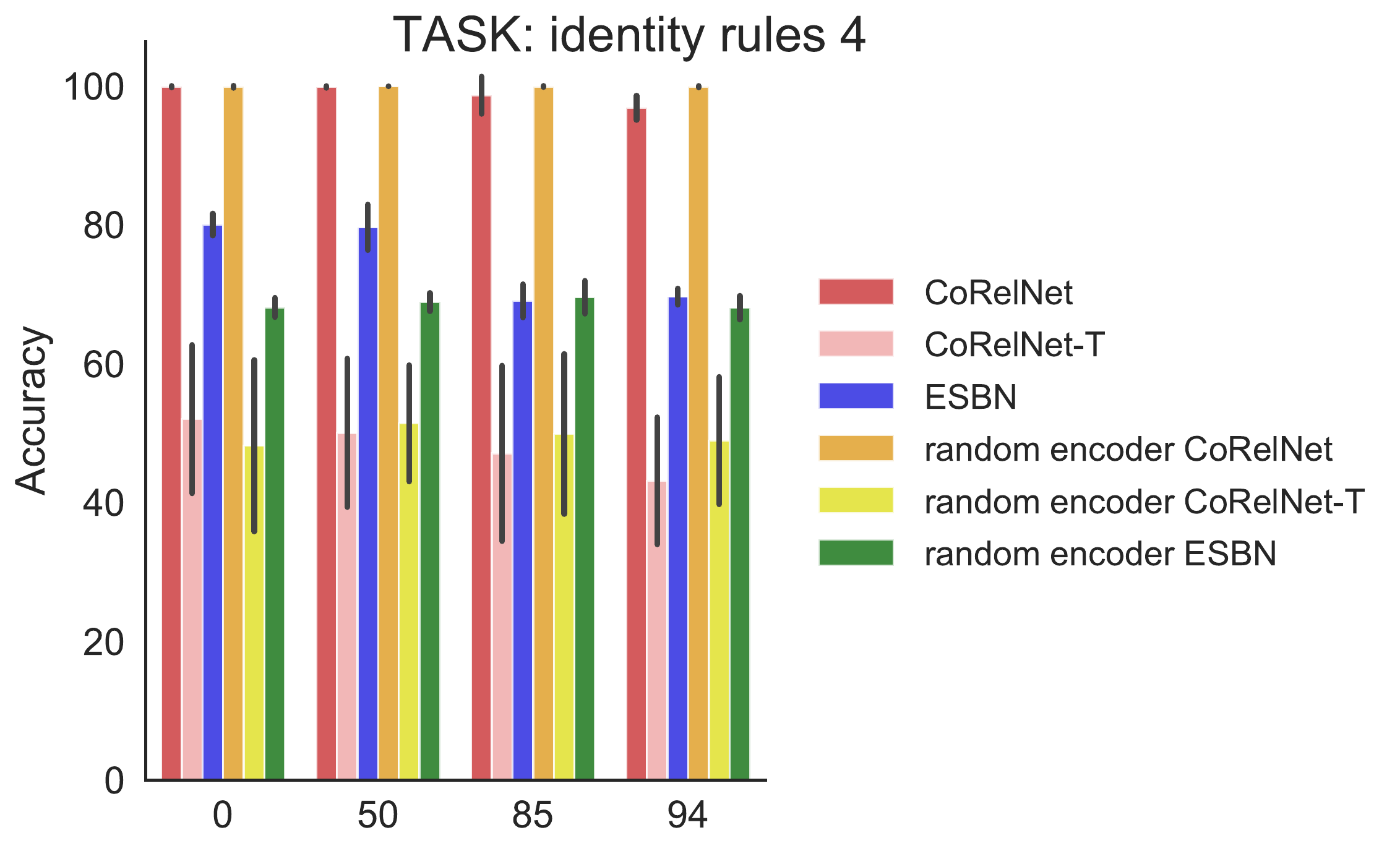}  
  \label{fig:random_results_identity_rules4}
\end{subfigure}
\caption{Full random encoder plots. Full detailed test accuracy results on the harder relational tasks with unseen relations, across the full range of values for $m$ (the number of heldout shapes during training, displayed on the $x$-axis). There are total of $\displaystyle{n} = 100$ shapes, hence $100-m$ of those are shown during training, and the test set consists only of the other $m$ shapes. The case $m=98$ corresponds to the most extreme OoD case for same/different 6, and $m=94$ for the other three tasks. The case $m=0$ corresponds to the in-distribution case, where the same 100 shapes are shown at testing and training. Here the prefix "random encoder" stands for randomly initializing the encoder but not updating it via backpropagation during training. The test result accuracies are averaged over $10$ random seeds.}
\label{fig:full_results_harder_unseen_freeze}
\end{figure}

\begin{figure}[t!]
\centering
\begin{subfigure}[c]{0.48\columnwidth}
  \centering
  \includegraphics[width=\textwidth]{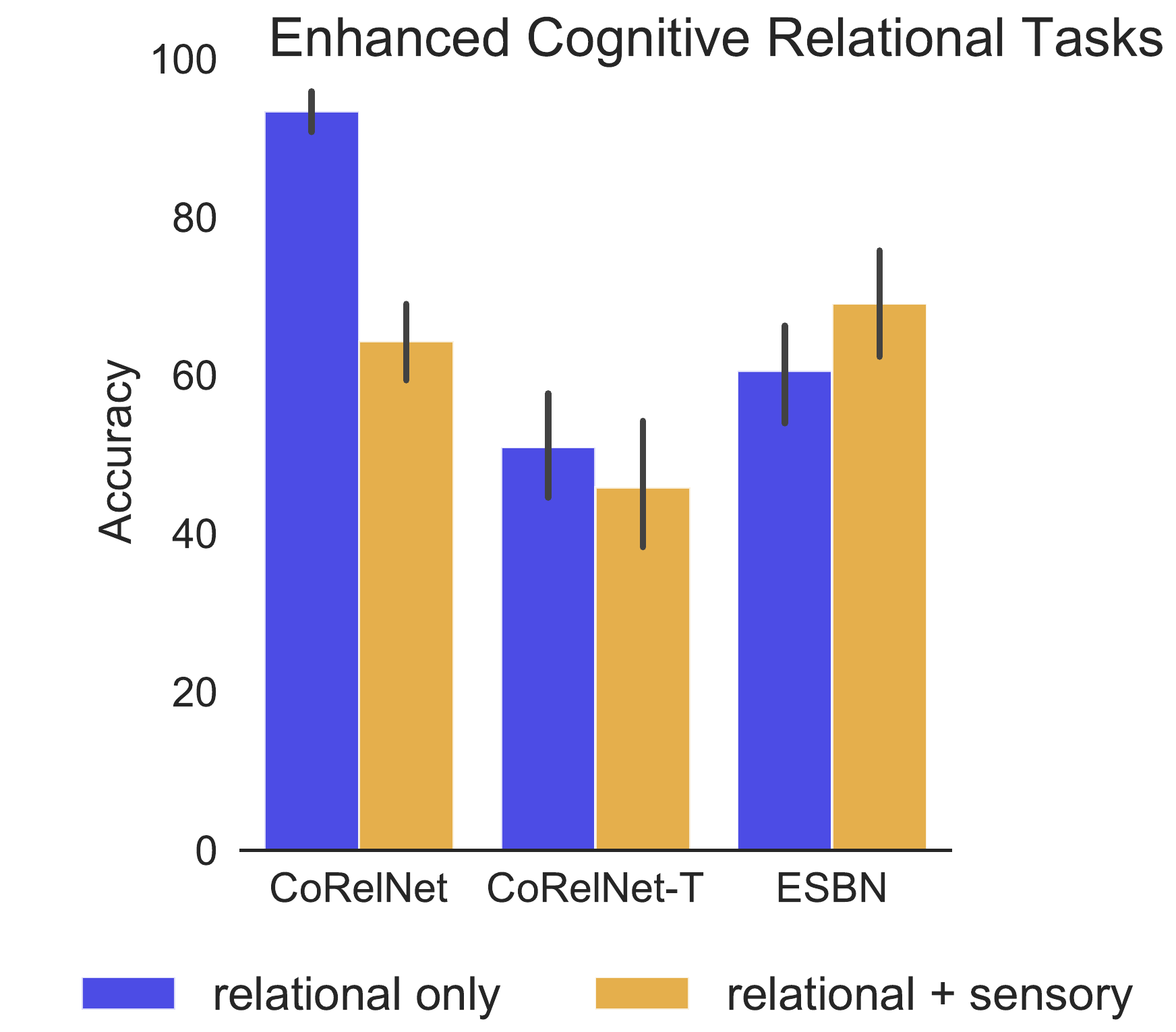}  
\end{subfigure}
\begin{subfigure}[c]{0.48\columnwidth}
  \centering
  \includegraphics[width=0.95\textwidth]{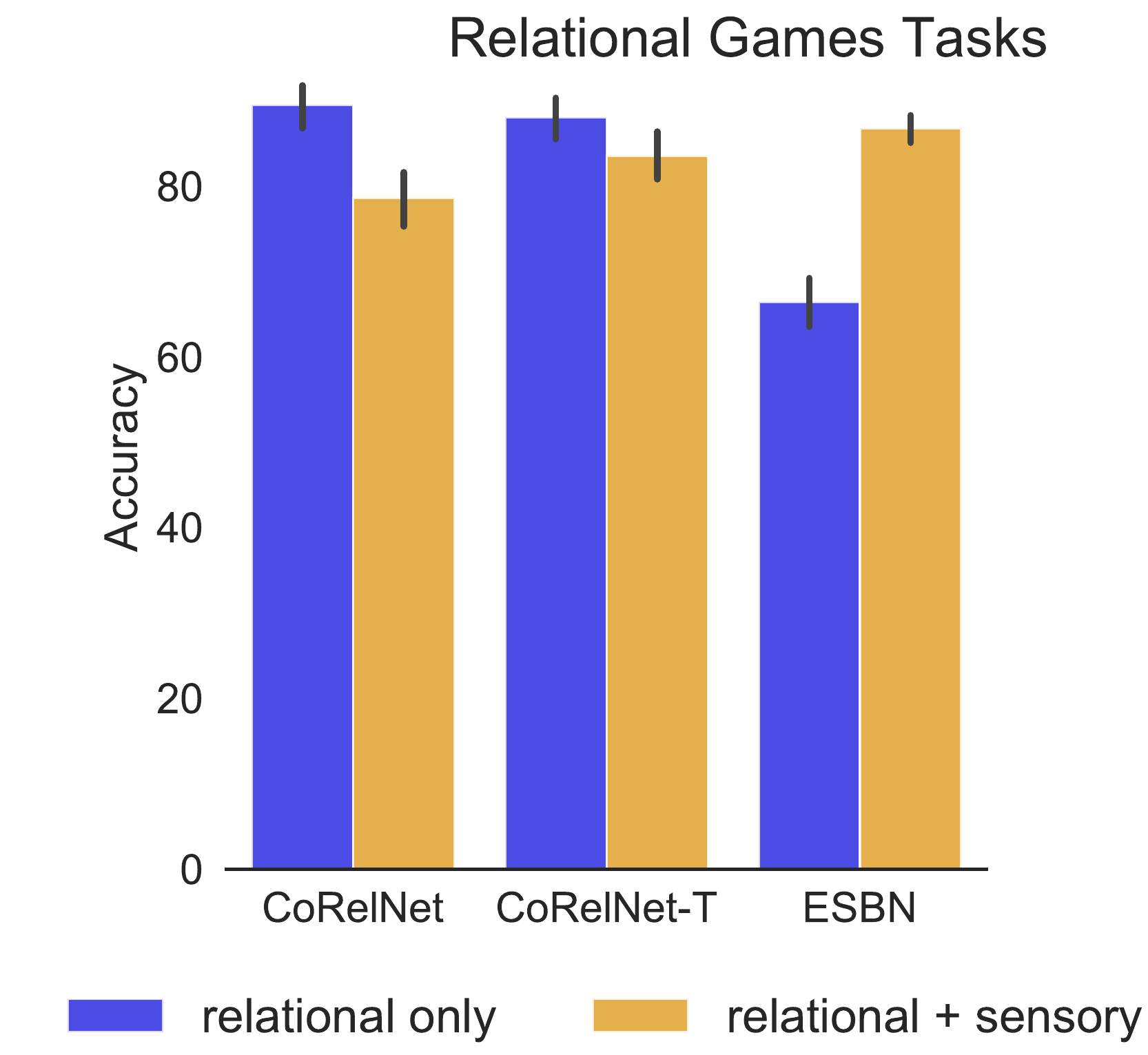}  
 
\end{subfigure}
\caption{Comparison relational only vs relational + sensory input models, all models are run on 10 seeds for each task. \textbf{Left.} Average OOD test performance across the three enhanced cognitive relational tasks same/diff6 ($m=98$), RMTS3 ($m=94$), identity rules 4 ($m=94$). \textbf{Right.} Average OOD test performance across the three enhanced cognitive relational tasks 'same', 'between', 'occurs', 'xoccurs', 'left-of', 'row matching', 'col./shape'. For the tasks involving stripes and hexominoes for OOD testing, both test performances were considered and the mean was computed, otherwise only hexominoes performance was considered. The average OOD test was taken across all tasks.}
\label{fig:additional_cat_results}
\end{figure}

\begin{figure}[t!]
\centering
\begin{subfigure}[c]{0.48\columnwidth}
  \centering
  \includegraphics[width=\textwidth]{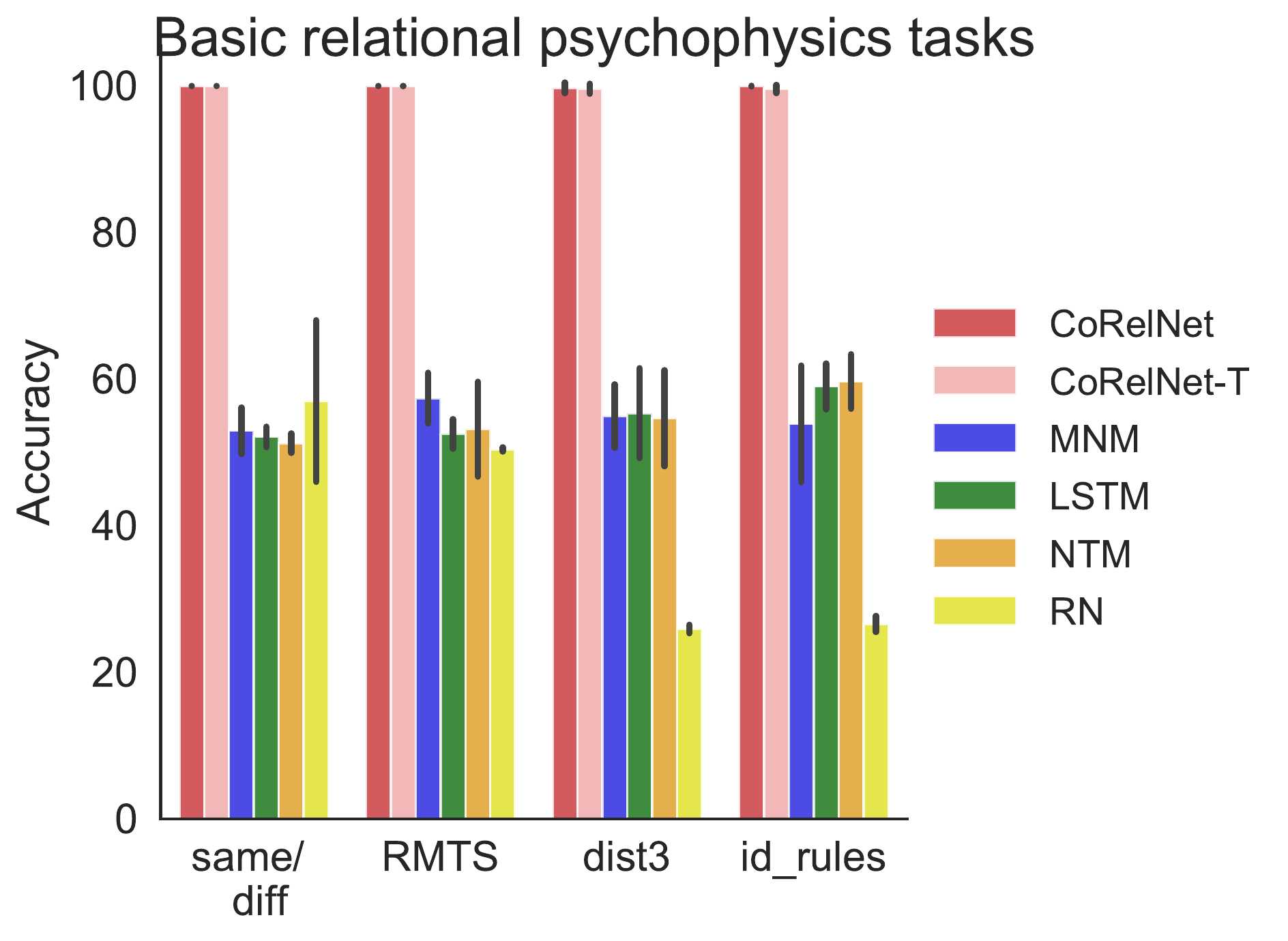}  
\end{subfigure}
\begin{subfigure}[c]{0.48\columnwidth}
  \centering

  \includegraphics[width=\textwidth]{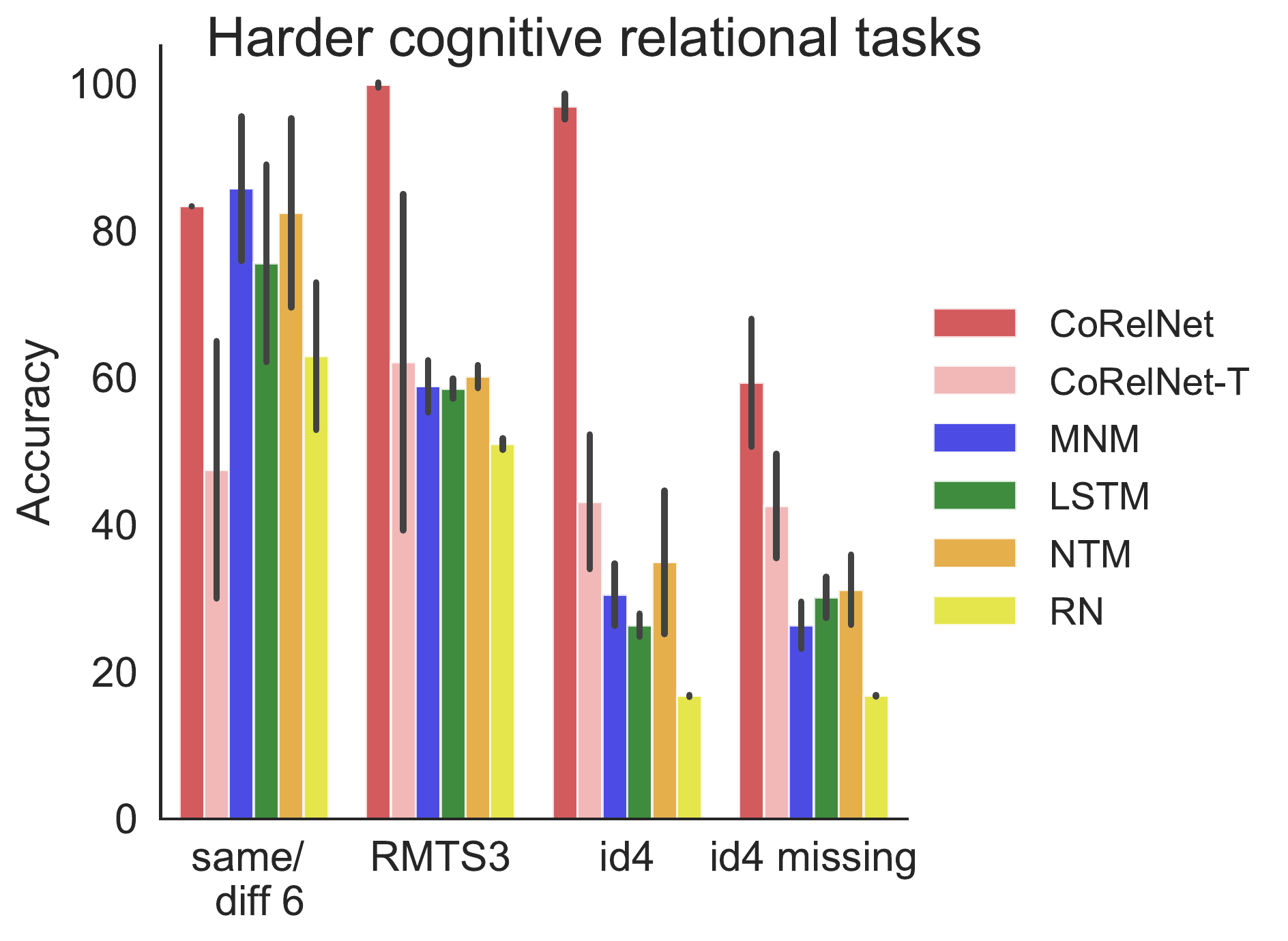}  
 
\end{subfigure}
\caption{Additional baselines are being shown, such as MNM, NTM, LSTM and RN. Results are showing OoD test accuracy averaged across $10$ seeds for each task. We took the most extreme OOD cases such as $m=98$ for same/diff and same/diff6, $m=95$ for RMTS, identity rules and dist3, and $m=94$ for identity rules 4, identity rules 4 missing,  RMTS3.}
\label{fig:additional_baselines_cog_tasks}
\end{figure}

\end{document}